\documentclass{article}
\usepackage{latexsym,amssymb,enumerate,amsmath,epsfig,amsthm,authblk,dsfont,bm}
\usepackage[margin=1in]{geometry}
\usepackage{setspace,color}
\usepackage{tikz}
\usepackage{floatrow}
\usepackage{multirow}
\usepackage{graphicx,subfig}
\usepackage[ruled]{algorithm2e}
\usepackage{epstopdf}
\usepackage{grffile}
\usepackage{makecell}
\usepackage{lineno}
\usepackage[numbers,sort&compress]{natbib}

\usepackage{hyperref}
\hypersetup{
	colorlinks=true,
	linkcolor=blue,
	citecolor=blue,
	filecolor=blue,      
	urlcolor=blue
}

\newcommand{\bc}{\mathbf{c}}
\newcommand{\RR}{\mathds{R}}

\newcommand{\bx}{\mathbf{x}}
\newcommand{\bz}{\mathbf{z}}

\newcommand{\bn}{\mathbf{n}}

\newcommand{\cL}{\mathcal{L}}
\newcommand{\cW}{\mathcal{W}}

\newcommand{\Sig}{\mathrm{Sig}}
\newcommand{\ReLU}{\mathrm{ReLU}}
\newcommand{\Per}{\mathrm{Per}}
\newcommand{\proj}{\mathrm{Proj}}

\newtheorem{remark}{Remark}[section]
\begin{document}
	\title{Double-well Net for Image Segmentation }
	\date{}
	\author{
		Hao Liu\thanks{Department of Mathematics, Hong Kong Baptist University, Kowloon Tong, Hong Kong. Email: haoliu@hkbu.edu.hk. The work of Hao Liu is partially supported by HKBU 179356, NSFC 12201530 and HKRGC ECS 22302123.} , 
		Jun Liu\thanks{ School of Mathematical Sciences, Laboratory of Mathematics and Complex Systems, Beijing Normal University,
			Beijing 100875, P.R. China. Email: jliu@bnu.edu.cn. The work of Jun Liu is partially supported by NSFC 12371527.},
		Raymond H. Chan\thanks{Lingnan University, Tuen Mun, Hong Kong SAR. Email: raymond.chan@ln.edu.hk. The work of Raymond H. Chan is partially supported by
			HKRGC GRF grants CityU1101120, CityU11309922, CRF grant C1013-21GF and HKITF MHKJFS Grant MHP/054/22.},
		Xue-Cheng Tai\thanks{Norwegian Research Centre (NORCE), Nyg\r{a}rdstangen, NO-5838 Bergen, Norway Email: xtai@norceresearch.no, xuechengtai@gmail.com. The work of Xue-Cheng Tai is partially supported by  HKRGC-NSFC Grant N-CityU214/19,  HKRGC CRF Grant C1013-21GF and NORCE Kompetanseoppbygging program.},
	}
	\maketitle
	\begin{abstract} 
		In this study, our goal is to integrate classical mathematical models with deep neural networks by introducing two novel deep neural network models for image segmentation  known as Double-well Nets. Drawing inspirations from the Potts model, our models leverage neural networks to represent a region force functional. We extend the well-know MBO (Merriman-Bence-Osher) scheme to solve the Potts model. 
		The widely recognized Potts model is approximated using a double-well potential and then solved by  an operator-splitting method, which turns out to be an extension  of the well-known MBO scheme. Subsequently, we replace the region force functional in the Potts model with a UNet-type network, which is data-driven and is designed to capture multiscale features of images, and also  introduce control variables to enhance effectiveness. The resulting algorithm is a neural network activated by a function that minimizes the double-well potential.
		What sets our proposed Double-well Nets apart from many existing deep learning methods for image segmentation is their strong mathematical foundation. They are derived from the network approximation theory and employ the MBO scheme  to approximately solve the Potts model. By incorporating mathematical principles, Double-well Nets bridge the MBO scheme and neural networks, and offer an alternative perspective for designing networks with mathematical backgrounds. Through comprehensive experiments, we demonstrate the performance of Double-well Nets, showcasing their superior accuracy and robustness compared to state-of-the-art neural networks.
		Overall, our work represents a valuable contribution to the field of image segmentation by combining the strengths of classical variational models and deep neural networks. The Double-well Nets introduce an innovative approach that leverages mathematical foundations to enhance segmentation performance.
	\end{abstract}
	
	\section{Introduction}
	Image segmentation is an important problem in image processing, computer vision, and object recognition. How to segment objects accurately from a given image has been an active research topic for a long time \cite{kass1988snakes,caselles1997geodesic,mumford1989optimal,chan2001active,bae2017augmented}. 
	In past decades, many mathematical models have been developed for image segmentation. One line of research focuses on contours and design forces to drive contours to objects' boundaries, such as the active contour model \cite{kass1988snakes} and the geodesic active contour model \cite{caselles1997geodesic}. Another class of methods, which was first introduced by Mumford and Shah in \cite{mumford1989optimal},  aims to find piece-wise smooth functions to approximate the given image. In \cite{mumford1989optimal}, the discontinuity set is the segmentation result, which is restricted to a smooth curve. The Mumford-Shah model inspired a lot of image processing models, among which a well-known one for image segmentation is the Chan-Vese model \cite{chan2001active}, in which the discontinuity set is restricted to closed curves. In \cite{osher1988fronts}, the authors use level set functions to represent the discontinuity set, which allows them to handle topology changes easily. Variants of the Chan-Vese model with various regularizers are studied in \cite{chan2005level, bae2017augmented,yan2020convexity}. 
	
	Another popular model for image segmentation is the Potts model \cite{boykov2001fast,chambolle2011first,chan2006algorithms,tai2021potts,weinan2017proposal}, which originates from statistical mechanics \cite{potts1952some} and can be taken as the generalization of the two-state Ising model for lattice \cite{pock2009convex}. The Potts model relates closely to graph cut algorithms. It is in fact a min-cut problem, which is equivalent to the max-flow and convex dual problems,  see \cite{yuan2010study,yuan2014spatially,tai2021potts,wei2018new} for some detailed explanations. Another interesting fact about the Potts model is that it can also be taken as a generalization of the Chan-Vese model \cite{sun2021efficient}. Other mathematical models for image segmentation include a smooth and threshold method \cite{cai2013two,li2020three}, conformal mapping \cite{zhang2021topology,zhang2021topology1}, to name a few.
	
	Most of the models mentioned above require solving some optimization problems. However, some models are complicated which are challenging to solve. Efficient and robust methods to solve these problems are also an active research field in image processing. A large class of numerical methods is based on the alternating direction method of multipliers (ADMM), which has been studied in \cite{glowinski2019finite}, see also some recent expositions \cite{yuan2010study,bae2017augmented,yashtini2016fast}. When the parameters are properly set, ADMM solves optimization problems very efficiently. However, parameters in ADMM need to be carefully tuned in order to get good results. Another class of methods that is not sensitive to parameters is the operator-splitting method, which decomposes a complicated problem into  subproblems so that each subproblem either has a closed-form solution or can be solved efficiently. Recently, operator-splitting methods have been successively applied in image processing \cite{deng2019new,liu2021color,liu2023elastica,duan2022fast}, surface reconstruction \cite{he2020curvature}, numerical PDEs \cite{liu2019finite,glowinski2019finite}, inverse problems \cite{glowinski2015penalization}, obstacle problem \cite{liu2023fast}, computational fluid dynamics \cite{bonito2017operator,bukavc2013fluid}. We suggest that readers refer to \cite{glowinski2016some,glowinski2017splitting} for a comprehensive discussion of operator-splitting methods. It has been shown in \cite{deng2019new,duan2022fast} that compared to ADMM, operator-splitting methods give similar (or slightly better) results but are more stable and efficient.  In fact, ADMM is a special type of operator-splitting method.
	
	In the last few years, deep neural networks have demonstrated impressive performances in many tasks, including image denoising \cite{li2023deep,li2023ewt,zhang2017beyond,yang2017bm3d} and image segmentation \cite{ronneberger2015u,zhou2018unet++,chen2018encoder}. Many network architectures are designed for image segmentation, such as UNet \cite{ronneberger2015u}, UNet++ \cite{zhou2018unet++} and DeepLabV3+ \cite{chen2018encoder}. While deep neural networks provide very good results, and in many cases, they are better than traditional image segmentation methods, they are black box algorithms, and their mathematical understanding is unclear. Making connections between deep neural networks and traditional mathematical models, as well as developing mathematically interpretative models, remain open questions. 
	
	Recently, several attempts have been made to connect mathematical models and deep neural networks. In \cite{weinan2017proposal}, neural networks are viewed as the discretization of continuous dynamical systems. PDE and ODE-motivated networks are proposed in \cite{haber2017stable,ruthotto2020deep}. The connections between deep neural networks and control problems are studied in \cite{ruiz2023neural,benning2019deep,onken2022neural}. Theories on the relation between neural ordinary differential equations and the controllability problem are established in \cite{elamvazhuthi2022neural}. Specifically, they show that for any  Lipschitz bounded vector field, neural ODEs can be used to
	approximate solutions of the continuity equation. Connections between deep neural networks with variational problems are pointed out in \cite{finlay2018lipschitz,thorpe2023deep}. The authors of \cite{finlay2018lipschitz} show that combining adversarial training and Lipschitz regularization improves adversarial robustness and is equivalent to total variation regularization. In \cite{thorpe2023deep}, the authors show that the deep layer limit of residual neural network coincides with a parameter estimation problem for a nonlinear ordinary differential equation.
	In \cite{nguyen2022fourierformer}, the authors replace the dot-product kernels in transformers with Fourier integral kernels, achieving higher accuracy in language modeling and image classification.
	MgNet, a network inspired by the multigrid method, is proposed in \cite{he2019mgnet}. The connections between neural networks and operator-splitting methods are pointed out in \cite{lan2023dosnet,liu2023connections}. Regularizers in popular mathematical models are used as priors in \cite{Jia2020,liu2022deep,Li2020b} to design networks that enable the segmentation results to have special priors. Recently, based on the Potts model, multigrid method, and operator-splitting method, the authors proposed PottsMGNet \cite{tai2023pottsmgnet}, which gives a clear mathematical explanation for existing encoder-decoder type of neural networks. 
.
	
	In this paper, we integrate the Potts model with deep neural networks and propose two networks inspired by operator-splitting methods for image segmentation. Consider the two-phase Potts model. It contains a region force term (see the first term in (\ref{eq.potts})) that is manually designed for good performance. Whether this term is optimal so that the resulting model gives the best segmentation results is an open question. In this paper, we consider a data-driven approach to determine it. We propose representing this term by networks, whose parameters will be learned from data. Starting from the Potts model, we first formulate an initial value problem (in the sense of gradient flow) to minimize the functional. Then, the initial value problem is time-discretized by operator-splitting methods. In the resulting scheme, each time stepping consists of two substeps. The first substep computes the segmentation linearly or nonlinearly, depending on the scheme. The second substep is a nonlinear step that approximately minimizes a double-well potential with a proximal term.  Such a splitting strategy is an extension of the well-known MBO scheme  of \cite{merriman1994motion,merkurjev2013mbo} as explained in  \cite{esedog2006threshold,tai2007image}.  Our proposed approach  is equivalent to a neural network consisting of several blocks: each time step corresponds to a block that is activated by approximately minimizing a double-well potential. The resulting scheme has an architecture that is similar to a convolutional neural network. We call our scheme Double-well Net (DN) and propose two variants of it: DN-I and DN-II. In general, a Double-well Net is an approximate solver for minimizing the Potts model using an extension of the  MBO scheme, where the region force term is represented by networks. It is also a bridge between classical numerical algorithms and neural networks as it unveils the connections between the well-known MBO scheme and networks.
	
	Compared to existing networks for image segmentation, such as UNet, the proposed Double-well Nets have several novelties: (1) DN-I uses a subnetwork as a bias term, while in existing networks, the bias term is a scalar. (2) DN-II uses the input image in each block instead of only at the input of the network, which is commonly done in existing networks. (3) Both DNs use an activation function derived from the double-well potential. (4) Both DNs have a mathematical background: they are operator-splitting algorithms that approximately solve the Potts model.
	PottsMGNet \cite{tai2023pottsmgnet} is another network for image segmentation which is based on the Potts model, operator splitting methods and has mathematical explanations. PottsMGNet uses operator-splitting methods together with multigrid methods to give an explanation of encoder-decoder based neural networks, and uses sigmoid function as activation. In this paper, the DNs are derived using the Potts model, the MBO scheme and network approximation theory, which have a mathematical explanation from another perspective. Furthermore, DNs use a double-well potential related operator as activation function.
	Our numerical results show that Double-well Nets give better results than the state-of-the-art segmentation networks on several datasets.
	
	This paper is structured as follows: We introduce the Potts model and derive the corresponding initial value problem in Section \ref{sec.potts}. Operator-splitting methods which are extensions of the well-known MBO scheme for solving these initial value problems and numerical discretization are discussed in Section \ref{sec.model}. We propose Double-well Nets in Section \ref{sec.doublewellnet}, and demonstrate their effectiveness by comprehensive numerical experiments in Section \ref{sec.experiments}. This paper is concluded in Section \ref{sec.conclusion}.
	
	\section{Potts model}\label{sec.potts}
	Let $\Omega\subset \mathbb{R}^2$ be a rectangular image domain. The continuous two-phase Potts model is in the form of \cite{potts1952some,chambolle2011first,yuan2010study,wei2018new,tai2021potts}
	\begin{align}
		\begin{cases}
			\min\limits_{\Sigma_0,\Sigma_1} \left\{ \frac{1}{2} \displaystyle\sum_{k=0}^1 \Per(\Sigma_k) +\sum_{k=0}^1 \displaystyle\int_{\Sigma_k} h_k(\bx) d\bx\right\},\\
			\Sigma_0, \Sigma_1\subset\Omega, \quad
			\Sigma_0\cup \Sigma_1=\Omega,\quad
			\Sigma_0\cap \Sigma_1=\emptyset,
		\end{cases}
		\label{eq.potts0}
	\end{align}
	where $\Sigma_k$ is regular enough so that its perimeter exists, denoted by $\Per(\Sigma_k)$, and $f_k(\bx)$'s are non-negative weight functions. A popular choice  of $h_k$ (as in \cite{chan2001active}) is 
	\begin{align}
		h_k(\bx)= (f(\bx)-r_k)^2/\alpha,
		\label{eq.CV}
	\end{align}
	in which $\alpha$ is a scaling parameter, and $r_k$ is the mean density of $f(\bx)$ on $\Sigma_k^*$. Here, we use $\Sigma_k^*$ to denote the `optimal' segmentation region, the ground truth segmentation of $f$ and is independent to any model. In practice, one has no information of $\Sigma_k^*$ and has to estimate $r_k$. How to estimate it is not the focus of this paper and is omitted. Here we want to emphasis that such a $r_k$ exists and only depends on $f$. As one can take $r_k$ and $\Sigma_k^*$ as functions depending on $f$, in this case, $h_k$'s are functions of the input image $f$ only. We do not require $\Sigma_k^*$ to be a minimizer of (\ref{eq.potts0}). But we hope there exists some $h_k$ so that the minimizer is close to $\Sigma_k^*$. Later, we will represent $h_k$ by a neural network.

	The Potts model for two-phase problems can be solved using binary representations as the following min-cut problem:
	\begin{align}
		\min_{v\in\{0,1\}}\int_{\Omega} F(f)vd\bx +\lambda \int_{\Omega}|\nabla v| d\bx,
		\label{eq.potts}
	\end{align}
	where $F(f)=f_1-f_0$ is called the region force depending on the input image $f$, and $\Omega$ is the domain on which the image is defined. Model (\ref{eq.potts}) requires the function $v$ to be binary. The above nonconvex  min-cut problem is equivalent to the following convex dual problem of a max-flow problem  as explained in \cite[Section 2.1]{wei2018new} and \cite[the section of (84) and (85)]{tai2021potts}:
	\begin{align}
		\min_{v\in [0,1]}\int_{\Omega} F(f)vd\bx +\lambda \int_{\Omega}|\nabla v| d\bx .
		\label{eq.potts-convex}
	\end{align}
	This is often called the convex relaxation of (\ref{eq.potts}). This model also recovers the well-known CEN model of Chan-Esedo\={g}lu-Nikolova \cite{chan2006algorithms}.
	If we use   the Ginzburg-Landau functional $\cL_{\varepsilon}$ to approximate the second term in (\ref{eq.potts}),  we then need to solve:
	\begin{align}
		\min_{v } \int_{\Omega} F(f)vd\bx +\lambda \cL_{\varepsilon}(v) d\bx,
		\label{eq.potts.relax}
	\end{align}
	with 
	\begin{align*}
		\cL_{\varepsilon}(v)=\int_{\Omega} \left[ \frac{\varepsilon}{2}|\nabla v|^2 + \frac{1}{\varepsilon} v^2(1-v)^2\right] d\bx,
	\end{align*}
	where the second term in $\cL_{\varepsilon}$ is the double-well potential. 
	It is shown that the minimizer of (\ref{eq.potts.relax}) converges to that of (\ref{eq.potts}) in the sense of Gamma-convergence as $\varepsilon\rightarrow 0$ \cite{modica1977esempio,modica1987gradient}. The Gamma-convergence of the graph-based Ginzburg-Landau functional was proved in \cite{van2012gamma}, and its applications in image processing, and data segmentation on graphs have been  studied in \cite{merkurjev2013mbo,garcia2014multiclass}. Global minimization for graph data using min-cut/max-flow approaches, c.f. \cite{yuan2010study,yuan2014spatially,bae2017augmented},   has also been studied  in \cite{merkurjev2015global}.
	
	Denote the minimizer of (\ref{eq.potts.relax}) by $u$. It satisfies the optimality condition
	\begin{align}
		F(f)-\lambda\varepsilon\nabla^2u +\frac{2\lambda}{\varepsilon}(2u^3-3u^2+u)=0,
		\label{eq.potts.relax.EL}
	\end{align}
	where $\nabla^2$ is the Laplacian operator.
	One way to solve (\ref{eq.potts.relax.EL}) for $u$ is to associate it with the following initial value problem (in the sense of gradient flow):
	\begin{align}
		\begin{cases}
			\frac{\partial u}{\partial t} =-F(f)+\lambda\varepsilon\nabla^2u -\frac{2\lambda}{\varepsilon}(2u^3-3u^2+u) \mbox{  in  }\Omega\times (0,T],\\
			\frac{\partial u}{\partial \bn}=0 \mbox{ on } \partial \Omega, \\
			u(0)=u_0 \mbox{  in  } \Omega,
		\end{cases}
		\label{eq.potts.relax.gf}
	\end{align}
	for some initial condition $u_0$ and fixed time $T$. Then solving (\ref{eq.potts.relax.EL}) is equivalent to finding the steady state solution of (\ref{eq.potts.relax.gf}).
	
	\section{The proposed models} \label{sec.model}
	In classical mathematical models for image segmentation, the operator $F$ is carefully designed, as (\ref{eq.CV}) in the Chan-Vese model. These models have demonstrated great performances in segmenting general images. For a specific type of images, certain choices of $F$ may give improved performances. In this paper, we consider a data-driven method to learn $F$.  
	
	Suppose we are given a training set of images $\{f_i\}_{i=1}^I$  with their foreground-background segmentation masks $\{g_i\}_{i=1}^I$ so that there exists some $F$ such that for each $f_i$, the minimizer of (\ref{eq.potts.relax}) is close to $g_i$. We will learn a data-driven operator $F$ so that for any given image $f$ with similar properties as the training set, the steady state of (\ref{eq.potts.relax.gf}) is close to its segmentation $g$. 
	Note that (\ref{eq.potts.relax.gf}) is an initial value problem of $u$. In practice, it might be difficult to solve (\ref{eq.potts.relax.gf}) unitl the steady state. Instead, a more practical way is to solve it until a finite time $t=T$, and use the solution (denoted by $u(\bx,T)$) as the segmentation. However, $u(\bx,T)$ may be far away from the ground truth segmentation.  To better control the evolutionary behavior of $u$ and borrow some of the ideas from \cite{tai2023pottsmgnet}, we introduce control variables $W(\bx,t), b(t)$ into (\ref{eq.potts.relax.gf}) and consider:
	\begin{align}
		\begin{cases}
			\frac{\partial u}{\partial t} =-F(f)+\lambda\varepsilon\nabla^2u -\frac{2\lambda}{\varepsilon}(2u^3-3u^2+u)+W(\bx,t)*u+b(t) \mbox{ in } \Omega\times(0,T],\\
			\frac{\partial u}{\partial \bn}=0 \mbox{ on } \partial \Omega,\\
			u(0)=u_0 \mbox{ in } \Omega,
		\end{cases}
		\label{eq.control}
	\end{align}
	where $*$ stands for the convolution operator and $W(\bx,t)$ is a convolution kernel depending on $\bx$ and $t$, and $b(t)$ is some function of $t$. The control variable will adjust the evolution of $u$ and  steer $u(\bx,T)$ to be close to the groundtruth segmentation.
	The control variables will be learned from the given dataset. We remark that here $W(\bx,t)$ is a convolution kernel instead of the double-well potential in many works related to the Ginzburg-Landau functional.
	
	We are going to present two models to solve (\ref{eq.control}). The first model is an operator-splitting scheme for (\ref{eq.control}), see Section \ref{sec.DN-I}. The second model generalize the first one in which we allow more complicated interactions among $F(f), W(\bx,t), b(t)$ and $u$ and represent these interactions by one operator, see Section \ref{sec.DN-II}. In our methods, we represent $F$ in the first model and the new operator in the second model as a neural network. Denote $\cW$ as the collection of all parameters to be determined from the data, i.e., the parameters in $F$ and the control variables $W(\bx,t), b(t)$. The initial condition $u_0=H(f)$ is taken as  a function of the input image $f$ with a properly chosen $H$, c.f.  (\ref{eq.u0}). Then the  solution of (\ref{eq.control}) at time $t=T$ only depends on $f$ and $\cW$. For each $f, \cW$, denote the solution for (\ref{eq.control}) at time $T$ by $R(\cW;f)$, which is a function map from $f$ to $R(\cW;f)$ (the solution at $T$). We will determine $\cW$ by solving 
	\begin{align}
		\min_{\cW} \frac{1}{K}\sum_{k=1}^K \ell (R(\cW;f_k),g_k),
		\label{eq.NNopti}
	\end{align}
	where $\ell(\cdot,\cdot)$ is a loss function measuring the differences between its arguments. Popular choices of the loss functional include the hinge loss, logistic loss, and $L^2$ norm, see \cite{rosasco2004loss} for a discussion of different loss functions.
	
	In this section, we focus on the solvers for the control problem (\ref{eq.control}). Specifically, we consider model (\ref{eq.control}) and a variant model and propose operator-splitting methods to solve them. We will show that the resulting scheme has a similar architecture as deep neural networks.
	
	\begin{remark}
		There is no problem for our proposed methods if we replace $W(\bx,t)*u+b(t)$ with some other general linear operator on $u$ in (\ref{eq.control}). The reason for choosing the convolution operator and the bias correction here is due to their success in applications related to  Deep Convolutional Neural Networks. 
	\end{remark}

	\subsection{Model I}
	\label{sec.DN-I}
	The first model we study is (\ref{eq.control}). 
	Note that it is well suited to be numerically solved by operator-splitting methods. For a comprehensive introduction to operator-splitting methods, we refer readers to \cite{glowinski2016some,glowinski2017splitting}. In this paper, we adopt the Lie scheme. The terms on the right-hand side of (\ref{eq.control}) can be classified into linear terms and nonlinear terms of $u$, based on how we split them into two substeps. Let $\tau$ be the time step. For $n\geq0$, we denote $t^n=n\tau$ and our numerical solution at $t^n$ by $u^n$.   Set $u_0=H(f)$ with a properly chosen $H$, c.f. (\ref{eq.u0}).
	We compute $u^{n+1}$ from $u^n$ via two substeps: $u^n\rightarrow u^{n+1/2}\rightarrow u^{n+1}$. The first substep focuses on linear terms; the second substep focuses on nonlinear terms. The details are as follows:\\
	\textbf{Substep 1}: Solve
	\begin{align}
		\begin{cases}
			\frac{\partial u}{\partial t} =-F(f)+\lambda\varepsilon\nabla^2u +W(\bx,t)*u+b(t) \mbox{ in } \Omega\times(t^n,t^{n+1}],\\
			\frac{\partial u}{\partial \bn}=0 \mbox{ on } \partial\Omega,\\
			u(t^n)=u^n \mbox{ in } \Omega,
		\end{cases}
		\label{eq.split.1}
	\end{align}
	and set $u^{n+1/2}=u(t^{n+1})$.\\
	\textbf{Substep 2}: Solve 
	\begin{align}
		\begin{cases}
			\frac{\partial u}{\partial t} = -\frac{2\lambda}{\varepsilon}(2u^3-3u^2+u) \mbox{ in } \Omega\times(t^{n},t^{n+1}],\\
			u(t^n)=u^{n+1/2} \mbox{ in } \Omega,
		\end{cases}
		\label{eq.split.2}
	\end{align}
	and set $u^{n+1}=u(t^{n+1})$.
	
	Scheme (\ref{eq.split.1})--(\ref{eq.split.2}) is semi-constructive since we still need to solve the two initial value problems. In this paper, we use a one-step forward Euler scheme to time discretize (\ref{eq.split.1}) and a one-step backward Euler scheme to time discretize (\ref{eq.split.2}):
	\begin{align}
		&\begin{cases}
			\frac{u^{n+1/2}-u^n}{\tau} =-F(f)+\lambda\varepsilon\nabla^2u^n +W^n*u^n+b^n \mbox{ in } \Omega,
			\\
			\frac{\partial u^{n+1/2}}{\partial \bn}=0 \mbox{ on } \partial \Omega,
		\end{cases}\label{eq.split.1.dis}\\
		&	\frac{u^{n+1}-u^{n+1/2}}{\tau} = -\frac{2\lambda}{\varepsilon}(2(u^{n+1})^3-3(u^{n+1})^2+u^{n+1}) \mbox{ in } \Omega,
		\label{eq.split.2.dis}
	\end{align}
	where the notations $W^n=W(\bx,t^n),b^n=b(t^n)$ are used.
	
	\begin{remark}
		In (\ref{eq.split.1.dis}), $\lambda\varepsilon \nabla^2 $ is a penalty term on the smoothness of the segmentation's boundary, and $W^n$ is a control variable that is learnable. The term $\lambda\varepsilon \nabla^2 $ can be absorbed by $W^n$ since both of them are linear in $u^n$. However, in Model I, we write $\lambda\varepsilon \nabla^2 u$ explicitly to explicitly drive the segmentation boundary to be smooth.
	\end{remark}
	
	\subsection{Model II}
	\label{sec.DN-II}
	In model (\ref{eq.control}), the functional $F(f)$ is fixed through all iterations. In our implementation, we will use a UNet class (see Section \ref{sec.doublewellnet} for details) to represent $F$, and learn $F$ and the control variables $W$ and $b$ from data.
In the second model, we learn a more general functional  $G(u,f,\bx,t)$ as shown below in (\ref{eq.control2}). 
In our implementation, we will use a UNet class to represent $G$, and learn it from data, without any assumption on the specific form of $G$. 
Our second model is as follows
\begin{align}
\begin{cases}
	\frac{\partial u}{\partial t} =\lambda\varepsilon\nabla^2u -\frac{2\lambda}{\varepsilon}(2u^3-3u^2+u)+G(u,f,\bx,t) \mbox{ in } \Omega\times(t^n,t^{n+1}],\\
	\frac{\partial u}{\partial \bn}=0 \mbox{ on } \partial\Omega,\\
	u(0)=u_0 \mbox{ in } \Omega.
\end{cases}
\label{eq.control2}
\end{align}
Note that $-F(f)+W(\bx,t)*u+b(t)$ in (\ref{eq.control}) is a special case of $G(u,f,\bx,t)$ in (\ref{eq.control2}). A linear relationship among $F(f), W(\bx,t), b(t)$ and $u$ is enforced in (\ref{eq.control}). By replacing $-F(f,t)+W(\bx,t)*u+b(t)$ by the functional $G(u,f,\bx,t)$, we allow more complicated interactions among them. Thus (\ref{eq.control2}) is a more general model.   

Similar to (\ref{eq.control}), we choose $u_0=H(f)$ as a function of the input image $f$ with a properly chosen $H$, c.f.  (\ref{eq.u0}), and use a Lie scheme to time discretize (\ref{eq.control2}). 
The splitting strategy is the same as (\ref{eq.split.1})--(\ref{eq.split.2}) except the first substep becomes
\begin{align}
\begin{cases}
	\frac{\partial u}{\partial t} =\lambda\varepsilon\nabla^2u +G(u,f,\bx,t) \mbox{ in } \Omega\times(t^n,t^{n+1}],\\
	\frac{\partial u}{\partial \bn}=0 \mbox{ on }\partial\Omega,\\
	u(t^n)=u^n \mbox{ in } \Omega.
\end{cases}
\label{eq.split2.1}
\end{align}
We time discretize Substep 1 by a forward Euler method and Substep 2 by a backward Euler method. The time-discretized scheme reads as
\begin{align}
&\begin{cases}
	\frac{u^{n+1/2}-u^n}{\tau} =\lambda\varepsilon\nabla^2u^n +G^n(u^n,f) \mbox{ in } \Omega,\\
	\frac{\partial u^{n+1/2}}{\partial \bn}=0 \mbox{ on } \partial\Omega,
\end{cases}
\label{eq.split2.1.dis}\\
&	\frac{u^{n+1}-u^{n+1/2}}{\tau} = -\frac{2\lambda}{\varepsilon}(2(u^{n+1})^3-3(u^{n+1})^2+u^{n+1}) \mbox{ in } \Omega,
\label{eq.split2.2.dis}
\end{align}
where $G^n(u,f)=G(u,f,\bx,t^n)$.

The main difference between Model I and Model II is how we treat $F(f)$ and the control variables. Model I is directly derived from (\ref{eq.control}). It is linear in the control variables, and the so called region force  $F(f)$ is independent of  $u$ and $t$, i.e., $F(f)$ stays the same over time. Model II generalizes Model I, in which we allow more complicated interactions among $u,F(f)$ and the control variables. The interaction complexity is determined by the functional used to represent $G(u,f,\bx,t)$. In later sections, we will use a network to represent $F(f)$ for Model I and $G(u,f,\bx,t)$ for Model II.
\begin{remark}
There are many ways to combine and decompose terms in the right-hand side of (\ref{eq.control}), among which Model I and II are just two special ones that provide good results. As we will show in Section \ref{sec.doublewellnet}, both Model I and II have similar architectures as some neural networks. Based on Model I, we further explore Model II because Model II can provide similar results with fewer parameters. 
\end{remark}

\begin{remark}
The operator-splitting schemes for Model I and II are extensions of  the Allen-Cahn-type MBO scheme \cite{merriman1994motion}, as explained in  \cite{esedog2006threshold,tai2007image}. The first substep focuses on linear operators and the second substep deals with nonlinear projection operators. 
In this work, only two-phase image segmentation is considered. Following the approaches proposed in \cite{tai2007image}, there is no problem to extend the method in this work to multiphase image segmentation problems using deep neural networks similar to Model I and II. 
\end{remark}

\begin{remark}
In this work, we present our method for two-phase  image segmentation. For graph data,  each vertex of the graph could be a high dimension vector. Using the ideas presented in this work, we can combine some existing graph neural networks, such as \cite{te2018rgcnn,defferrard2016convolutional},  with the graph 
Ginzburg-Landau model of \cite{merkurjev2013mbo,garcia2014multiclass}
and the graph min-cut/max-flow approach of \cite{merkurjev2015global,wei2018new,yin2018effective}    to get some graph neural network similar to Model I and II for high-dimensional data analysis. 
\end{remark}

\subsection{Connections to neural networks}
In Model I and II, for each time step, our scheme has two substeps: the first substep is linear operation in $u^{n}$, and the second substep is a nonlinear operation. This structure is the same as the building block (layers) of neural networks (NN). An NN usually consists of several layers. Each layer conducts the computation 
\begin{align}
\sigma(Ax+b),
\label{eq.NN}
\end{align}
where $A$ is a weight matrix, $b$ is a bias term (a vector), and $\sigma$ is a nonlinear activation function applied pointwisely. In (\ref{eq.NN}), when the input to $\sigma$ is a linear operation, it is similar to our first substep; and when the operation of $\sigma$ is nonlinear, it is similar to our second substep.

\subsection{Accommodate periodic boundary conditions}
In image processing, it is common to use the periodic boundary conditions. Let $\Omega$ be a rectangular domain $[0,L_1]\times [0,L_2]$. We denote the two spatial directions by $x_1,x_2$.  To accommodate the periodic boundary conditions, we only need to replace  
(\ref{eq.split.1.dis}) by
\begin{align}
\begin{cases}
	\frac{u^{n+1/2}-u^n}{\tau} =-F(f)+\lambda\varepsilon\nabla^2u^n +W^n*u^n+b^n \mbox{ in } \Omega,
	\\
	u^{n+1/2}(0,x_2)=u^{n+1/2}(L_1,x_2), \ 0\leq x_2\leq L_2,\\
	u^{n+1/2}(x_1,0)=u^{n+1/2}(x_1,L_2), \ 0\leq x_1\leq L_1,
\end{cases}
\label{eq.split.1.dis.periodic}
\end{align}
and replace (\ref{eq.split2.1.dis}) by
\begin{align}
\begin{cases}
	\frac{u^{n+1/2}-u^n}{\tau} =\lambda\varepsilon\nabla^2u^n +G^n(u^n,f) \mbox{ in } \Omega,\\
	u^{n+1/2}(0,x_2)=u^{n+1/2}(L_1,x_2), \ 0\leq x_2\leq L_2,\\
	u^{n+1/2}(x_1,0)=u^{n+1/2}(x_1,L_2), \ 0\leq x_1\leq L_1.
\end{cases}
\label{eq.split2.1.dis.periodic}
\end{align}
In the rest of this paper, we always assume that $u$ satisfies the periodic boundary condition. 
\subsection{Spatial discretization}
\label{sec.spatial}

In our discretization, we use spatial steps $\Delta x_1=\Delta x_2=h$ for some $h>0$. We denote the spatially-discretized $u^n$ and $W^n$ by $\widehat{u}^n$ and $\widehat{W}^n$, respectively.

From (\ref{eq.split.1.dis.periodic}), an updating formula of $u^{n+1/2}$ is given as
\begin{align*}
u^{n+1/2}=u^n-\tau F(f)+\tau\lambda\varepsilon(\nabla^2 u^n)+\tau W^n*u^n+\tau b^n.
\end{align*}
Note that $\nabla^2 u^n$ is the Laplacian of $u^n$. We approximate it by central difference, which can be realized by convolution $\widehat{W}_{\Delta}*\widehat{u}^n$ with
\begin{align*}
\widehat{W}_{\Delta}=\frac{1}{h^2}\begin{bmatrix}
	0 & 1 & 0\\
	1 & -4 & 1\\
	0 & 1 & 0
\end{bmatrix}.
\end{align*}
Then the updating formula becomes
\begin{align}
\widehat{u}^{n+1/2}=\widehat{u}^n-\tau F(f)+\tau\lambda\varepsilon (\widehat{W}_{\Delta}* \widehat{u}^n)+\tau \widehat{W}^n*\widehat{u}^n+\tau b^n.
\label{eq.split.1.formula}
\end{align}
Similarly, for $u^{n+1/2}$ in model (\ref{eq.split2.1.dis.periodic}), the updating formula is 
\begin{align}
\widehat{u}^{n+1/2}=\widehat{u}^n+\tau\lambda\varepsilon (\widehat{W}_{\Delta}* \widehat{u}^n)+\tau G^n(\widehat{u}^n,f).
\label{eq.split2.1.formula}
\end{align}

For updating $\widehat{u}^{n+1}$, note that in continuous case $u^{n+1}$ is the minimizer of 
\begin{align}
\int_{\Omega} \frac{1}{2}(u-u^{n+1/2})^2 + \frac{\tau\lambda}{\varepsilon} u^2(1-u)^2 d\bx
\label{eq.step2.min}
\end{align}
and (\ref{eq.split.2.dis}) is the Euler-Lagrange equation of (\ref{eq.step2.min}). Denote $\alpha=2\tau\lambda/\varepsilon$. Then $u^{n+1}$ satisfies
\begin{align}
(1+\alpha)u^{n+1}=u^{n+1/2}-\alpha (2(u^{n+1})^3-3(u^{n+1})^2).
\end{align}
After discretizing the equation above, we solve for $\widehat{u}^{n+1}$ pointwisely using a fixed-point iteration: For $i=1,...,M_1, j=1,...,M_2$, set $\widehat{v}^0=\widehat{u}^{n+1/2}$, and we update $\widehat{v}^m\rightarrow v^{m+1}$ as
\begin{align}
\widehat{v}^{m+1}=\frac{1}{1+\alpha}(\widehat{u}^{n+1/2}-\alpha (2(\widehat{v}^{m})^3-3(\widehat{v}^{m})^2)).
\label{eq.fixed}
\end{align}
Proving the convergence of (\ref{eq.fixed}) requires a dedicated analysis of the problem, for which we leave as our future work. Empirically, when the time step $\tau$ (as well as $\alpha$) is sufficiently small, we observe that iteration (\ref{eq.fixed}) converges. A discussion on how to numerically improve its convergence behavior is presented in Section \ref{sec.DNI}.
Let $\widehat{v}^*$ denote the point $\widehat{v}^{m}$ converges to. We then set $\widehat{u}^{n+1}=v^*$.  

\section{Double-well net} \label{sec.doublewellnet}
In our framework, we solve (\ref{eq.NNopti}) to learn $\cW$, which contains the parameters in $F(f)$ and the control variables $W(\bx,t^n),b(t^n)$ for Model I, or the parameters in $G(u,f,\bx,t^n)$ for Model II. 
To learn $F(f)$ and $G(u,f,\bx,t^n)$, we need to assume a general form for them and then learn their weights by solving (\ref{eq.NNopti}). In this paper, we will use neural networks to represent $F(f)$ and $G(u,f,\bx,t^n)$.

The representation theory for neural networks has been an active topic in past decades, see \cite{barron1993universal,yarotsky2017error,  hon2022simultaneous,zhou2020universality,oono2019approximation,liu2021besov, liu2022benefits,liu2022deepoperator,song2023approximation}. These works show that when the network size (depending on the width and depth) is sufficiently large, neural networks can approximate functions, functionals, or operators with certain regularity to arbitrary accuracy
If the target function has low-dimensional structures, the network size can be reduced to maintain the same accuracy, i.e., the curse of dimensionality is mitigated, see \cite{chen2019efficient,liu2021besov, liu2022benefits,chui2018deep, chen2022nonparametric,shen2019deep} for details.

For image segmentation, many neural networks have demonstrated great performances. In this paper, we adopt the well-known UNet \cite{ronneberger2015u} type architecture  which has  provided remarkable results in various segmentation tasks \cite{huang2020unet,cao2022swin}. In this section, we first introduce a class of UNet architecture. This class is then used to represent $F(f)$ and $G(u,f,\bx,t^n)$ in the proposed models.

\subsection{UNet class}
UNet has an encoder-decoder architecture and is proposed in \cite{ronneberger2015u, tai2023pottsmgnet} for medical image segmentation. UNet's architecture consists of three parts: an encoding part, a decoding part, and a bottleneck part. 
\begin{itemize}
\item The encoding part consists of several convolutional layers and pooling layers. Given an image $f$, this part gradually reduces the resolution of the image while increasing the number of channels. Specifically, for every two layers, the number of channels is doubled and a pooling layer is used to reduce the resolution. 

Here the number of channels corresponds to the third dimension of the output. Consider a convolution layer with output $\bz\in \mathbb{R}^{d_1\times d_2 \times d_3}$. The number of channels is $d_3$. Each $\bz(:,:,i)$ for $i=1,...,d_3$ is the result of computing the convolution of the input with a convolution kernel. Thus the number of channels also determines the number of convolution kernels a convolution layer contains.

\item Following the encoding part, the bottleneck part downsamples the output of the encoding part again, doubles the number of channels, and conducts two convolutional layers. The result is then upsampled and passed to the decoding part. 
\item The decoding part consists of convolutional layers and upsampling layers. It upsamples the output of the bottleneck part from the coarsest resolution to the original image resolution and gradually reduces the number of channels. Specifically, for every two layers, the number of channels is halved and an upsampling layer is used to increase the resolution.
\item Besides the three parts mentioned above, there are skip connections between the encoding and decoding parts. For each resolution, there are two convolutional layers in the encoding part and two in the decoding part. Skip connections pass the output of the layers in the encoding part directly to the corresponding layers of the decoding part.
\end{itemize}

In UNet, there are five levels of resolutions: four levels for the encoding and decoding part and one level for the bottleneck part. For each resolution, there are two convolutional layers in the encoding and decoding part (or the bottleneck part). Each downsampling operator reduces the resolution of each dimension by half, and each upsampling operator doubles the resolution of each dimension. Every time the intermediate result is downsampled, the number of channels doubles. Every time the intermediate result is upsampled, the number of channels is reduced by half.

\begin{figure}[t!]
\centering
\includegraphics[width=\textwidth]{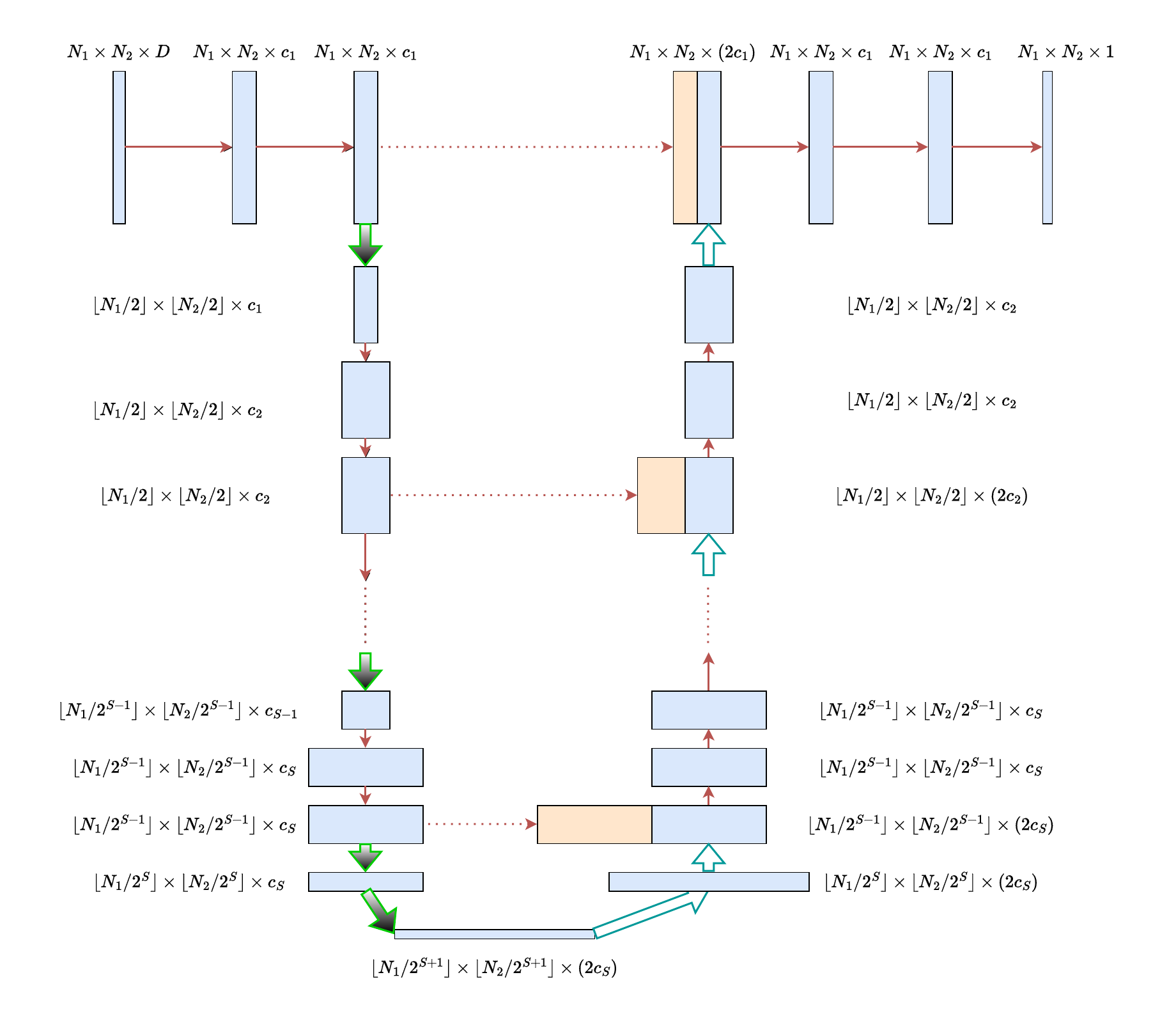}
\caption{Illustration of UNet type network with input of size $N_1\times N_2\times D$. The left branch is the encoding part, the right branch is the decoding part, and the bottom rectangle denotes the bottleneck. $S$ denotes the number of resolution scales in the encoding part and decoding part. $c_k$  denotes the number of channels at resolution scale $k$. Wide arrows with gradient shadow represent downsampling operations. Wide arrows without gradient shadow represent upsampling operations. Horizontal dashed arrows represent skip connections. The orange rectangles denote the outputs of the encoding part that are passed to the decoding part via the skip connections. The length and width of the rectangle represent the output resolution and number of channels, respectively.}
\label{fig.UNetClass}
\end{figure}

Based on the architecture of UNet, we define a more general class in which we have the flexibility to control the number of resolution levels and number of channels. For each resolution level for the encoding and decoding parts, we manually set the number of channels. For the bottleneck part, we double the number of channels of the coarsest resolution level of the encoding and decoding parts. Considering the resolution levels for the encoding and decoding parts in the order from finest to the coarsest, we denote their corresponding number of channels as a vector $\bc=[c_1,...,c_S]$ for some positive integers $\{c_s\}_{s=1}^S$, where $S$ denotes the number of resolution levels. Every network in this class can be fully characterized by the channel vector $\bc$: given a $\bc\in \RR^S$, the corresponding network has $S+1$ resolution levels, $c_s$ channels at resolution level $s$ for $1\leq s \leq S$, and $2c_S$ channels at resolution level $S+1$. For an input with dimension $N_1\times N_2\times D$, the structure of such a class is illustrated in Figure \ref{fig.UNetClass}. The architecture is designed to capture multiscale features of images: each resolution level corresponds to features of one scale. For the original UNet, it has $\bc=[64,128,256,512]$. In our model, we will choose $F(f)$ and $G(u,f,\bx,t^n)$'s in the UNet class with different $\bc$'s.

\subsection{Double-well net for model (\ref{eq.control})} 
\label{sec.DNI}

Assume $f$ is an image of size $N_1\times N_2\times D$.  We approximate $F(f)$ by a UNet class specified by channel vector $\bc$ (to be specified in Section \ref{sec.experiments}).

\textbf{A double-well block I}.
We next define a double-well block I (DB-I), which represents a step that updates $\widehat{u}^n$ to $\widehat{u}^{n+1}$ using (\ref{eq.split.1.formula}) and (\ref{eq.split.2.dis}). We first compute $\widehat{u}^{n+1/2}$ according to (\ref{eq.split.1.formula}) as
\begin{align}
\widehat{u}^{n+1/2}=\widehat{u}^n-\tau F(f)+\tau \lambda \varepsilon(\widehat{W}_{\Delta}*\widehat{u}^n)+\tau (\widehat{W}^n*\widehat{u}^n+b^n),
\end{align}
where $\widehat{W}^n$ and $b^n$ are the convolution kernel and bias in the $n$-th residual block, respectively. To compute $\widehat{u}^{n+1}$, one can solve the fixed-point updating  (\ref{eq.fixed}) with fixed number of iterations.

For efficiency, deep neural networks are trained using stochastic gradient descent with back-propagation.  
Since all we want is an approximate minimizer of (\ref{eq.potts.relax}), intermediate solutions ($\widehat{u}^n$'s) are not so important. If the scheme converges, it is not necessary to solve the problem (\ref{eq.step2.min}) exactly at every step. For practical consideration, to make the training more efficient, we solve for $\widehat{u}^{n+1}$ by using only few steps of the fixed-point iteration (\ref{eq.fixed}). In our experiments, three iterations of the fixed-point iteration already make the segmentation result good.

For robustness, note that the second fractional step (\ref{eq.split.2.dis}) solves (\ref{eq.step2.min}). Pointwisely, the solver of (\ref{eq.step2.min}) drives $\widehat{u}^{n+1/2}$ towards 1 for $\widehat{u}^{n+1/2}>0.5$, and drives $\widehat{u}^{n+1/2}$ towards 0 for $\widehat{u}^{n+1/2}<0.5$. Our numerical experiments show that the fixed-point iteration (\ref{eq.fixed}) is robust (convergent) for large time steps when $0\leq \widehat{u}^{n+1/2}\leq 1$.  
However, if $\widehat{u}^{n+1/2}$ is larger than 1 or smaller than 0, the fixed-point iteration may diverge. To make the algorithm robust (convergent for large time step), especially when $\widehat{u}^{n+1/2}>1$ or $\widehat{u}^{n+1/2}<0$, one has to use a very small $\alpha$ (recall that $\alpha=2\tau\lambda/\varepsilon$ is defined in Section \ref{sec.spatial}), implying a small time step $\tau$. With few fixed point iterations and a small $\tau$, the updated $\widehat{u}^{n+1}$ and $\widehat{u}^{n+1/2}$ are very close to each other, making the effect of this substep negligible. 

To use a relatively large $\tau$ while keeping the algorithm stable, we need to restrict $\widehat{u}^{n+1/2}$ in $[0,1]$. In this paper, we consider two methods to achieve that. In the first method, we use a projection operator to project $\widehat{u}^{n+1/2}$ to $[0,1]$ before it is passed to the fixed point iteration. For any value $a$, we project it to $[0,1]$ by solving
$$
\min_{z\in[0,1]} (z-a)^2. 
$$
The explicit solution is given by the following projection operator:
\begin{align*}
\proj_{[0,1]}(a)=\begin{cases}
	0 & \mbox{ if } a\leq 0,\\
	a & \mbox{ if } 0<a<1,\\
	1 & \mbox{ if } a\geq1.
\end{cases}
\end{align*}
Then $\proj_{[0,1]}(\widehat{u}^{n+1/2})$ is applied pointwisely and projects (truncates) $\widehat{u}^{n+1/2}$ at each pixel to $[0,1]$. Denote the fixed-point iteration (\ref{eq.fixed}) with $\gamma$ iterations by $Q_{\gamma}$. We update $\widehat{u}^{n+1}$ as
\begin{align}
\widehat{u}^{n+1}=Q_{\gamma}\circ \proj_{[0,1]}(\widehat{u}^{n+1/2}).
\label{eq.activateQ.proj}
\end{align}
The projection operator outputs a real number in $[0,1]$ and pointwisely preserves the relation between $|\widehat{u}^{n+1/2}-0|$ and $|\widehat{u}^{n+1/2}-1|$: if $|\widehat{u}^{n+1/2}-0|<|\widehat{u}^{n+1/2}-1|$, then $|\proj_{[0,1]}(\widehat{u}^{n+1/2})-0|<|\proj_{[0,1]}(\widehat{u}^{n+1/2})-1|$. In other words, if $\widehat{u}^{n+1/2}$ is closer to 0 than 1, then $\proj_{[0,1]}(\widehat{u}^{n+1/2})$ is closer to 0 than 1. Applying $\proj_{[0,1]}(\widehat{u}^{n+1/2})$ before the fixed-point iteration has two benefits: (i) Since the output of $\proj_{[0,1]}(\widehat{u}^{n+1/2})$ is in $[0,1]$, we can use a larger time step $\tau$ while having a convergent fixed-point iteration. (ii) The relation among $\widehat{u}^{n+1/2},0$ and 1 is preserved. When $\tau$ is sufficiently small, we have $Q_{\infty}(\widehat{u}^{n+1/2})=Q_{\infty}\circ \proj_{[0,1]}(\widehat{u}^{n+1/2})$ pointwisely: When $\widehat{u}^{n+1/2}<0.5$, we have $Q_{\infty}(\widehat{u}^{n+1/2})=0$ and $\proj_{[0,1]}(\widehat{u}^{n+1/2})<0.5$ since the relation among $\widehat{u}^{n+1/2},0$ and 1 is preserved by the projection. Because $\proj_{[0,1]}(\widehat{u}^{n+1/2})<0.5$, we have $Q_{\infty}\circ \proj_{[0,1]}(\widehat{u}^{n+1/2})=0=Q_{\infty}(\widehat{u}^{n+1/2})$. The same argument can be used to show $Q_{\infty}\circ \proj_{[0,1]}(\widehat{u}^{n+1/2})=1=Q_{\infty}(\widehat{u}^{n+1/2})$ when $\widehat{u}^{n+1/2}> 0.5$ and $Q_{\infty}\circ \proj_{[0,1]}(\widehat{u}^{n+1/2})=0.5=Q_{\infty}(\widehat{u}^{n+1/2})$ when $\widehat{u}^{n+1/2}= 0.5$.   Note that the projection operator can be realized by a two-layer ReLU network:
\begin{align*}
\proj_{[0,1]}(a)=\max\{\min\{a,1\},0\}= \ReLU(1-\ReLU(1-a)),
\end{align*}
which does not introduce extra complexity to the network.

The second method considered in this paper is to utilize the sigmoid function defined by:
$$
\Sig(a)=\frac{1}{1+\exp(-a)}.
$$
Compared to $\proj$ which is the composition of two ReLU functions, $\Sig$ is a more common activation function in the deep learning community.
We first compute $\Sig(\widehat{u}^{n+1/2}-0.5)$. Such an operation outputs a real number in $(0,1)$ and pointwisely preserves the relation between $|\widehat{u}^{n+1/2}-0|$ and $|\widehat{u}^{n+1/2}-1|$: if $|\widehat{u}^{n+1/2}-0|<|\widehat{u}^{n+1/2}-1|$, then $|\Sig(\widehat{u}^{n+1/2}-0.5)-0|<|\Sig(\widehat{u}^{n+1/2}-0.5)-1|$. In other words, if $\widehat{u}^{n+1/2}$ is closer to 0 than 1, then $\Sig(\widehat{u}^{n+1/2}-0.5)$ is closer to 0 than 1. Thus using $\Sig$ has the same two benefits as using $\proj_{[0,1]}$ discussed above.

Note that the formula of $\widehat{u}^{n+1/2}$ has a bias term $\tau b^n$. In the second method, we can combine the shifting scalar $0.5$ with the bias by shifting the bias by 0.5. Therefore, our updating formula for $\widehat{u}^{n+1}$ with $\Sig$ is
\begin{align}
\widehat{u}^{n+1}=Q_{\gamma}\circ \Sig(\widehat{u}^{n+1/2}).
\label{eq.activateQ} 
\end{align}
In our numerical experiments, the proposed methods with activation $Q_{\gamma}\circ \Sig$ work slightly better than that with $Q_{\gamma}\circ \proj_{[0,1]}$. In the rest of this section, we present our methods using $\Sig$ and activation function $Q_{\gamma}\circ \Sig$. We remark that $\Sig$ and $Q_{\gamma}\circ \Sig$ can be replaced by $\proj_{[0,1]}$ and $Q_{\gamma}\circ \proj_{[0,1]}$, respectively.

\begin{figure}[t!]
\centering
\begin{tabular}{c}
	(a)\\
	\includegraphics[width=0.8\textwidth]{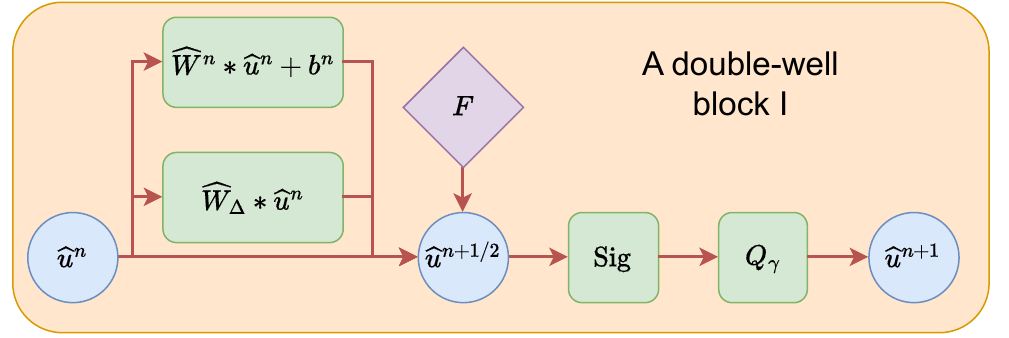}\\
	(b)\\
	\includegraphics[width=0.8\textwidth]{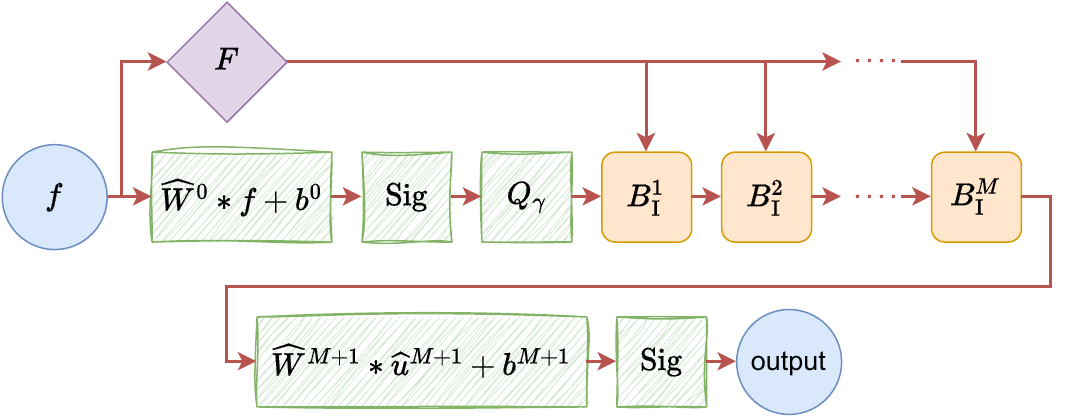}
\end{tabular}
\caption{For model (\ref{eq.control}): (a) An illustration of a double-well block I (DB-I) with activation $Q_{\gamma}\circ \Sig$. (b) An illustration of the double-well net I (DN-I) with activation $Q_{\gamma}\circ \Sig$. The architecture in (a) is the detailed representation of the block $B_{\rm I}^k$'s in (b). In both figures, the disk represents input, output and intermediate variables. The diamond represents the functional $F$. In (a), rectangles represent operations applied in a DB-I. In (b), sketched rectangles represent operations in the input layer and final layer, normal rectangles represent DB-I's. To better present the architecture, some scalar factors are omitted.}
\label{fig.block}
\end{figure}

We call the procedure $\widehat{u}^n\rightarrow \widehat{u}^{n+1/2} \rightarrow \widehat{u}^{n+1}$ a DB-I, denoted by $B^{n+1}_{\rm I}$, see Figure \ref{fig.block}(a) for an illustration with activation $Q_{\gamma}\circ \Sig$. The $B_{\rm I}^{n+1}$ contains trainable parameters $\widehat{W}^n,b^n$. The input for $B^n_{\rm I}$ includes the output of the previous block $u^n$, and $F(f)$.

In a DB-I, we have 
\begin{align}
\widehat{u}^{n+1/2}=\widetilde{W}^n*\widehat{u}^n+\widetilde{b}^n
\end{align}
with
\begin{align*}
\widetilde{W}^n=I_{\rm id}+\tau\lambda\epsilon \widehat{W}_{\Delta}+\tau \widehat{W}^n, \quad \widetilde{b}^n=\tau(b^n-F(f)),
\end{align*}
where $I_{\rm id}$ denotes the identity kernel so that $I_{\rm id}*\widehat{u}=\widehat{u}$ for any $\widehat{u}$. Thus a DB-I is a convolutional layer with weight kernel $\widetilde{W}^n$, a heavy bias term $\widetilde{b}^n$, and activation $Q_{\gamma}\circ \proj_{[0,1]}$ or $Q_{\gamma}\circ \Sig$.

\begin{figure}[t!]
\centering
\begin{tabular}{c}
	(a)\\
	\includegraphics[width=0.8\textwidth]{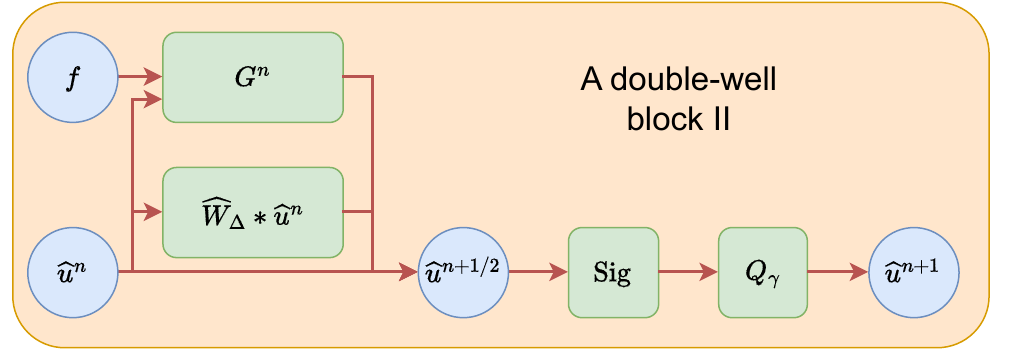}\\
	(b)\\
	\includegraphics[width=0.8\textwidth]{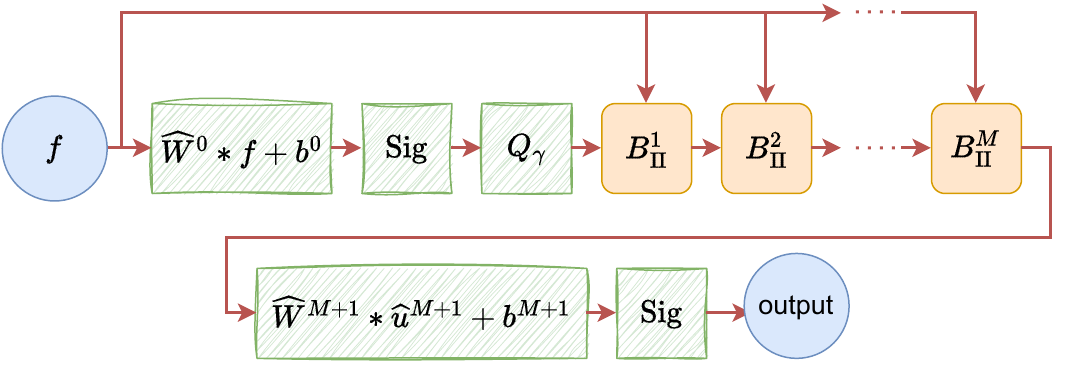}
\end{tabular}
\caption{For model (\ref{eq.control2}): (a) An illustration of a double-well block II (DB-II) with activation $Q_{\gamma}\circ \Sig$. (b) An illustration of the double-well net II (DN-II) with activation $Q_{\gamma}\circ \Sig$. The architecture in (a) is the detailed representation of the block $B_{\rm II}^k$'s in (b). In both figures, the disk represents input, output and intermediate variables. In (a), rectangles represent operations applied in a DB-II. In (b), sketched rectangles represent operations in the input layer and final layer, normal rectangles represent DB-II's. To better present the architecture, some scalar factors are omitted.}
\label{fig.block2}
\end{figure}

\textbf{Double-well net I.}
Given an image $f$, in order to use DB-I's to solve (\ref{eq.control}), we need an initial condition $u^0$. We propose to use a convolutional layer followed by a sigmoid function and fixed point iterations to generate $\widehat{u}^0$:
\begin{align}
\widehat{u}^0=H(f)=Q_{\gamma}\circ\Sig(\widehat{W}^0*f+b^0).
\label{eq.u0}
\end{align}
Since $\widehat{u}^n\in \mathbb{R}^{N_1\times N_2}$ for any $n$, we set $\widehat{W}^0$ as a convolution kernel with one output channel. The input image $f$ is also passed to $F$ to compute $F(f)$, where $F$ will be chosen as a UNet class demonstrated in Figure \ref{fig.UNetClass}. Then $\widehat{u}^0$ is passed through every DB-I sequentially and $F(f)$ is passed to all DB-I's. Assume there are $M$ blocks (DB-I's), and denote the $m$-th DB-I by $B^m_{\rm I}$. After $B^M_{\rm I}$, we add a convolution layer followed by a sigmoid function. Denote the kernel and bias in the last convolution layer by $\widehat{W}^{M}$ and $b^{M}$, respectively. The double-well net I (DN-I) is formulated as
\begin{align}
P_{\rm I}(f)=\Sig(\widehat{W}^{M}*(B^M_{\rm I}(\cdots B^2_{\rm I}( B^1_{\rm I}(\widehat{u}^0,F(f)),F(f))\cdots,F(f))+b^{M}),
\end{align}
where $u^0$ is defined in (\ref{eq.u0}). We illustrate the architecture of DN-I in Figure \ref{fig.block}(b). 

In general, DN-I is a convolutional neural network (CNN) with $M+1$ layers. The main differences between DN-I from classical CNN are: (1) DN-I uses a subnetwork $F(f)+b^n$ as the bias term in each layer, and the subnetwork $F(f)$ is the same across all layers, whereas existing CNNs only use $b^n$ as bias; (2) for the activation function, DN-I uses $Q_{\gamma}\circ\Sig$ or $Q_{\gamma}\circ \proj_{[0,1]}$, which are derived from the double-well potential, while classical CNN uses ReLU or $\Sig$, which are designed empirically; (3) unlike existing networks, DN-I has a mathematical background: it is an operator-splitting scheme that approximately solves the Potts model. 

PottsMGNet \cite{tai2023pottsmgnet} is another neural network derived from the Potts model and operator-splitting method which gives a clear explanation for encode-decoder type of neural networks like UNet. However, the mathematical treatments and splitting strategy of DN-I are different from those of PottsMGNet. For the mathematical treatment of the Potts model, to drive the segmentation function $u$ to be binary, a regularized softmax operator \cite{liu2022deep} is used in PottsMGNet, leading to the activation function being a sigmoid function. DN-I uses the double-well potential which leads to the activation function being a fixed-point iteration that minimize the potential. In the splitting strategy, inspired by the multigrid method, PottsMGNet uses control variables to represent the region force $F(f)$ at several scales. Thus $F(f)$ evolves over time and is linear in $f$ and $u$. In DN-I, $F(f)$ is assumed to be a complicated function of $f$. Inspired by the approximation theory of deep neural networks, it is approximated by a subnetwork. Such an approximation is fixed and independent of time. This subnetwork is used as a heavy bias in each layer of DN-I.

\subsection{Double-well net for model (\ref{eq.control2})}
For  $G(u,f,\bx,t)$ in Model II in  (\ref{eq.control2}) at each time step $t=t^n$, we use a UNet  specified by channel vector $\bc$  to represent  it,  see Section \ref{sec.experiments} for details.

\noindent\textbf{A Double-well block II.} In a Double-well Block II (DB-II), we first compute $\widehat{u}^{n+1/2}$ according to (\ref{eq.split2.1.formula}), and then apply (\ref{eq.activateQ.proj}) or (\ref{eq.activateQ}) to get $\widehat{u}^{n+1}$. A DB-II with activation $Q_{\gamma}\circ \Sig$ is visualized in Figure \ref{fig.block2}(a).

DB-II is different from DB-I in terms of how we treat the UNet class and its dependency on time. In DB-I, the UNet class is contained in the bias term and does not apply any convolution to $\widehat{u}^n$. This class is independent of $t$ in the whole network, i.e., all blocks share the same UNet class and weight parameters. In DB-II, the UNet class is an operator that is directly applied to $\widehat{u}^n$. In DB-II, the UNet class depends on $t$, and its weight parameters can be different from block to block.

\noindent\textbf{Double-well net II.} For double-well net II (DN-II), we use (\ref{eq.u0}) as an initial condition. Denote the $m$-th DB-II by $B^m_{\rm II}$ and assume we use $M$ blocks. After $B_{\rm II}^M$, we add a convolutional layer followed by a sigmoid function. The DN-II is formulated as
\begin{align}
P_{\rm II}(f)=\Sig(\widehat{W}^{M}*(B^M_{\rm II}(\cdots (B^2_{\rm II}\circ B^1_{\rm II}(\widehat{u}^0,f),f)\cdots, f)+b^{M}).
\end{align}
An illustration of DN-II is shown in Figure \ref{fig.block2}(b).

Compared to DN-I, DN-II is more like the classical CNNs, while it still has three differences from existing networks for image segmentation: (1) in DN-II, the input image $f$ is used at each block, instead of only at the beginning of the whole network; (2) For the activation function, DN-II uses $Q_{\gamma}\circ\Sig$ or $Q_{\gamma}\circ \proj_{[0,1]}$, which are derived from the double-well potential,  while classical CNN uses ReLU or $\Sig$, which are designed empirically; (3) unlike existing networks, DN-II has a mathematical background: it is an operator-splitting scheme that approximately solves the Potts model. Unlike DN-I, which fixes $F(f)$ over time and assumes $F(f)$ is only a function of the input image $f$, DN-II combines $F(f)$ and the control variables and represents the combination using a subnetwork which is applied to both $f$ and $u$ and whose parameters evolve over time.

\section{Numerical experiments} \label{sec.experiments}
In this section, we demonstrate the performance of the proposed models by systematic numerical experiments. Our PyTorch code is available at \url{https://github.com/liuhaozm/Double-well-Net}. We consider three datasets: (i) The MARA10K data set \cite{cheng2014global}, which contains 10000 images with pixel-level salient labeling. We resize the images to the size $192\times 256$ and select 2500 images for training and 400 images for testing. (ii) The Extended Complex Scene Saliency Dataset (ECSSD) \cite{shi2015hierarchical}. ECSSD is a semantic segmentation data set containing 1000 images with complex backgrounds and manually labeled masks. We resize all images to the size $192\times 256$, using 800 images for training and 200 images for testing. (iii) The Retinal Images vessel Tree Extraction dataset (RITE) \cite{hu2013automated}. RITE is a dataset for the segmentation and classification of arteries and veins on the retinal fundus. This dataset contains 40 images. We resize all images to the size $256\times 256$, use 20 images for training and 20 images for testing. We train each model with Adam optimizer \cite{kingma2014adam} and 400 epochs for MARA10K, and 600 epochs for ECSSD and RITE. Without further specification, for model DN-I, we use the UNet class with channels $\bc=[128,128,128,128,256]$, and 10 blocks. We set $\tau=0.2$. For DN-II, we use the UNet class with channels $\bc=[64,64,64,128,128]$ for each block and 3 blocks. We set $\tau=0.5$. In both models, without specification, we set $\gamma=3,\lambda\varepsilon=1$, and $\alpha=2\tau\lambda/\varepsilon=15$. We use (\ref{eq.u0}) to compute $u^0$, where $W^0$ and $b^0$ are to be learned from data. In all experiments, we use DN-I and DN-II to denote the networks with activation $Q_{\gamma}\circ \Sig$, and use DN-I Proj and DN-II Proj to denote the networks with activation $Q_{\gamma}\circ \proj_{[0,1]}$.

We compare the proposed models with state-of-the-art image segmentation networks, including UNet \cite{ronneberger2015u}, UNet++ \cite{zhou2018unet++}, MANet \cite{fan2020ma}, and DeepLabV3+ \cite{chen2018encoder}. In our experiments, UNet++, MANet and DeepLabV3+ are implemented using the Segmentation Models PyTorch package \cite{Yakubovskiy2019}. 
For these models, the final layer is a sigmoid function. Thus, the output is a matrix with elements between 0 and 1. To obtain a binary segmentation result, we use a simple threshold $T$  as
\begin{align}
T\circ P(f)=\begin{cases}
	1 & \mbox{ if } P(f)\geq 0.5,\\
	0 & \mbox{ if } P(f)<0.5.
\end{cases}
\end{align}

To measure the similarity between a model prediction and the given mask, we use accuracy and dice score. Given a set of test images $\{f_k\}_{k=1}^{K}$ of size $N_1\times N_2$ and their segmentation masks $\{v_k\}_{k=1}^{K}$ , for model $P$, we define the accuracy as
\begin{align}
\mbox{accuracy}=\frac{1}{K}\sum_{k=1}^{K}\left[ \frac{\left|[T\circ P(f_k)]\cap v_k\right|}{N_1N_2}\times 100\% \right] 
\end{align}
and the dice score as
\begin{align}
\mbox{dice}=\frac{1}{K}\sum_{k=1}^{K}\left[ \frac{2\left|[T\circ P(f_k)]\cap v_k\right|}{|T\circ P(f)|+ |v|} \right],
\end{align}
where $|v|$ denotes the number of nonzero elements of a binary function $v$, and $\cap$ is the logic `and' operation. 

\subsection{Comparison with other networks}
\label{sec.compare}

We first consider MARA10K. The comparison of accuracy and dice score between the proposed models and other networks are presented in Table \ref{tab.compare}. In terms of performance, both activations $Q_{\gamma}\circ \Sig$ and $Q_{\gamma}\circ \proj_{[0,1]}$ work similarly for DN-I and DN-II. Compared to UNet, the accuracy of the proposed models is 1\% higher. Compared to other networks, the proposed models also provide higher accuracy and dice score. 

\begin{figure}[t!]
\centering
\begin{tabular}{ccccc}
	Image &\raisebox{-.5\height}{\includegraphics[width=0.16\textwidth]{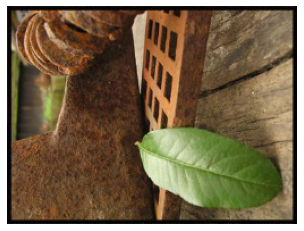}} &
	\raisebox{-.5\height}{\includegraphics[width=0.16\textwidth]{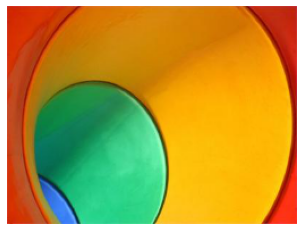} }&
	\raisebox{-.5\height}{\includegraphics[width=0.16\textwidth]{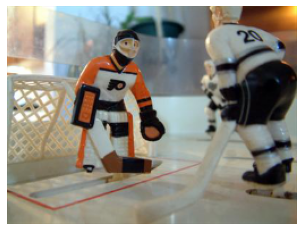}} &
	\raisebox{-.5\height}{\includegraphics[width=0.16\textwidth]{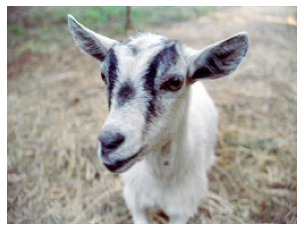}}\\
	\makecell{Ground\\truth} &\raisebox{-.5\height}{\includegraphics[width=0.16\textwidth]{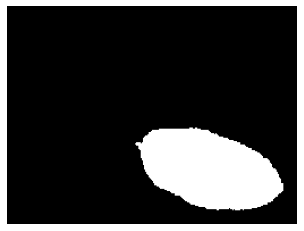}} &
	\raisebox{-.5\height}{\includegraphics[width=0.16\textwidth]{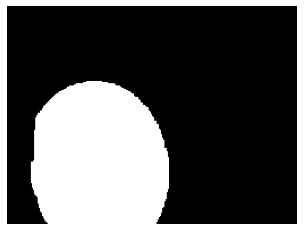}} & 
	\raisebox{-.5\height}{\includegraphics[width=0.16\textwidth]{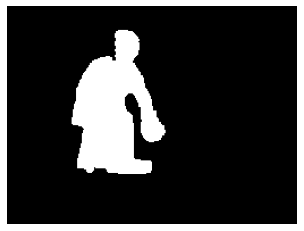}} &
	\raisebox{-.5\height}{\includegraphics[width=0.16\textwidth]{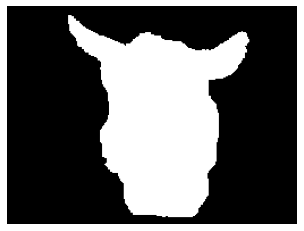}}\\
	DN-I &\raisebox{-.5\height}{\includegraphics[width=0.16\textwidth]{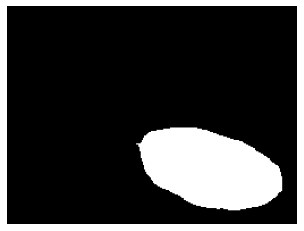}} & 
	\raisebox{-.5\height}{\includegraphics[width=0.16\textwidth]{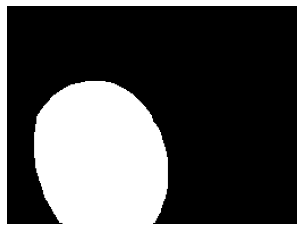}} &
	\raisebox{-.5\height}{\includegraphics[width=0.16\textwidth]{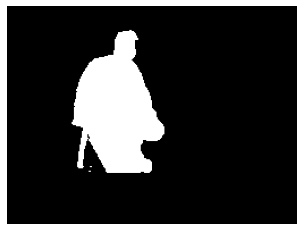}} &
	\raisebox{-.5\height}{\includegraphics[width=0.16\textwidth]{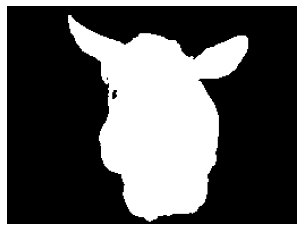}}\\
	DN-II &\raisebox{-.5\height}{\includegraphics[width=0.16\textwidth]{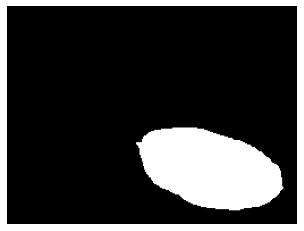}} &
	\raisebox{-.5\height}{\includegraphics[width=0.16\textwidth]{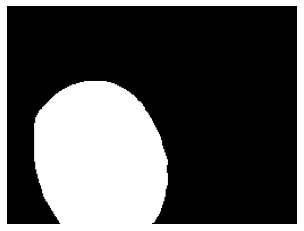}} &
	\raisebox{-.5\height}{\includegraphics[width=0.16\textwidth]{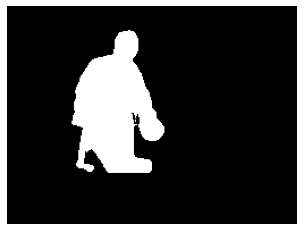}} &
	\raisebox{-.5\height}{\includegraphics[width=0.16\textwidth]{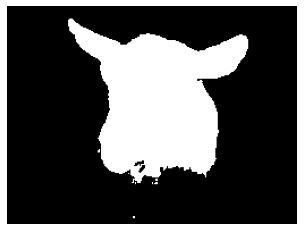}}\\
	UNet &\raisebox{-.5\height}{\includegraphics[width=0.16\textwidth]{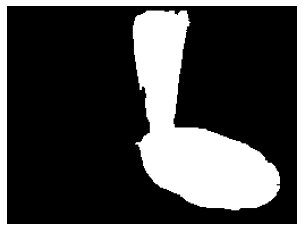}} &
	\raisebox{-.5\height}{\includegraphics[width=0.16\textwidth]{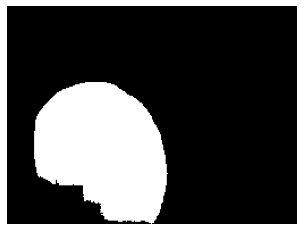}} &
	\raisebox{-.5\height}{\includegraphics[width=0.16\textwidth]{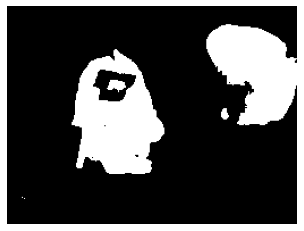}} &
	\raisebox{-.5\height}{\includegraphics[width=0.16\textwidth]{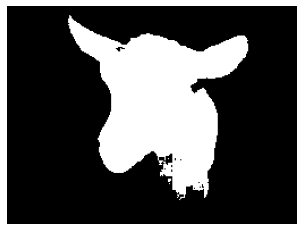}}\\
	UNet++ &\raisebox{-.5\height}{\includegraphics[width=0.16\textwidth]{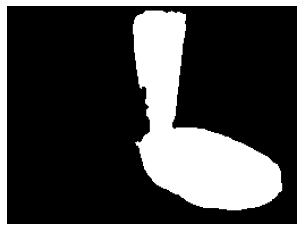}} &
	\raisebox{-.5\height}{\includegraphics[width=0.16\textwidth]{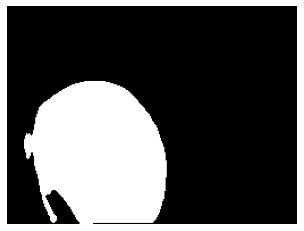}} &
	\raisebox{-.5\height}{\includegraphics[width=0.16\textwidth]{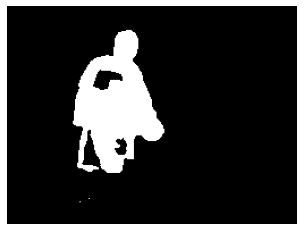}} &
	\raisebox{-.5\height}{\includegraphics[width=0.16\textwidth]{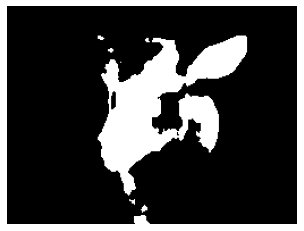}}\\
	MANet & \raisebox{-.5\height}{\includegraphics[width=0.16\textwidth]{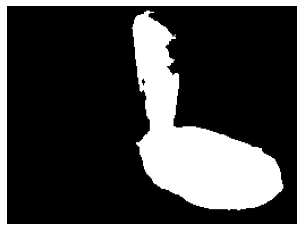}} &
	\raisebox{-.5\height}{\includegraphics[width=0.16\textwidth]{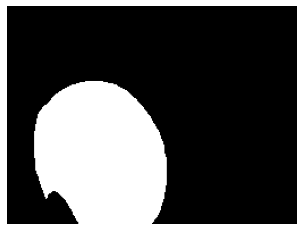}} &
	\raisebox{-.5\height}{\includegraphics[width=0.16\textwidth]{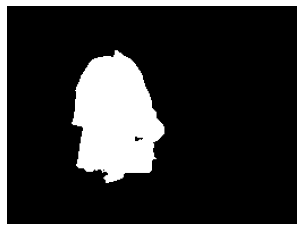}} &
	\raisebox{-.5\height}{\includegraphics[width=0.16\textwidth]{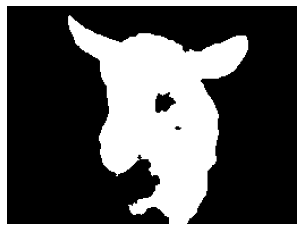}}\\
	DeepLabV3+ &\raisebox{-.5\height}{\includegraphics[width=0.16\textwidth]{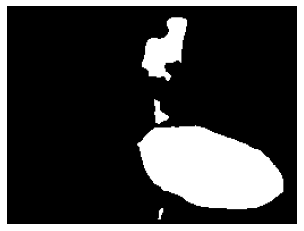}} &
	\raisebox{-.5\height}{\includegraphics[width=0.16\textwidth]{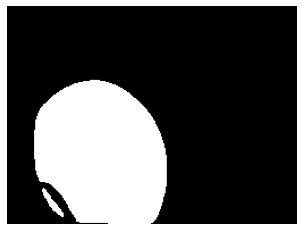}} &
	\raisebox{-.5\height}{\includegraphics[width=0.16\textwidth]{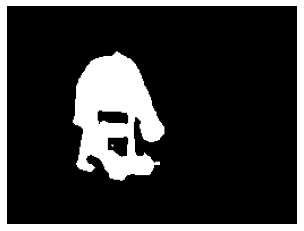}}&
	\raisebox{-.5\height}{\includegraphics[width=0.16\textwidth]{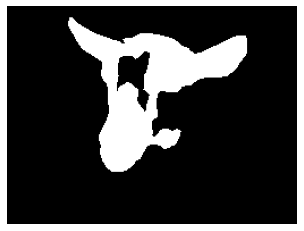}}\\
\end{tabular}
\caption{Segmentation examples in the comparison of DN-I and DN-II with UNet, UNet++, MANet, and DeepLabV3+ on MARA10K. }
\label{fig.compare}
\end{figure}

We present some selected segmentation results in Figure \ref{fig.compare}. The selected results are not the best or worst ones. There are many test images with similar results. These images are selected to show the visual differences between the proposed methods and existing networks. In these examples, the proposed models accurately segment the given images. The predicted results are very similar to the given masks, while the other models more or less mistakenly segment some objects or miss some parts. 

To measure the complexity of different models, we also present the number of parameters of each model in Table \ref{tab.compare}. The proposed models have the least number of parameters compared to other models. The number of parameters of the proposed models is less than one-third of that of UNet and MANet, and less than half of that of UNet++ and DeepLabV3+.

\begin{figure}[t!]
\centering
\begin{tabular}{ccc}
	(a) & (b) & (c)\\
	\includegraphics[trim={0.2cm 0 0.2cm 0},clip,height=0.295\textwidth]{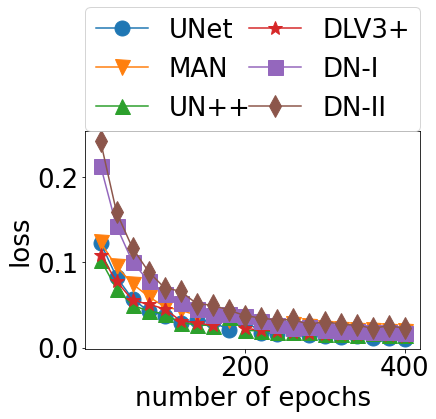}&
	\includegraphics[trim={0.2cm 0 0.2cm 0},clip,height=0.295\textwidth]{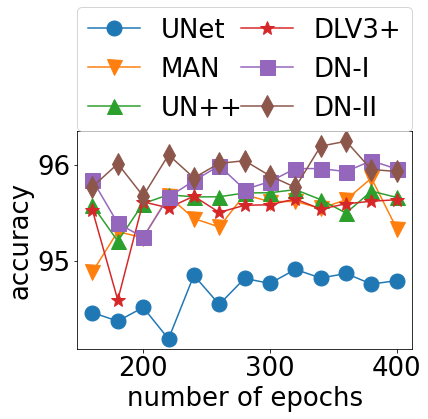}&
	\includegraphics[trim={0.2cm 0 0.2cm 0},clip,height=0.295\textwidth]{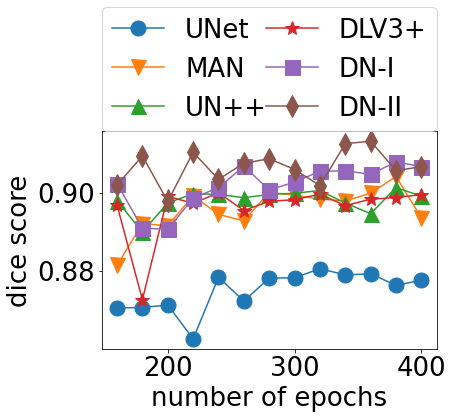}
\end{tabular}
\caption{Comparison of the histories of (a) training loss, (b) accuracy and (c) dice score of DN-I and DN-II with UNet, UNet++ (UN++), MANet (MAN), and DeepLabV3+ (DLV3+) on MARA10K. 
}
\label{fig.loss}
\end{figure}

We present the evolution of loss, accuracy, and dice score of all models in Figure \ref{fig.loss}. They are plotted every 20 epochs. In Figure \ref{fig.loss}(a), we observe that the proposed models can reduce the loss as effectively as other models. Figures  \ref{fig.loss}(b) and (c) show that for most of the time during training, the proposed models consistently give the highest accuracy and dice score.

\begin{table}[t!]
\centering
\begin{tabular}{c||c|c|c}
	\hline
	& Accuracy & Dice score & Number of Parameters\\
	\hline
	DN-I & {\bf 95.95}\% & {\bf 0.9067} &9.86$\times 10^{6}$\\
	\hline
	DN-I Proj & {\bf 95.95}\% &  0.9059 &9.86$\times 10^{6}$\\
	\hline
	DN-II & 95.93\% & {\bf 0.9067} &{\bf 9.21$\times 10^{6}$}\\
	\hline
	DN-II Proj& 95.68\% & 0.9003 &{\bf 9.21$\times 10^{6}$}\\
	\hline
	UNet & 94.74\% & 0.8758 & 31.04$\times 10^{6}$\\
	\hline
	UNet++ & 95.66\% & 0.8992 & 26.08$\times 10^{6}$\\
	\hline
	MANet & 95.33\% & 0.8936 & 31.78$\times 10^{6}$\\
	\hline
	DeepLabV3+ & 95.41\% & 0.8937 & 22.44$\times 10^{6}$\\
	\hline
\end{tabular}
\caption{Comparison of the accuracy, dice score, and number of parameters of DN-I and DN-II with UNet, UNet++, MANet, and DeepLabV3+ on MARA10K. Here DN-I and DN-II denote the proposed networks with activation $Q_{\gamma}\circ \Sig$, and DN-I Proj and DN-II Proj denote the proposed networks with activation $Q_{\gamma}\circ \proj_{[0,1]}$.}
\label{tab.compare}
\end{table}

\begin{figure}[t!]
\centering
\begin{tabular}{ccccc}
	Image &\raisebox{-.5\height}{\includegraphics[width=0.17\textwidth]{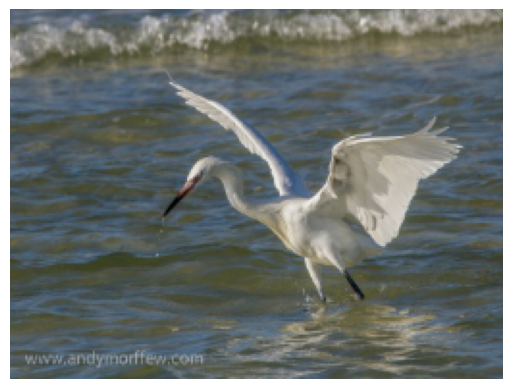}} &
	\raisebox{-.5\height}{\includegraphics[width=0.17\textwidth]{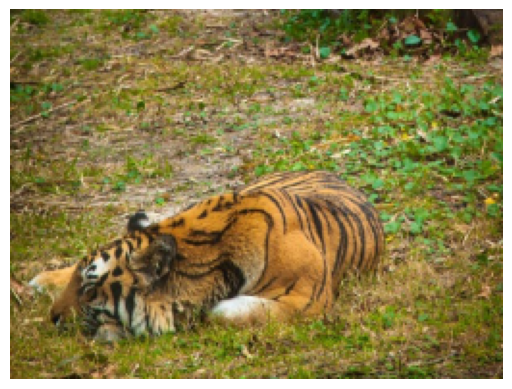} }&
	\raisebox{-.5\height}{\includegraphics[width=0.15\textwidth]{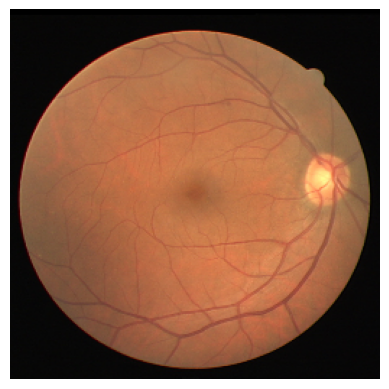}} &
	\raisebox{-.5\height}{\includegraphics[width=0.15\textwidth]{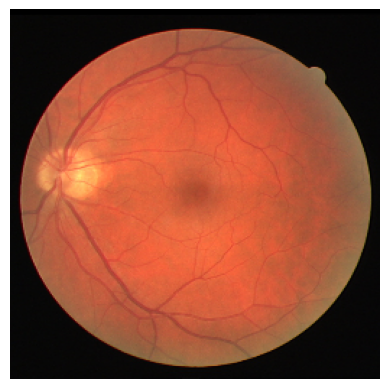}}\\
	\makecell{Ground\\truth}  &\raisebox{-.5\height}{\includegraphics[width=0.17\textwidth]{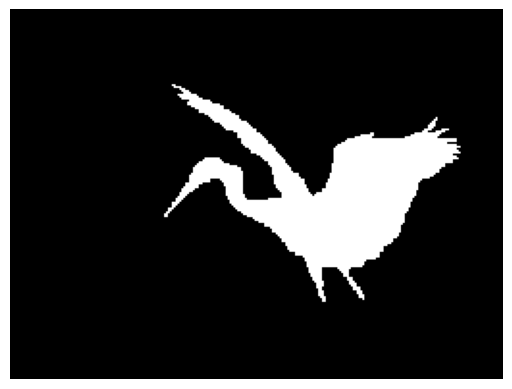}} &
	\raisebox{-.5\height}{\includegraphics[width=0.17\textwidth]{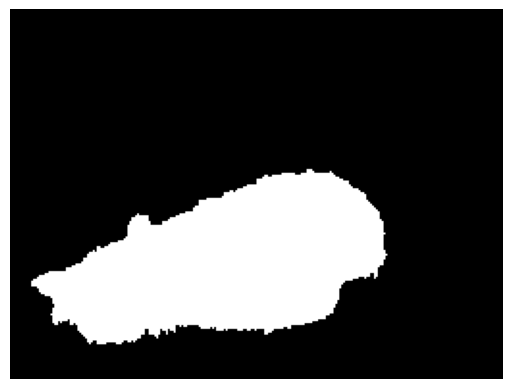}} & 
	\raisebox{-.5\height}{\includegraphics[width=0.15\textwidth]{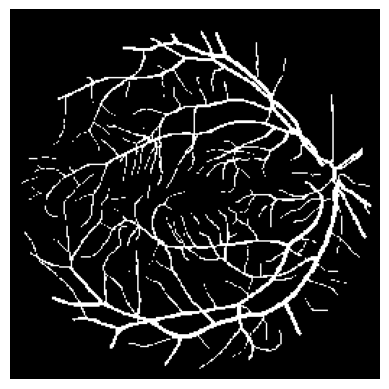}} &
	\raisebox{-.5\height}{\includegraphics[width=0.15\textwidth]{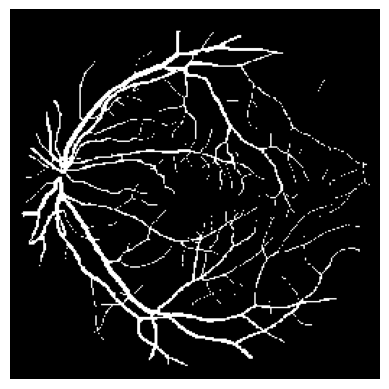}}\\
	DN-I &\raisebox{-.5\height}{\includegraphics[width=0.17\textwidth]{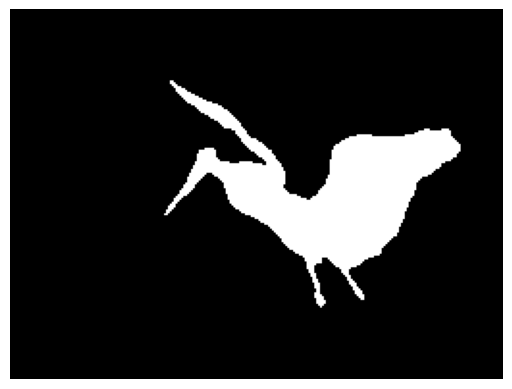}} & 
	\raisebox{-.5\height}{\includegraphics[width=0.17\textwidth]{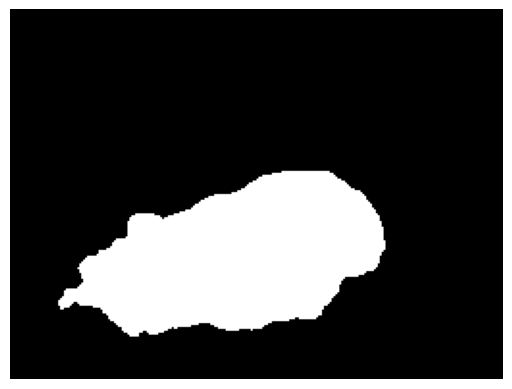}} &
	\raisebox{-.5\height}{\includegraphics[width=0.15\textwidth]{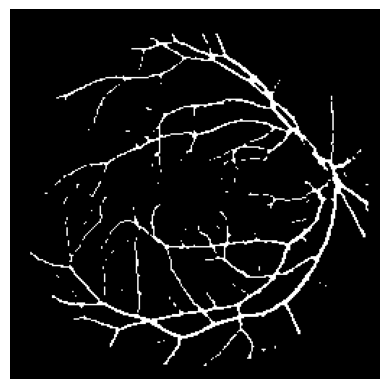}} &
	\raisebox{-.5\height}{\includegraphics[width=0.15\textwidth]{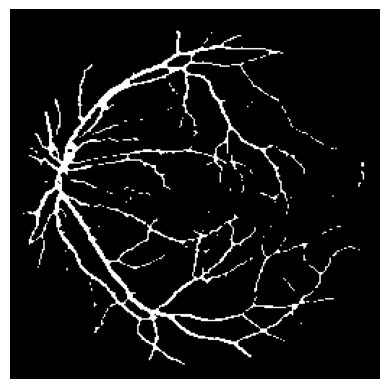}}\\
	DN-II &\raisebox{-.5\height}{\includegraphics[width=0.17\textwidth]{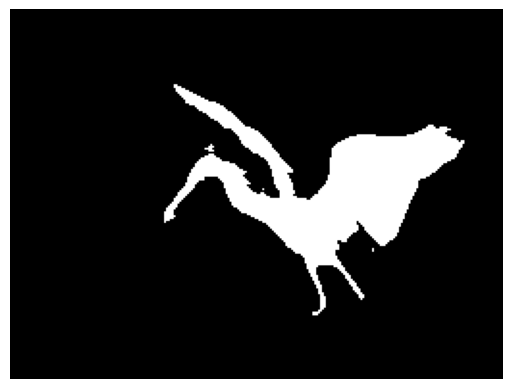}} &
	\raisebox{-.5\height}{\includegraphics[width=0.17\textwidth]{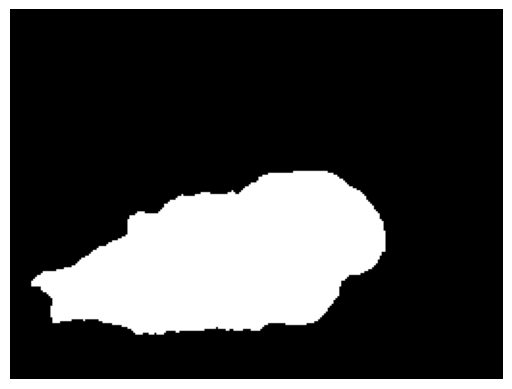}} &
	\raisebox{-.5\height}{\includegraphics[width=0.15\textwidth]{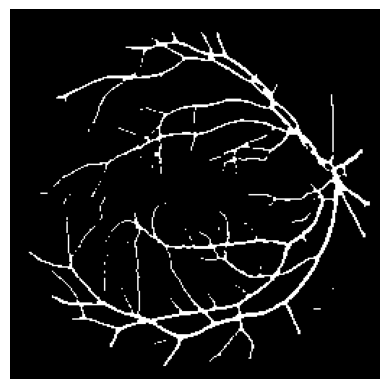}} &
	\raisebox{-.5\height}{\includegraphics[width=0.15\textwidth]{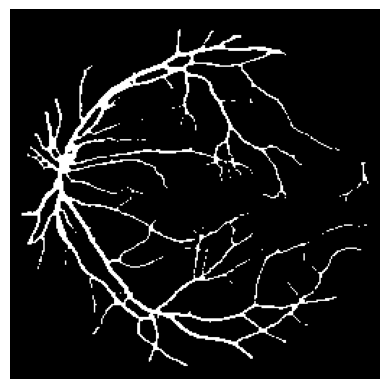}}\\
	UNet &\raisebox{-.5\height}{\includegraphics[width=0.17\textwidth]{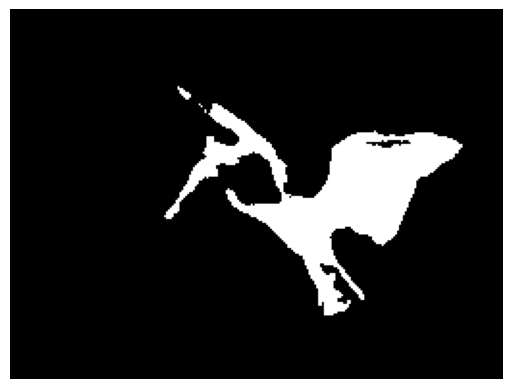}} &
	\raisebox{-.5\height}{\includegraphics[width=0.17\textwidth]{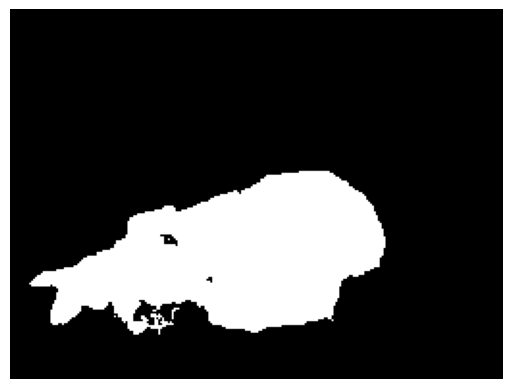}} &
	\raisebox{-.5\height}{\includegraphics[width=0.15\textwidth]{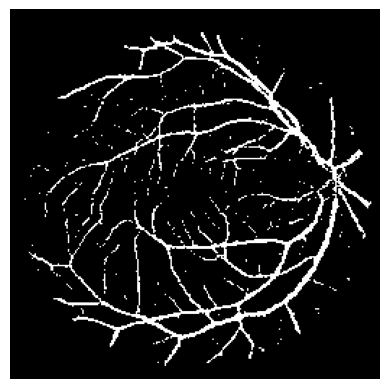}} &
	\raisebox{-.5\height}{\includegraphics[width=0.15\textwidth]{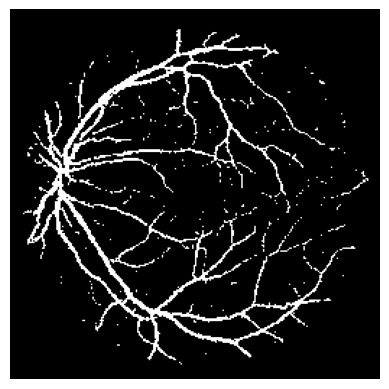}}\\
	UNet++ &\raisebox{-.5\height}{\includegraphics[width=0.17\textwidth]{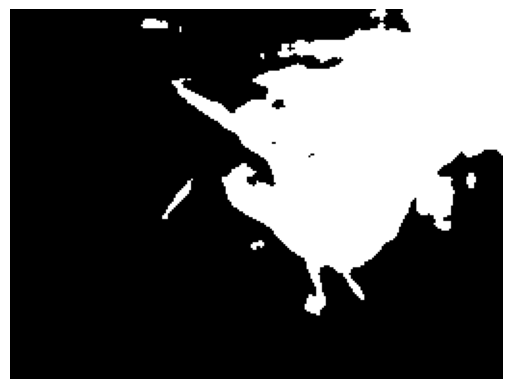}} &
	\raisebox{-.5\height}{\includegraphics[width=0.17\textwidth]{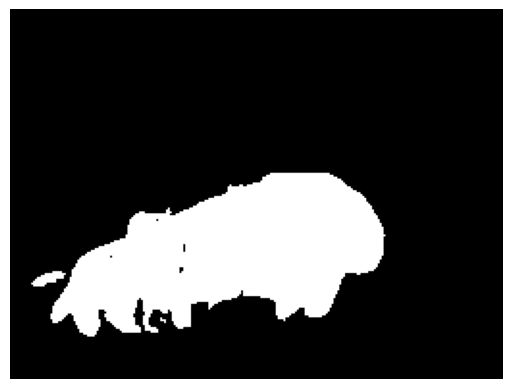}} &
	\raisebox{-.5\height}{\includegraphics[width=0.15\textwidth]{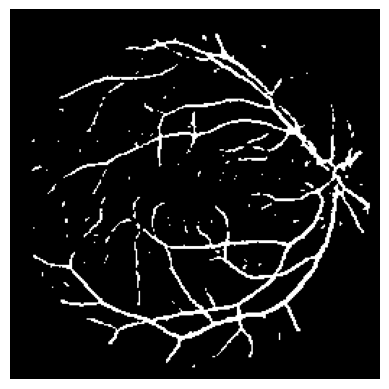}} &
	\raisebox{-.5\height}{\includegraphics[width=0.15\textwidth]{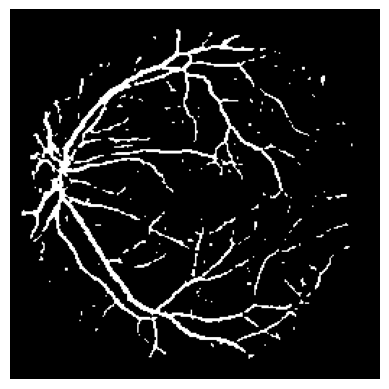}}\\
	MANet & \raisebox{-.5\height}{\includegraphics[width=0.17\textwidth]{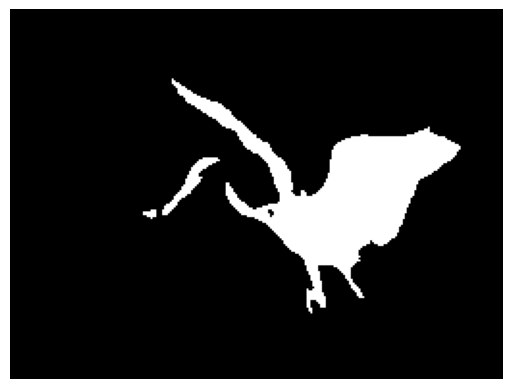}} &
	\raisebox{-.5\height}{\includegraphics[width=0.17\textwidth]{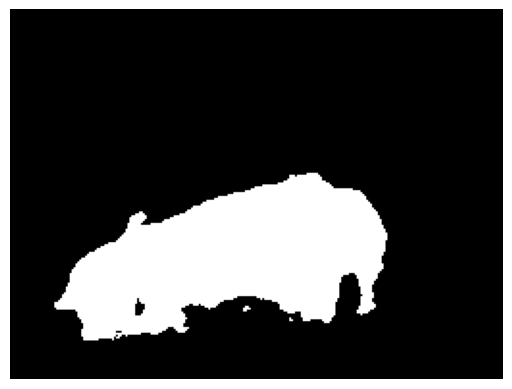}} &
	\raisebox{-.5\height}{\includegraphics[width=0.15\textwidth]{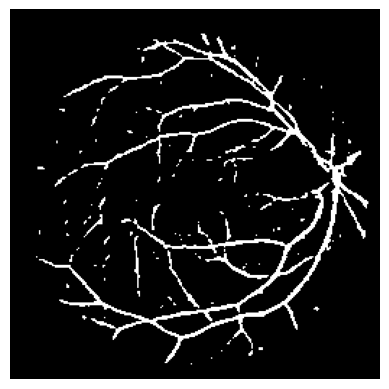}} &
	\raisebox{-.5\height}{\includegraphics[width=0.15\textwidth]{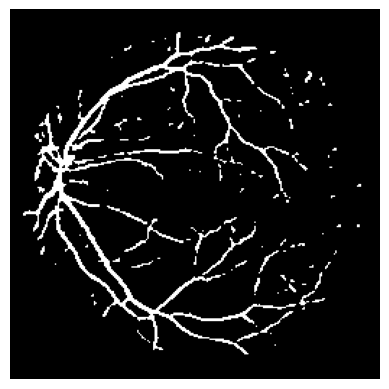}}\\
	DeepLabV3+ &\raisebox{-.5\height}{\includegraphics[width=0.17\textwidth]{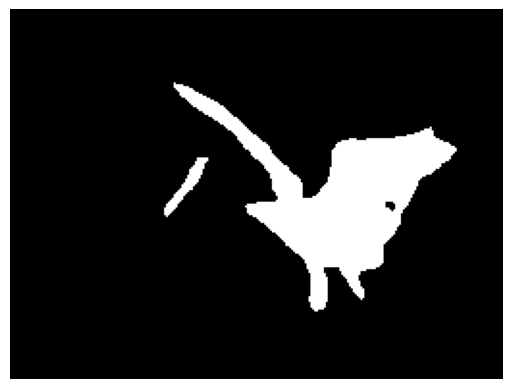}} &
	\raisebox{-.5\height}{\includegraphics[width=0.17\textwidth]{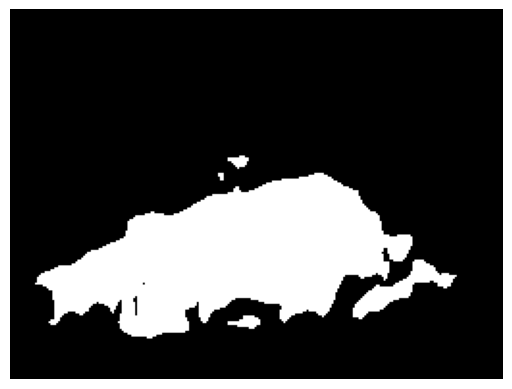}} &
	\raisebox{-.5\height}{\includegraphics[width=0.15\textwidth]{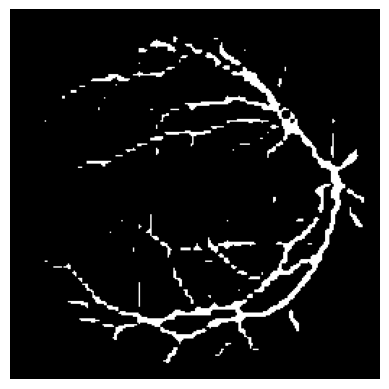}}&
	\raisebox{-.5\height}{\includegraphics[width=0.15\textwidth]{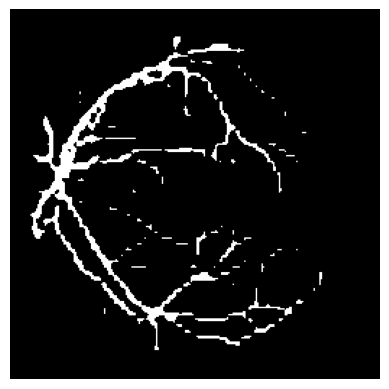}}\\
\end{tabular}
\caption{Comparison between the proposed models and UNet, UNet++, MANet, and DeepLabV3+ on ECSSD (the first two columns) and RITE (the last two columns). }
\label{fig.compare.more}
\end{figure}

We next consider ECSSD and RITE. The comparisons of DN-I and DN-II with other networks are shown in Table \ref{tab.compare.more}. For RITE, because the targets are vessels that are very thin, a smaller weight for the length penalty term should be used. We set $\lambda\varepsilon=0.25$ for both DN-I and DN-II, $\lambda\varepsilon=0.025$ for DN-I Proj, and $\lambda\varepsilon=0.05$ for DN-II Proj. For ECSSD and RITE, both DN-I and DN-II outperform other networks, with DN-I Proj providing the highest accuracy and dice score for ECSSD and DN-II providing the highest accuracy and dice score for RITE.  The comparison of some selected segmentation results is shown in Figure \ref{fig.compare.more}. For ECSSD, DN-I and DN-II can successfully segment the target with smooth boundaries, while other networks either fail to segment the target or have noise in the results. For RITE, DN-I and DN-II provide clear main branches of vessels, while results by other networks contain scattered noise.

\begin{table}[t!]
\centering
\begin{tabular}{c||c|c||c|c}
	\hline
	& \multicolumn{2}{c||}{ECSSD} & \multicolumn{2}{c}{RITE}\\
	\hline
	& Accuracy & Dice Score & Accuracy & Dice Score\\
	\hline
	DN-I &  91.33\% &  0.8063 & 95.78\% & 0.7174\\
	\hline
	DN-I Proj& {\bf 91.47}\% & {\bf 0.8084} & 95.06\% & 0.6809\\
	\hline
	DN-II & 91.17\% & 0.8056 & {\bf 96.06}\%&{\bf 0.7368} \\
	\hline
	DN-II Proj& 90.74\% & 0.7945 &  95.89\%& 0.7145 \\
	\hline
	UNet & 90.94\% &0.7941 & 95.00\% &0.6760\\
	\hline
	UNet++ & 88.00\% & 0.7305 & 95.00\% & 0.6734\\
	\hline
	MANet & 89.40\% & 0.7575 & 94.94\% & 0.6602\\
	\hline
	DeepLabV3+ & 89.17\% & 0.7571 & 93.19\% & 0.5209\\
	\hline
\end{tabular}

\caption{Comparison of the accuracy and dice score of DN-I and DN-II with UNet, UNet++, MANet, and DeepLabV3+ on ECSSD and RITE. Here DN-I and DN-II denote the proposed networks with activation $Q_{\gamma}\circ \Sig$, and DN-I Proj and DN-II Proj denote the proposed networks with activation $Q_{\gamma}\circ \proj_{[0,1]}$.}
\label{tab.compare.more}
\end{table}

\subsection{Effects of hyperparameters of DN-I}
We explore the effects of hyperparameters of DN-I using MARA10K.
Except for the weight parameters in the neural network, DN-I has several hyperparameters: the number of blocks $M$, the network architecture in each block which is specified by $\bc$, the time step $\tau$, the coefficient of diffusion $\lambda$, and the number of iterations of the fixed point iteration $\gamma$. In this subsection, we fix the network architecture as $\bc = [128,128,128,128,256]$ and explore the effects of other hyperparameters. For better visualization of the comparison, we present results from the 200th to 400th epoch. The accuracy, dice score, and loss are plotted every 20 epochs.

\begin{figure}[t!]
\centering
\begin{tabular}{ccc}
	(a) & (b) & (c)\\
	\includegraphics[trim={0.2cm 0 0.2cm 0},clip,width=0.3\textwidth]{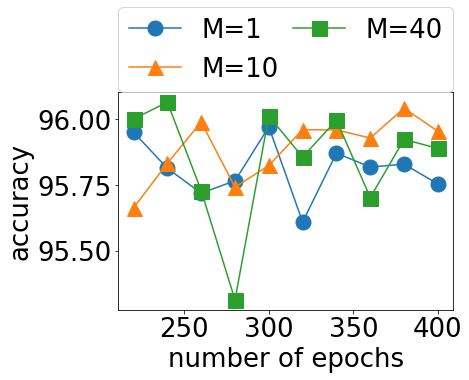}&
	\includegraphics[trim={0.2cm 0 0.2cm 0},clip,width=0.3\textwidth]{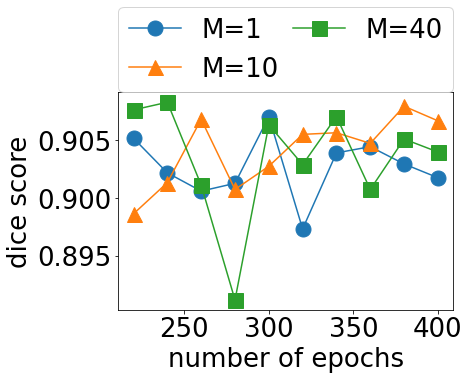}&
	\includegraphics[trim={0.2cm 0 0.2cm 0},clip,width=0.3\textwidth]{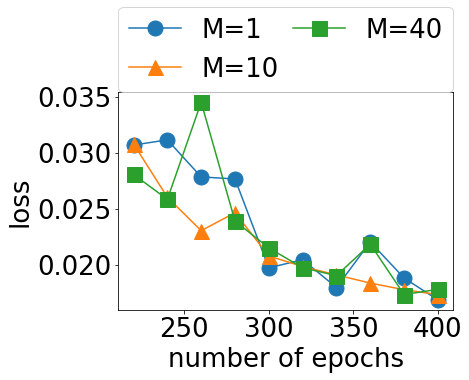}
\end{tabular}
\caption{Effect of the number of blocks $M$ for DN-I.}
\label{fig.simpleU.blk}
\end{figure}

For the number of blocks, we test $M=1,10,40$ and fix $\tau=0.2, \lambda\varepsilon=1,\gamma=3$. For a fixed $F$, the number of blocks corresponds to the number of iterations (time stepping) in the operator-splitting algorithm. The comparisons of the accuracy and dice score are shown in Figure \ref{fig.simpleU.blk}. We observe that DN-I with $M=10$ provides the best results. Note that adding additional blocks only slightly increases the number of parameters of DN-I (the number of parameters in the models in Figure \ref{fig.simpleU.blk} are all around 9.86$\times 10^{6}$), but significantly increases the model complexity. When $M$ is too small, the model is too simple and cannot accomplish the segmentation task well. But a too-large $M$ makes the model too complicated so that it can easily get stuck at a local minimizer during training. This test implies that as long as $F$ is sufficiently good, a few operator-splitting iterations are enough to accomplish the segmentation task. In other words, we can find an effective $F$ so that a few operator-splitting method iterations give desired segmentation.

\begin{figure}[t!]
\centering
\begin{tabular}{ccc}
	(a) & (b) & (c)\\
	\includegraphics[trim={0.2cm 0 0.2cm 0},clip,width=0.3\textwidth]{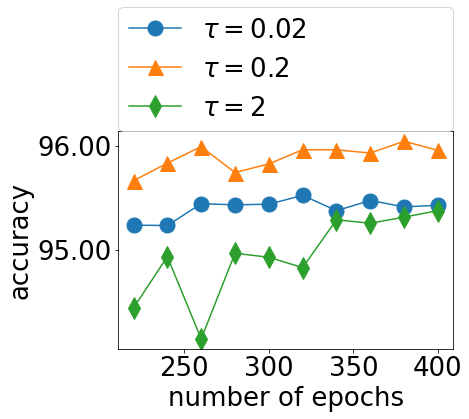}&
	\includegraphics[trim={0.2cm 0 0.2cm 0},clip,width=0.3\textwidth]{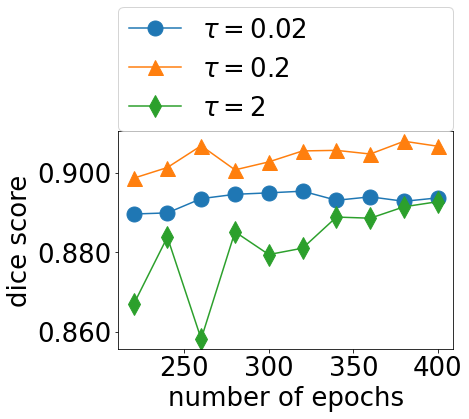}&
	\includegraphics[trim={0.2cm 0 0.2cm 0},clip,width=0.3\textwidth]{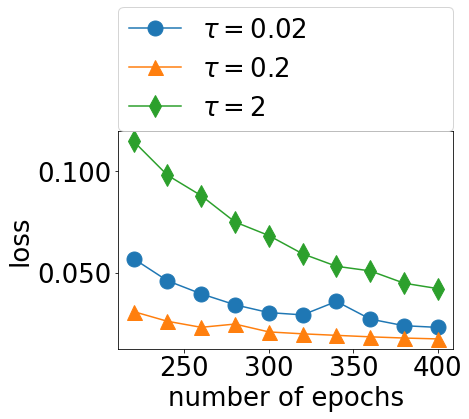}
\end{tabular}
\caption{Effect of the time step $\tau$ for DN-I.}
\label{fig.simpleU.time}
\end{figure}

We then set $M=10,\lambda\varepsilon=1,\gamma=3$, and test $\tau=0.02,0.2$ and $2$. The results are shown in Figure \ref{fig.simpleU.time}. In our model, $\tau$ corresponds to the time step in the operator-splitting algorithm. In numerical methods, small $\tau$ makes the solution evolve very slowly, and large $\tau$ may make the algorithm unstable. This phenomenon is observed in Figure \ref{fig.simpleU.time}. Too small or too large $\tau$ leads to slower decay of the loss and less accurate results.

\begin{figure}[t!]
\centering
\begin{tabular}{ccc}
	(a) & (b) & (c)\\
	\includegraphics[trim={0.2cm 0 0.2cm 0},clip,width=0.3\textwidth]{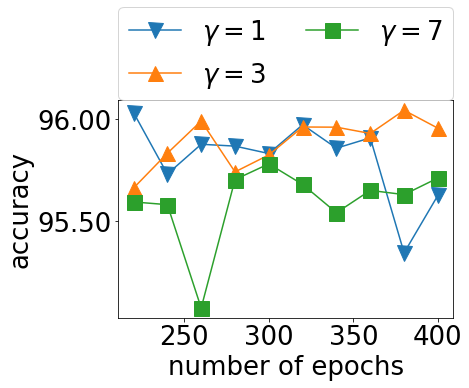}&
	\includegraphics[trim={0.2cm 0 0.2cm 0},clip,width=0.3\textwidth]{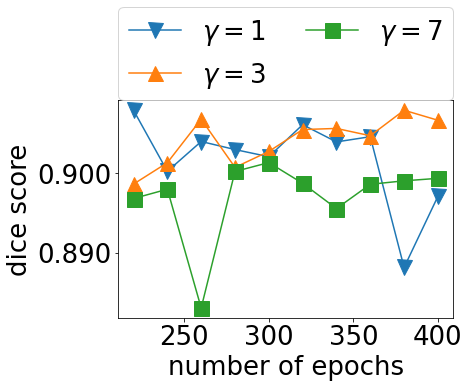}&
	\includegraphics[trim={0.2cm 0 0.2cm 0},clip,width=0.3\textwidth]{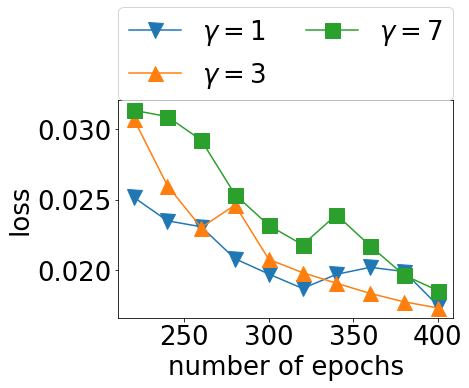}
\end{tabular}
\caption{Effect of the number of fixed point iteration $\gamma$ for DN-I.}
\label{fig.simpleU.iter}
\end{figure}

The next test is for the number of fixed-point iterations. We fix $M=10,\tau=0.2, \lambda\varepsilon=1$ and test $\gamma=0,3,7$. The results are shown in Figure \ref{fig.simpleU.iter}. The number of fixed point iterations determines how well the result after each iteration approximates the binary function, i.e., the function value is either 0 or 1. Larger $\gamma$ gives a better approximation and thus a better approximation of the Potts model. In Figure \ref{fig.simpleU.iter}, too small $\gamma$ makes the results worse, validating our argument. When $\gamma$ is too large, like 7, the result also gets worse. That is because each fixed point iteration is not a simple function. If $\gamma$ is too large, the operator, in general, gets very complicated, i.e., the loss functional has more local minimizers, making training the network more difficult.

\begin{figure}[t!]
\centering
\begin{tabular}{ccc}
	(a) & (b) & (c)\\
	\includegraphics[trim={0.2cm 0 0.2cm 0},clip,width=0.3\textwidth]{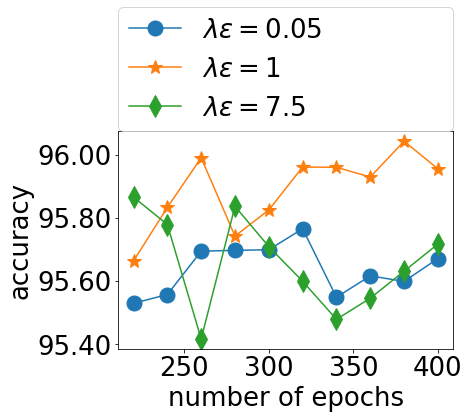}&
	\includegraphics[trim={0.2cm 0 0.2cm 0},clip,width=0.3\textwidth]{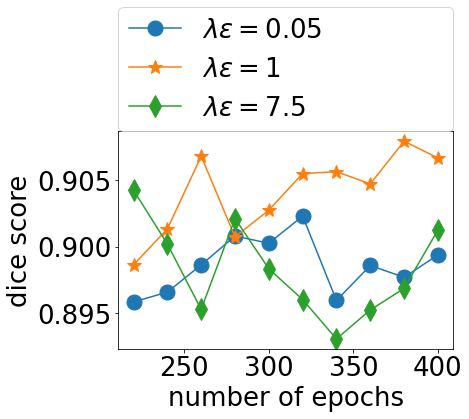}&
	\includegraphics[trim={0.2cm 0 0.2cm 0},clip,width=0.3\textwidth]{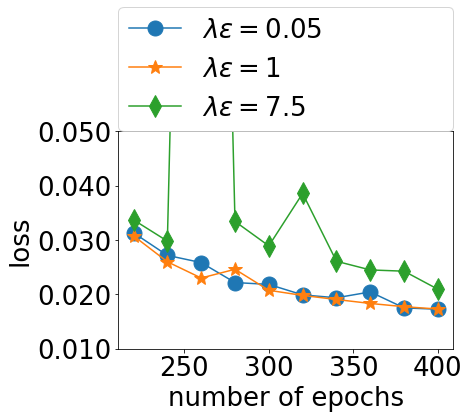}
\end{tabular}
\caption{Effect of the diffusion coefficient $\lambda$ for DN-I. For the case $\lambda\varepsilon=7.5$ (the green curve with diamond) in (c), the loss has a large value of 0.2852 at the 260-th epoch. To better visualize the evolution of loss, the vertical axis is truncated to $[0.01,0.05]$, making that point out of range and a sharp increase and decrease in the plot.}
\label{fig.simpleU.diff}
\end{figure}

In the last test, we study the effect of the diffusion coefficient $\lambda$. In this test, we set $M=10,\tau=0.2,\gamma=3$ and test %$\kappa=0.01,0.2$
$\lambda\varepsilon=0.05,1$ and 7.5. The results are shown in Figure \ref{fig.simpleU.diff}. Although the diffusion operator is a linear operator which can be absorbed by $W^n$ in each iteration, Figure \ref{fig.simpleU.diff} suggests that explicitly adding this diffusion operator with a proper coefficient improves the result.

\subsection{Effects of hyperparameters of DN-II}
Similar to DN-I, DN-II has hyperparameters: the number of blocks $M$, the network architecture in each block which is specified by $\bc$, the time step $\tau$, the coefficient of diffusion $\lambda$, and the number of iterations of the fixed point iteration $\gamma$. We fix the network architecture as $\bc = [64,64,64,128,128]$ and explore the effects of other hyperparameters. In this subsection, for better visualization of the comparisons, we present results from the 200th to 400th epoch. 
The accuracy, dice score, and loss are plotted every 20 epochs.

\begin{figure}[t!]
\centering
\begin{tabular}{ccc}
	(a) & (b) & (c)\\
	\includegraphics[trim={0.2cm 0 0.2cm 0},clip,width=0.3\textwidth]{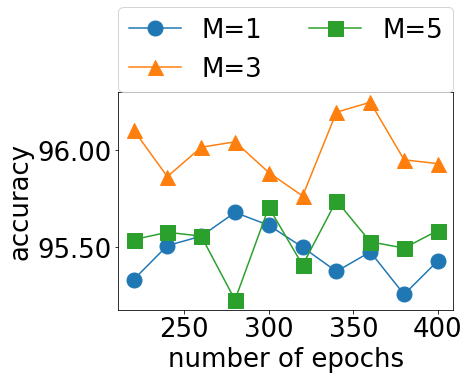}&
	\includegraphics[trim={0.2cm 0 0.2cm 0},clip,width=0.3\textwidth]{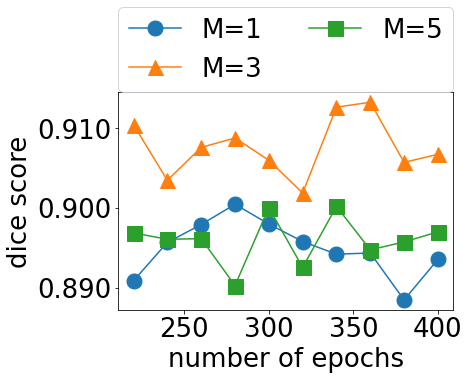} &
	\includegraphics[trim={0.2cm 0 0.2cm 0},clip,width=0.3\textwidth]{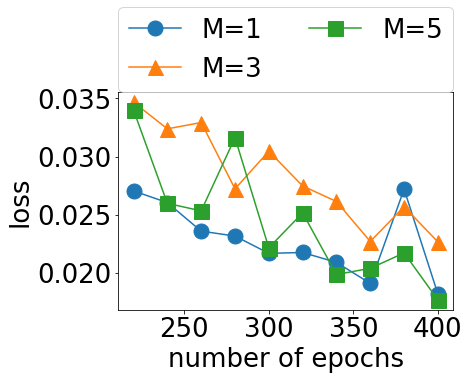}
\end{tabular}
\caption{Effect of the number of blocks $M$ for DN-II.}
\label{fig.concaUG.blk}
\end{figure}

For the number of blocks, we set $\tau=0.5, \lambda\varepsilon=1,\gamma=3$, and test $M=1,3,5$. The number of blocks corresponds to the number of iterations in the operator-splitting algorithm. In DN-II, since every block is a UNet class but with different weight parameters, the number of weight parameters scales almost linearly with $M$. For $M=1,3,5$, the number of parameters of the corresponding model is 3.07M, 9.21M, and 15.36M. The comparison of the accuracy, dice score, and training loss are shown in Figure \ref{fig.concaUG.blk}. We observe that the model with $M=3$ gives the highest accuracy. We think it is because we need sufficiently many parameters so that the model can learn a good approximate solver of the Potts model. When $M=1$, the number of parameters is too small, while when $M=5$, the increased number of parameters makes the model more complicated and more difficult to train.

\begin{figure}[t!]
\centering
\begin{tabular}{ccc}
	(a) & (b) & (c)\\
	\includegraphics[trim={0.2cm 0 0.2cm 0},clip,width=0.3\textwidth]{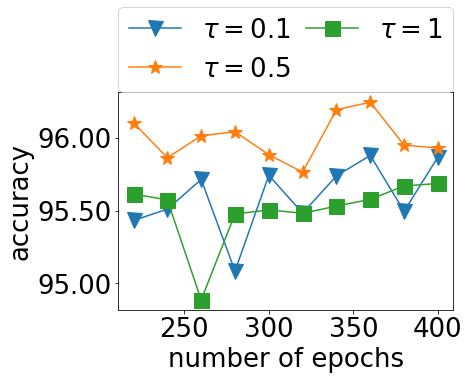}&
	\includegraphics[trim={0.2cm 0 0.2cm 0},clip,width=0.3\textwidth]{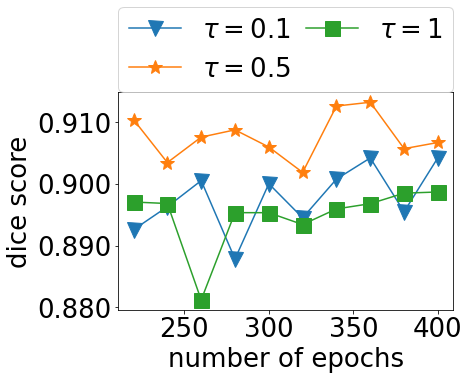} &
	\includegraphics[trim={0.2cm 0 0.2cm 0},clip,width=0.3\textwidth]{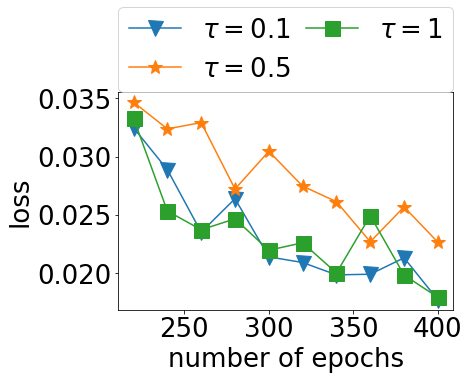}
\end{tabular}
\caption{Effect of the time step $\tau$ for DN-II.}
\label{fig.concaUG.time}
\end{figure}

We then set $M=3,\lambda\varepsilon=1,\gamma=3$ and test $\tau=0.1,0.5$ and $1$. The results are shown in Figure \ref{fig.concaUG.time}. Similar to our observations for DN-I, too small or too large $\tau$ leads to less accurate results.

\begin{figure}[t!]
\centering
\begin{tabular}{ccc}
	(a) & (b) & (c)\\
	\includegraphics[trim={0.2cm 0 0.2cm 0},clip,width=0.3\textwidth]{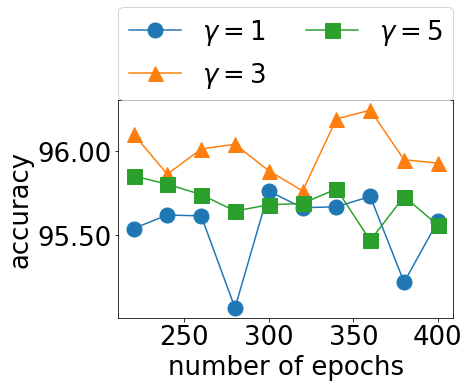}&
	\includegraphics[trim={0.2cm 0 0.2cm 0},clip,width=0.3\textwidth]{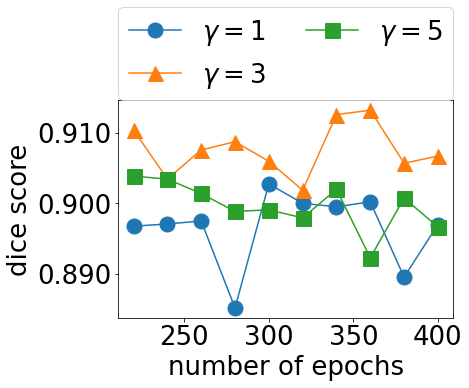} &
	\includegraphics[trim={0.2cm 0 0.2cm 0},clip,width=0.3\textwidth]{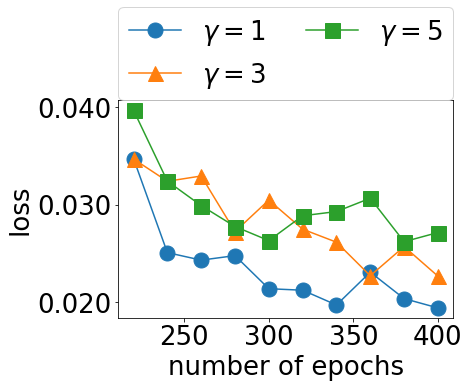}
\end{tabular}
\caption{Effect of the number of fixed point iteration $\gamma$ for DN-II.}
\label{fig.concaUG.iter}
\end{figure}

The next test is for the number of fixed-point iterations. We fix $M=3,\tau=0.5, \lambda\varepsilon=1$ and test $\gamma=1,3,5$. The results are shown in Figure \ref{fig.concaUG.iter}. Our observation is similar to that of DN-I. The model with $\gamma=3$ gives the best result.

\begin{figure}[t!]
\centering
\begin{tabular}{ccc}
	(a) & (b) & (c)\\
	\includegraphics[trim={0.2cm 0 0.2cm 0},clip,width=0.3\textwidth]{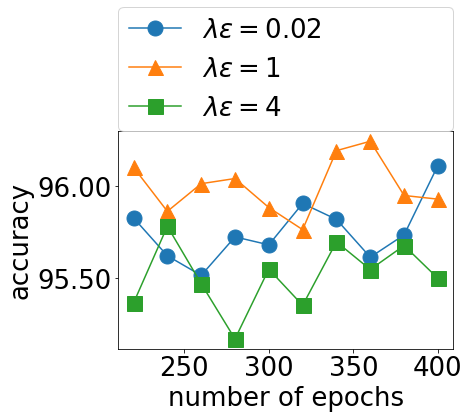}&
	\includegraphics[trim={0.2cm 0 0.2cm 0},clip,width=0.3\textwidth]{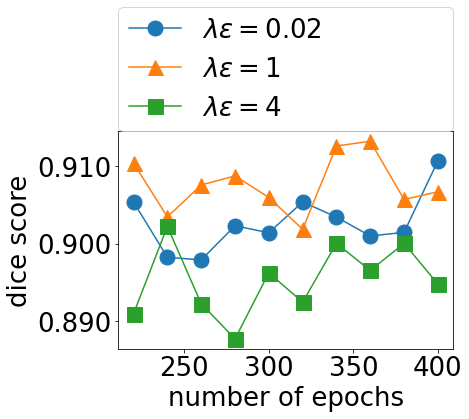} &
	\includegraphics[trim={0.2cm 0 0.2cm 0},clip,width=0.3\textwidth]{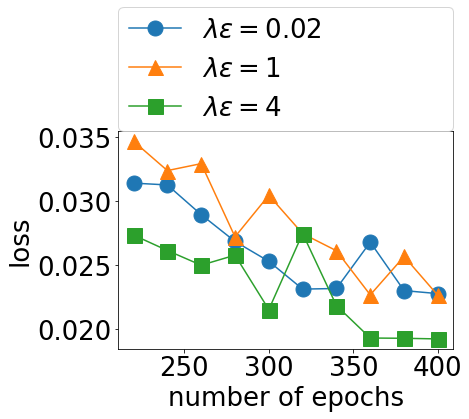}
\end{tabular}
\caption{Effect of the diffusion coefficient $\lambda$ for DN-II.}
\label{fig.concaUG.diff}
\end{figure}

In the last test, we study the effect of the diffusion coefficient $\lambda$. In this test, we set $M=3,\tau=0.5,\gamma=3$ and test %$\kappa=0.01,0.5, 2$
$\lambda\varepsilon=0.02,1$ and 4. The results are shown in Figure \ref{fig.concaUG.diff}. Again, Figure \ref{fig.concaUG.diff} suggests that explicitly adding this diffusion operator with a proper coefficient improves the result.

\subsection{The learned region force $F(f)$ of DN-I}

Note that DN-I is an operator-splitting scheme solving (\ref{eq.control}), in which $F(f)$ is the region force functional for  the Potts model. By training DN-I, the region force is learned from the data. Under the same setting as in Section \ref{sec.compare} for MARA10K, in Figure \ref{fig.F}, we show the learned region force $F(f)$ for two images. We observe the learned region force $F(f)$ highlights the region to be segmented in white, while  edges and textures information is also reflected in the learner region force $F(f)$. This clearly shows that  DB-I provides a data-driven way to learn the region force functional in the Potts model. The evolution of the segmentation results are shown in Figure \ref{fig.F_int}. The initial condition, output of the 3rd, 8th, 10th and the final layer are presented. Note that the final segmentation is generated by thresholding the output of the final layer. Starting from the initial condition, the segmentation is driven by region force $F(f)$ and control variables towards the ground truth segmentation.

We then examine the learned region force $F(f)$ on noisy data. For the network training, we adopt the progressive training strategy proposed in \cite[Sec 7.3]{tai2023pottsmgnet}. During training, we gradually increase the noise standard deviation (SD) as 0, 0.3, 0.5, 0.8, 1.0. For each noise level, we train the network for 400 epochs and then use the trained network parameters as the initial value for training the network for the next SD. The performance of the final trained network {on training data with SD=1.0 is then tested on images with various noise levels. Two examples with SD=0.2 and SD=0.5  are shown in Figure \ref{fig.F.prog}. Even with substantial noise, DN-I still gives good results. The learned region force $F(f)$ highlights the region of the segmentation target. The evolution of the segmentation results are shown in Figure \ref{fig.F_int.prog}. Similar to our observation in Figure \ref{fig.F_int}, the segmentation is driven by region force $F(f)$ and control variables towards the ground truth segmentation.

\begin{figure}[t!]
	\centering
	\begin{tabular}{cccc}
		Image & Ground truth & \makecell{DB-I Segmentation}& \makecell{Region force\\$(1-F(f))$}\\
		\includegraphics[width=0.2\textwidth]{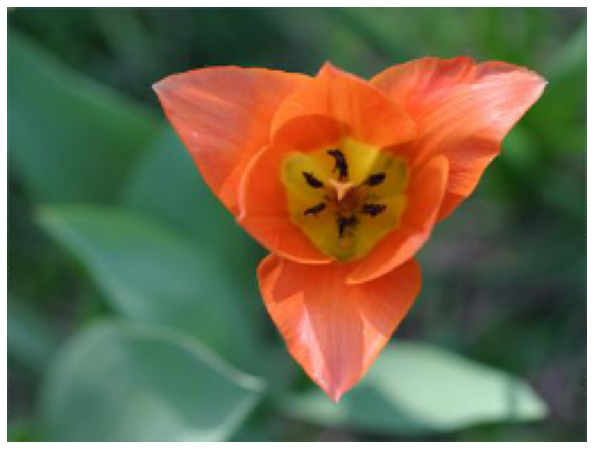} & 
		\includegraphics[width=0.2\textwidth]{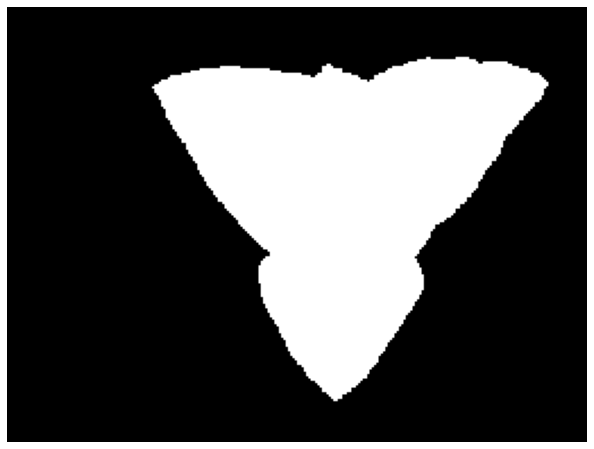} &
		\includegraphics[width=0.2\textwidth]{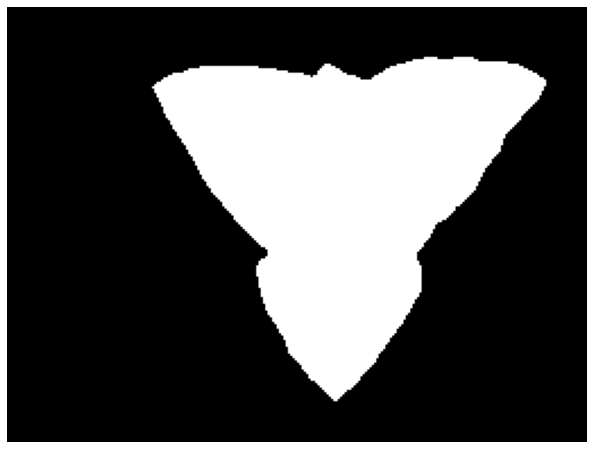} &
		\includegraphics[width=0.2\textwidth]{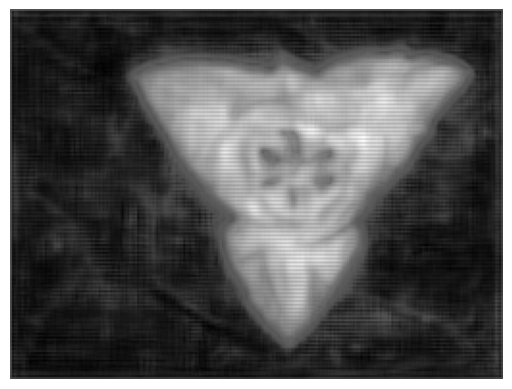}\\
		\includegraphics[width=0.2\textwidth]{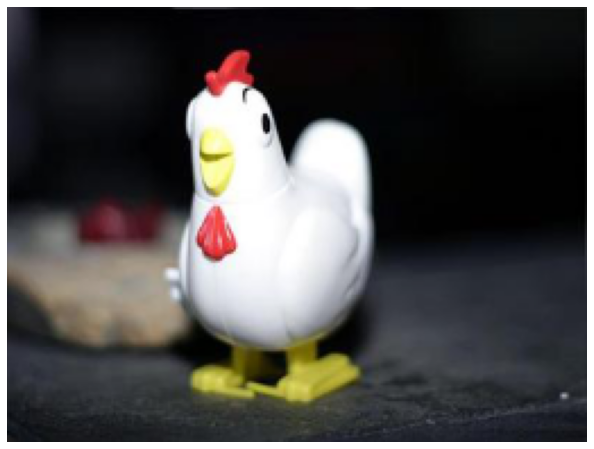} & 
		\includegraphics[width=0.2\textwidth]{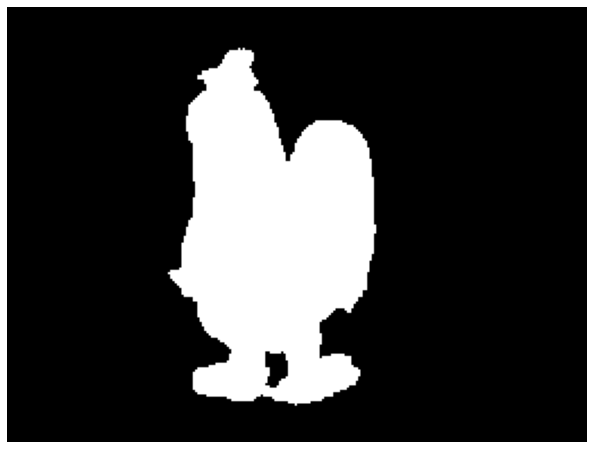} &
		\includegraphics[width=0.2\textwidth]{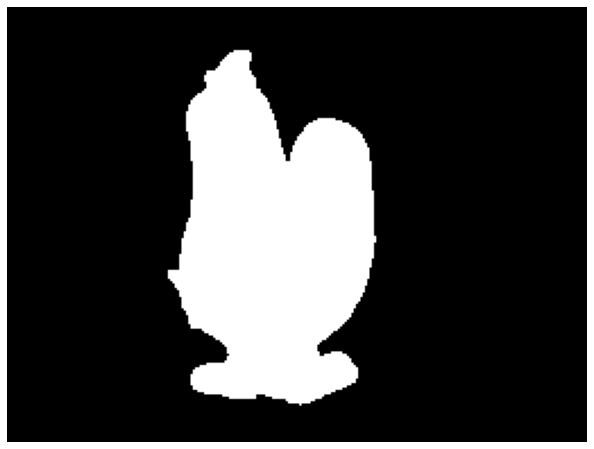} &
		\includegraphics[width=0.2\textwidth]{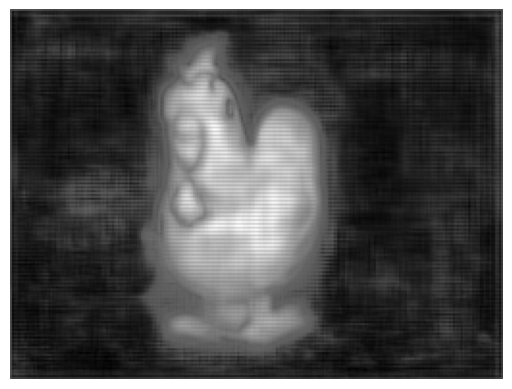}
	\end{tabular}
	\caption{The learned region force $F(f)$ in DN-I on clean images. For better visualization, we normalized $F(f)$ to be in $[0,1]$.}
	\label{fig.F}
\end{figure}

\begin{figure}[t!]
	\centering
	\begin{tabular}{ccccc}
		Initial & Block 3 & Block 8& Block 10 & Final layer\\
		\includegraphics[width=0.16\textwidth]{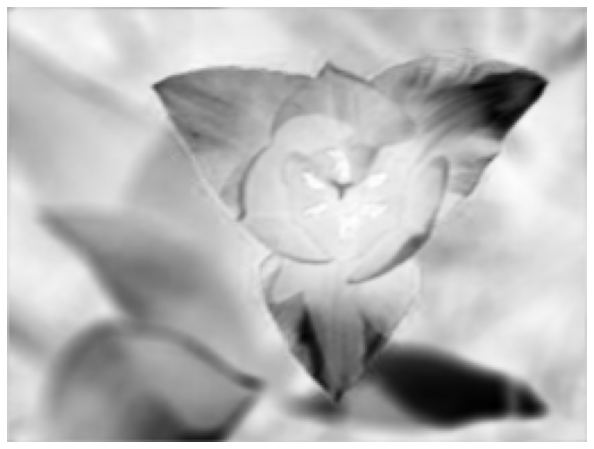} & 
		\includegraphics[width=0.16\textwidth]{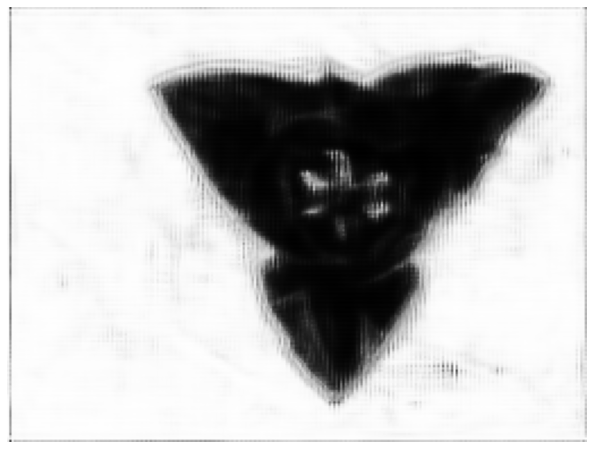} & 
		\includegraphics[width=0.16\textwidth]{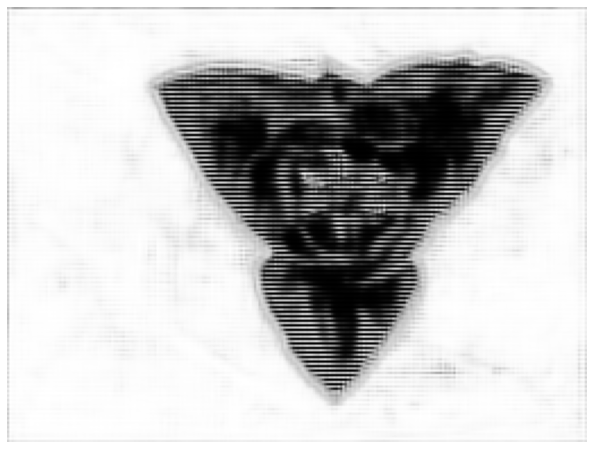} & 
		\includegraphics[width=0.16\textwidth]{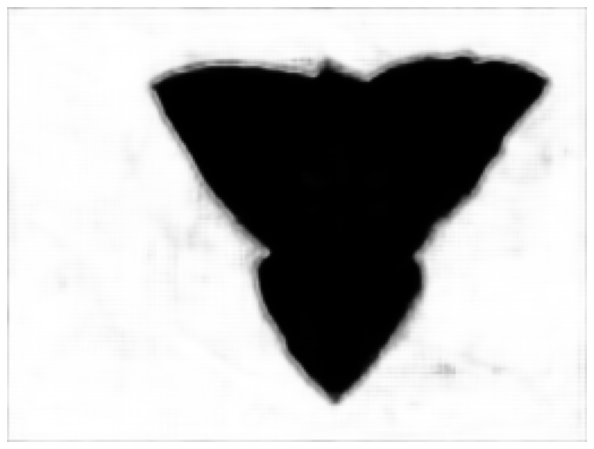} &
		\includegraphics[width=0.16\textwidth]{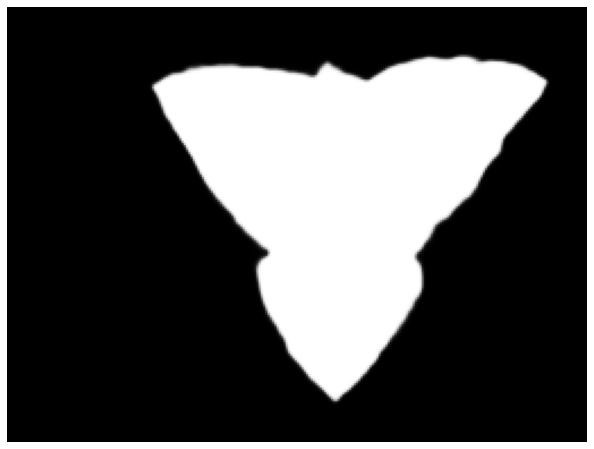} \\
		\includegraphics[width=0.16\textwidth]{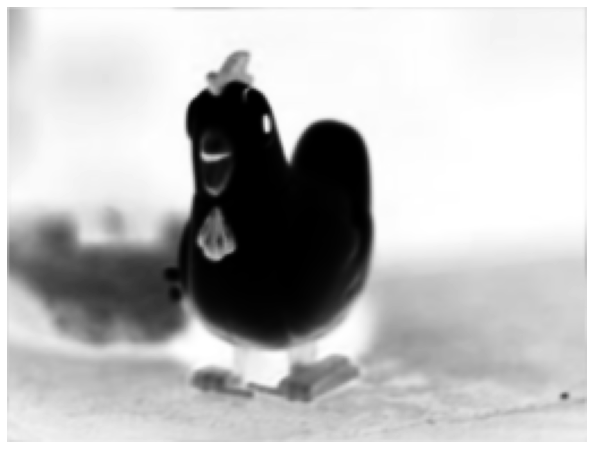} & 
		\includegraphics[width=0.16\textwidth]{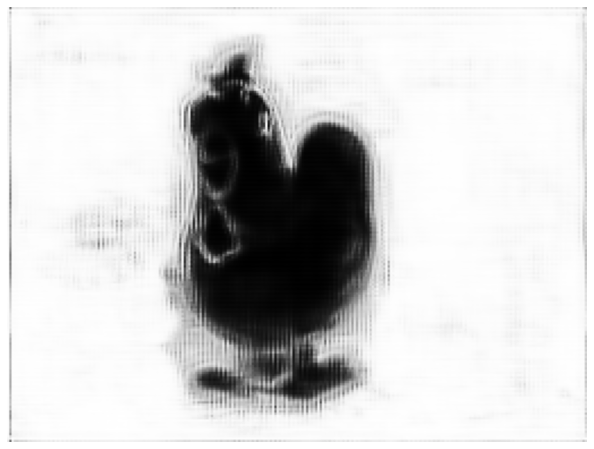} & 
		\includegraphics[width=0.16\textwidth]{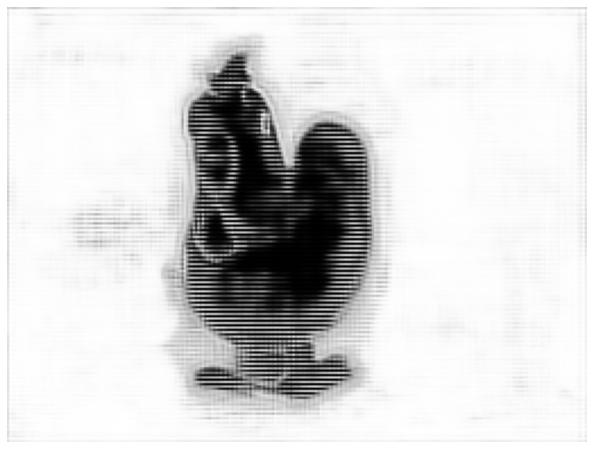} & 
		\includegraphics[width=0.16\textwidth]{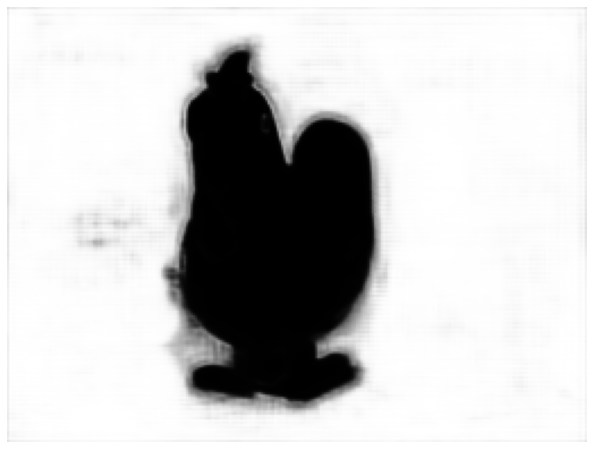} &
		\includegraphics[width=0.16\textwidth]{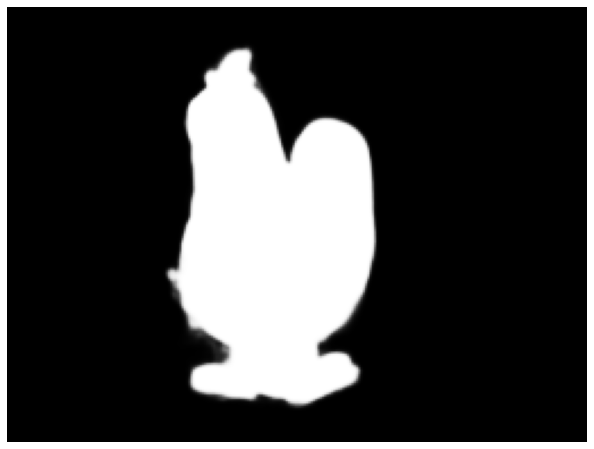} 
	\end{tabular}
	\caption{Evolution of the segmentation across DB-I's in DN-I. Ten DB-I's are used in this experiment.}
	\label{fig.F_int}
\end{figure}

\begin{figure}[t!]
	\centering
	\begin{tabular}{cccc}
		Image & Ground truth & DB-I Segmentation& \makecell{Region force \\ $(1-F(f))$}\\
		\includegraphics[width=0.2\textwidth]{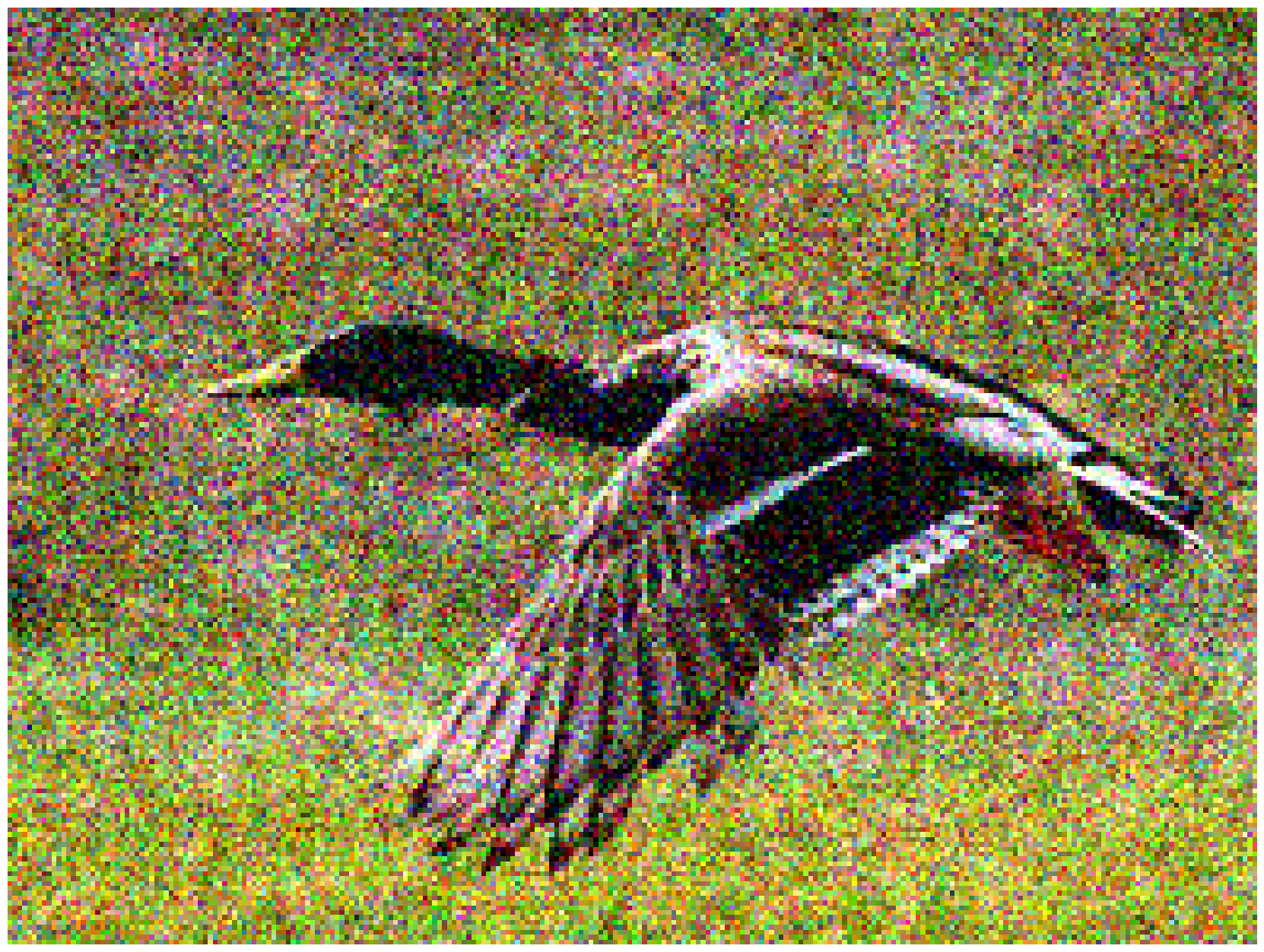} & 
		\includegraphics[width=0.2\textwidth]{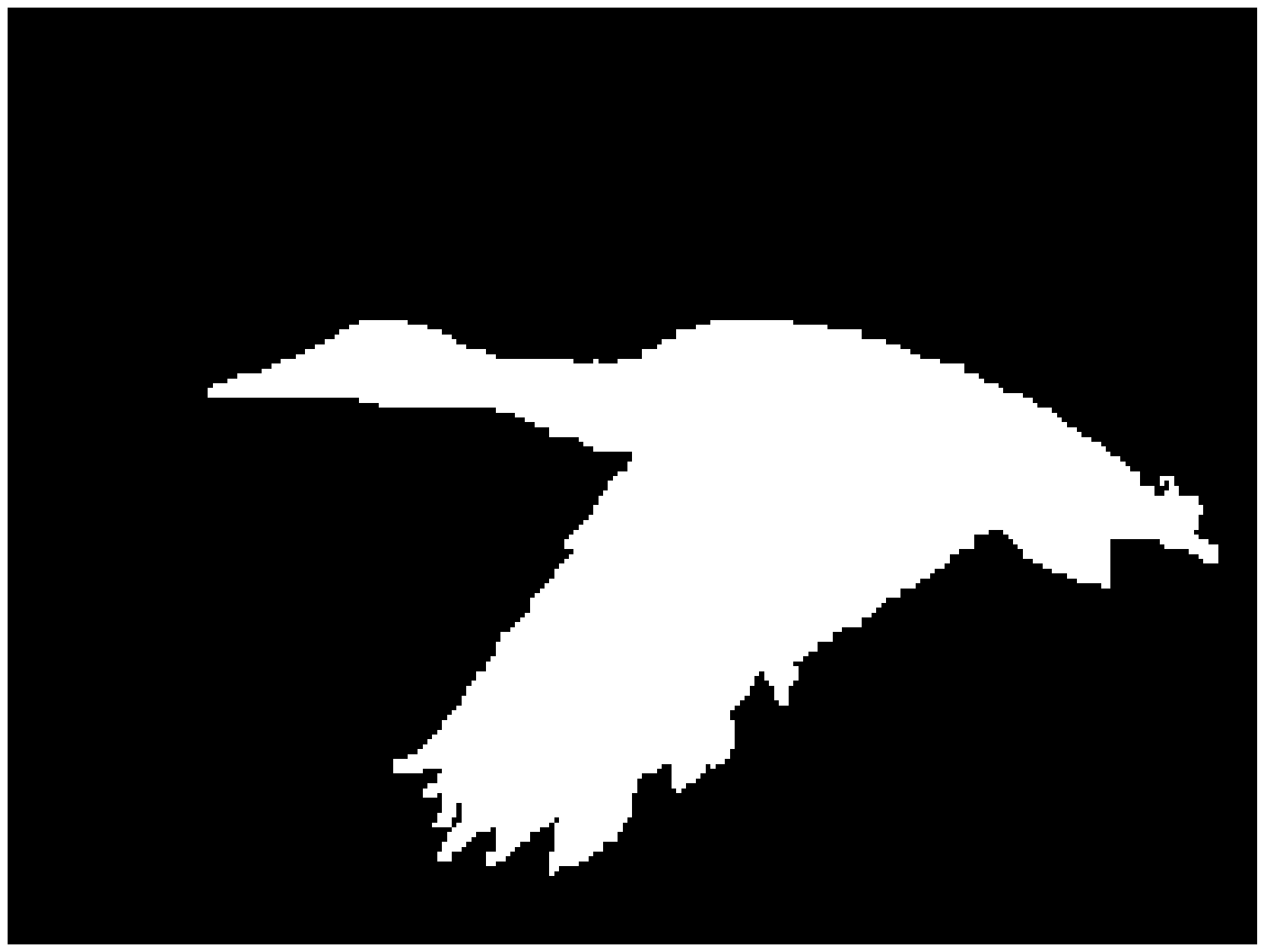} &
		\includegraphics[width=0.2\textwidth]{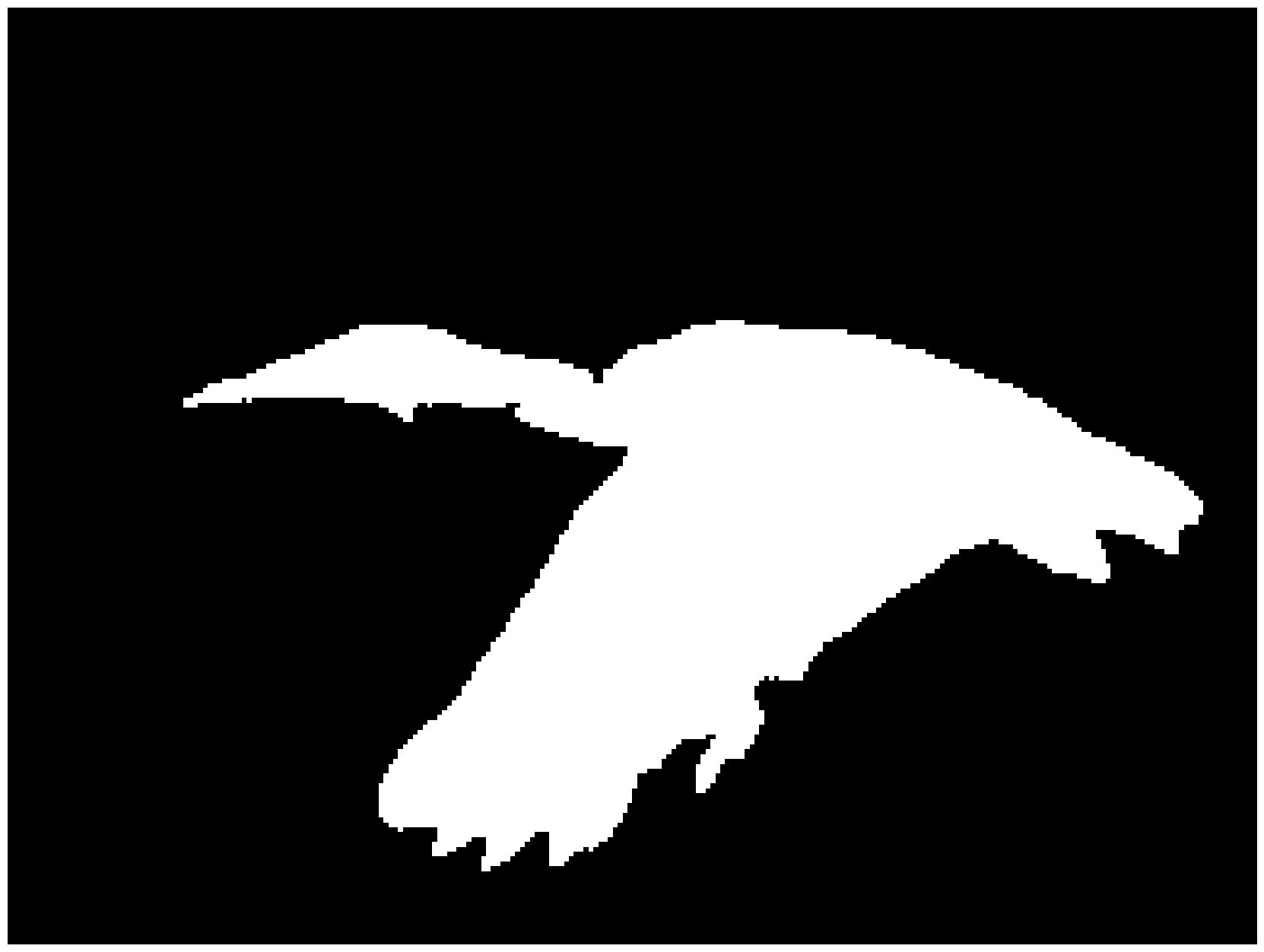} &
		\includegraphics[width=0.2\textwidth]{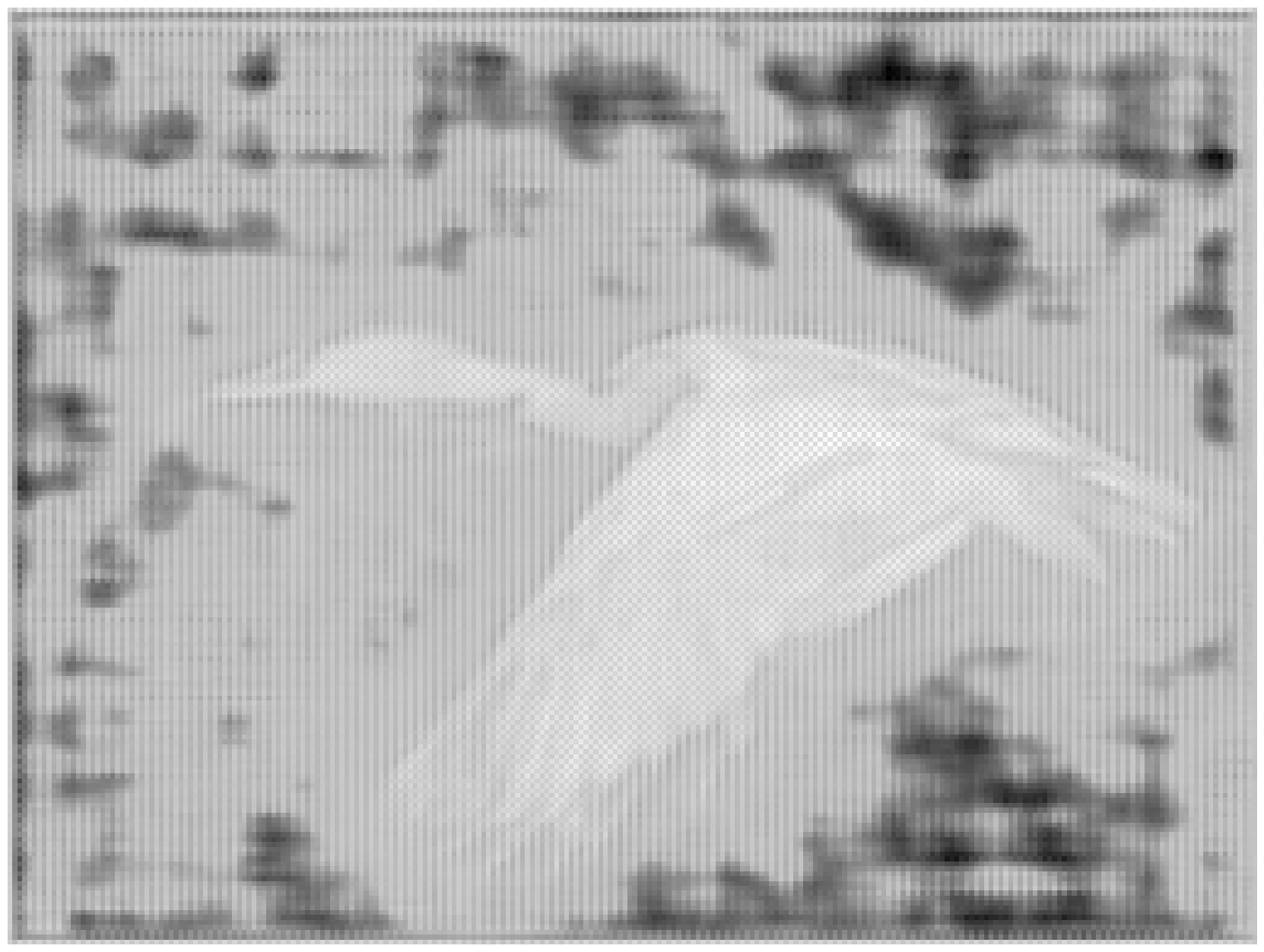}\\
		\includegraphics[width=0.2\textwidth]{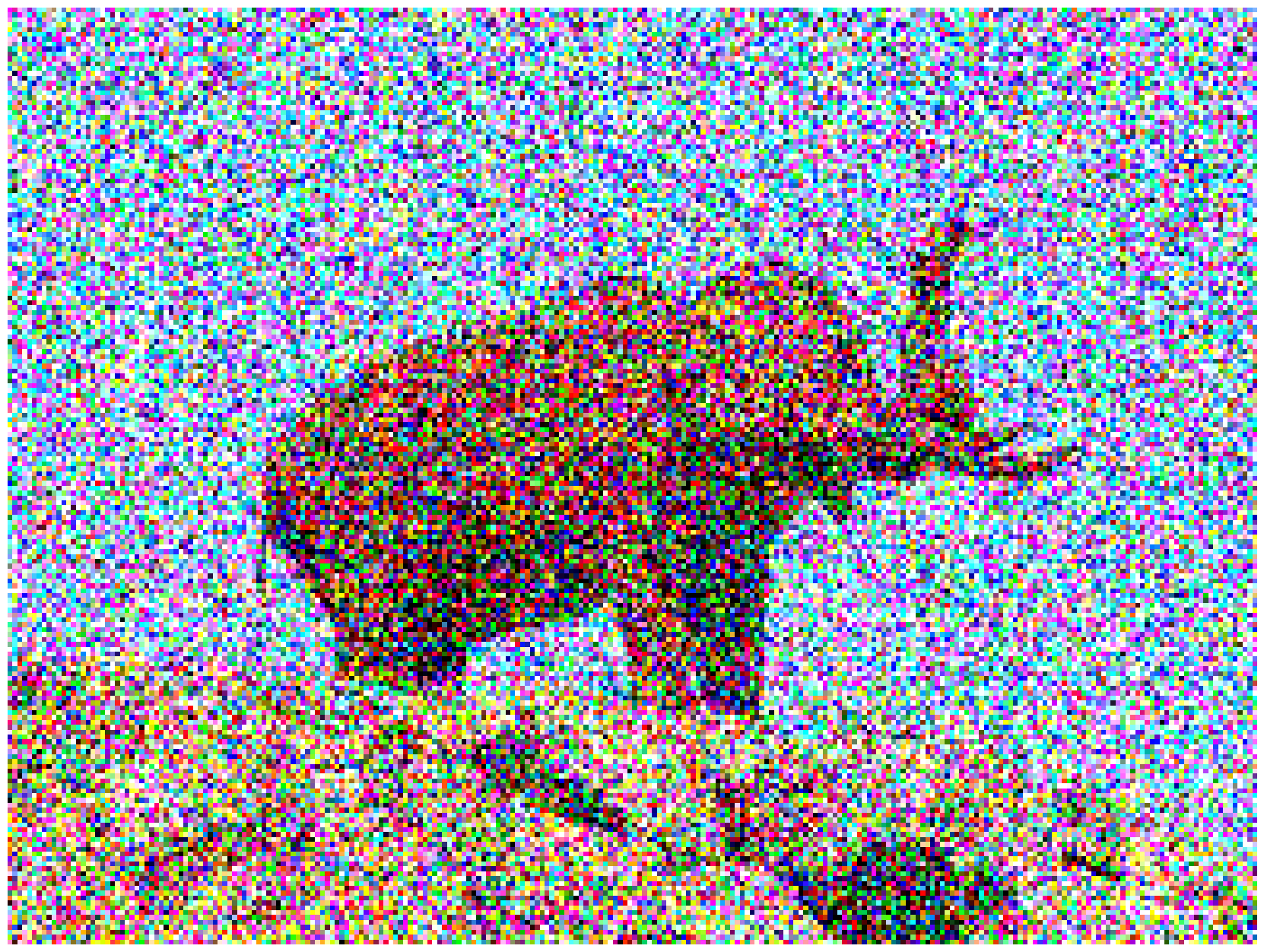} & 
		\includegraphics[width=0.2\textwidth]{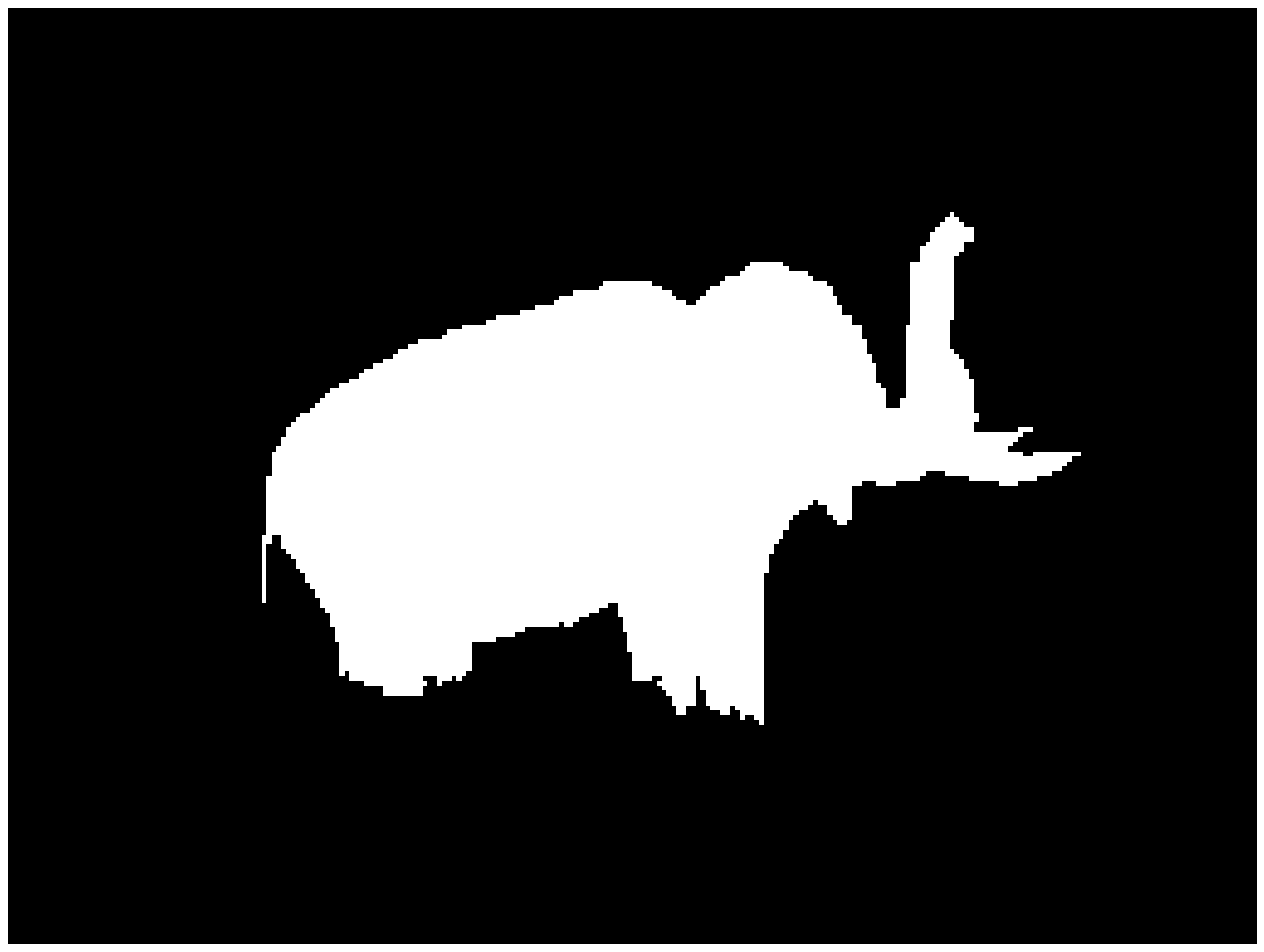} &
		\includegraphics[width=0.2\textwidth]{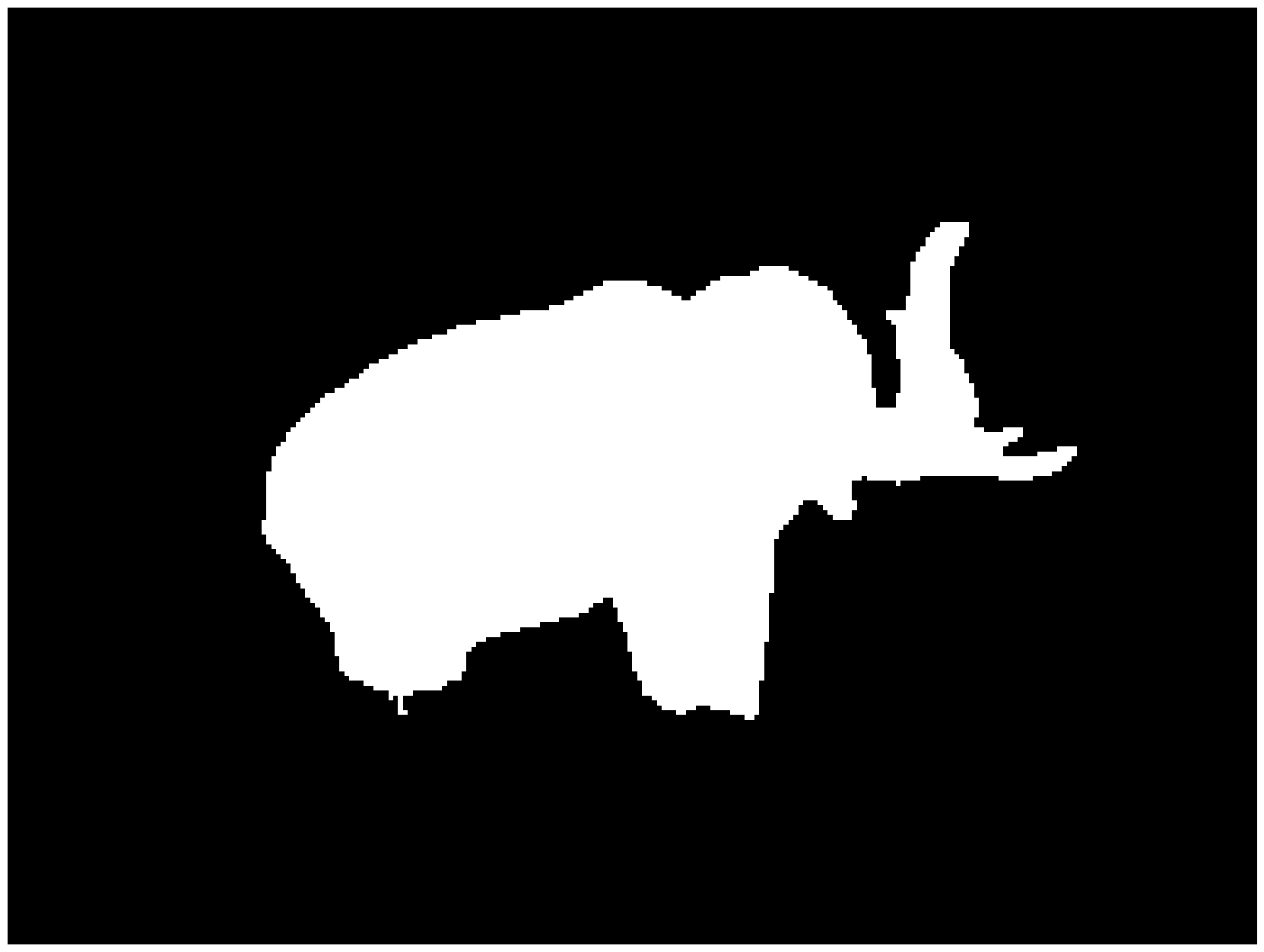} &
		\includegraphics[width=0.2\textwidth]{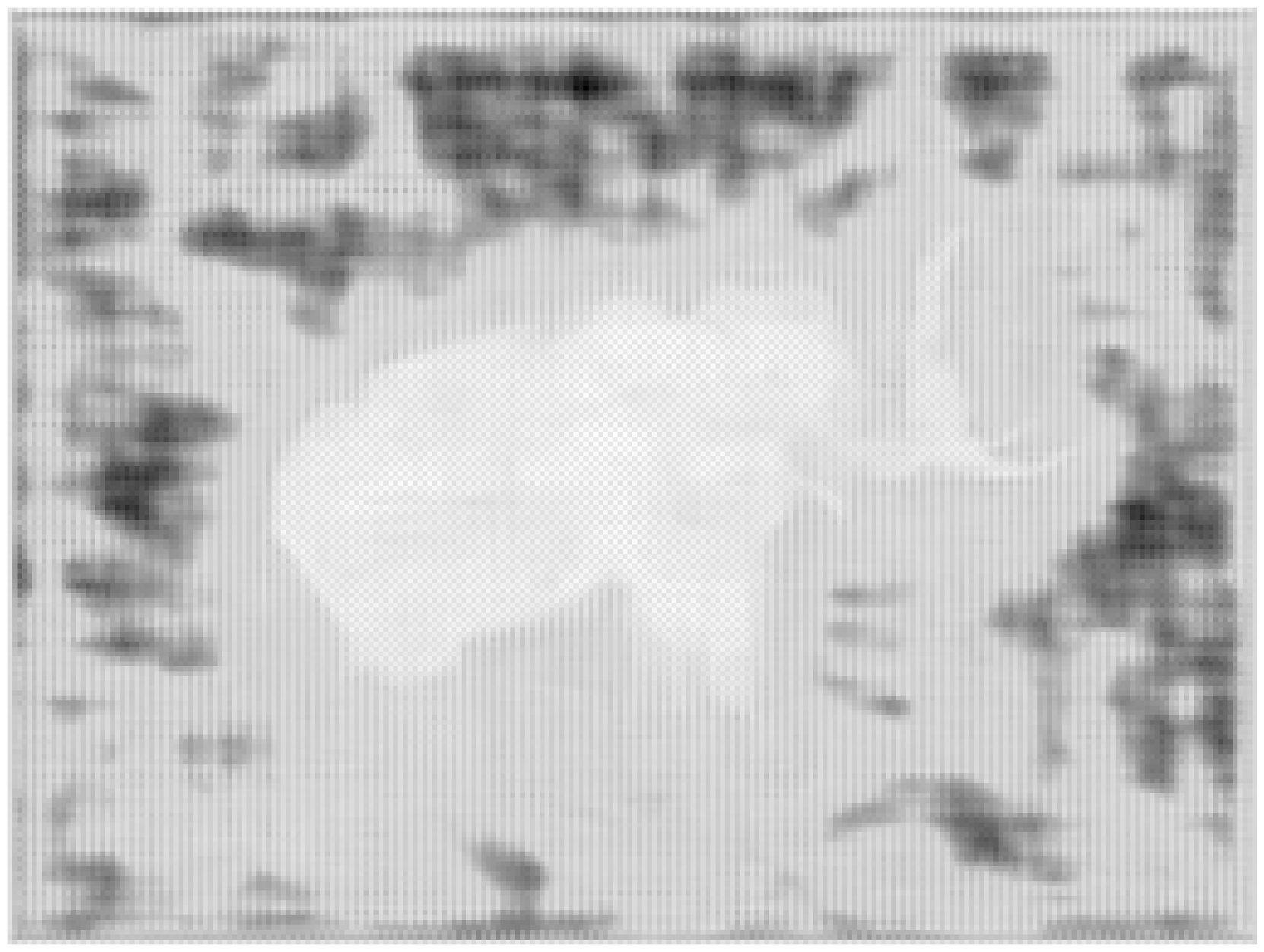}
	\end{tabular}
	\caption{The learned region force $F(f)$ in DN-I on noisy images with (a) SD 0.2 and (b) SD 0.5. The DN-I is trained by progressive training. For better visualization, we normalized $F(f)$ to be in $[0,1]$ and display $1-F(f)$. }
	
	\label{fig.F.prog}
\end{figure}

\begin{figure}[t!]
	\centering
	\begin{tabular}{ccccc}
		Initial & Block 3& Block 8& Block 10& Final layer\\
		\includegraphics[width=0.16\textwidth]{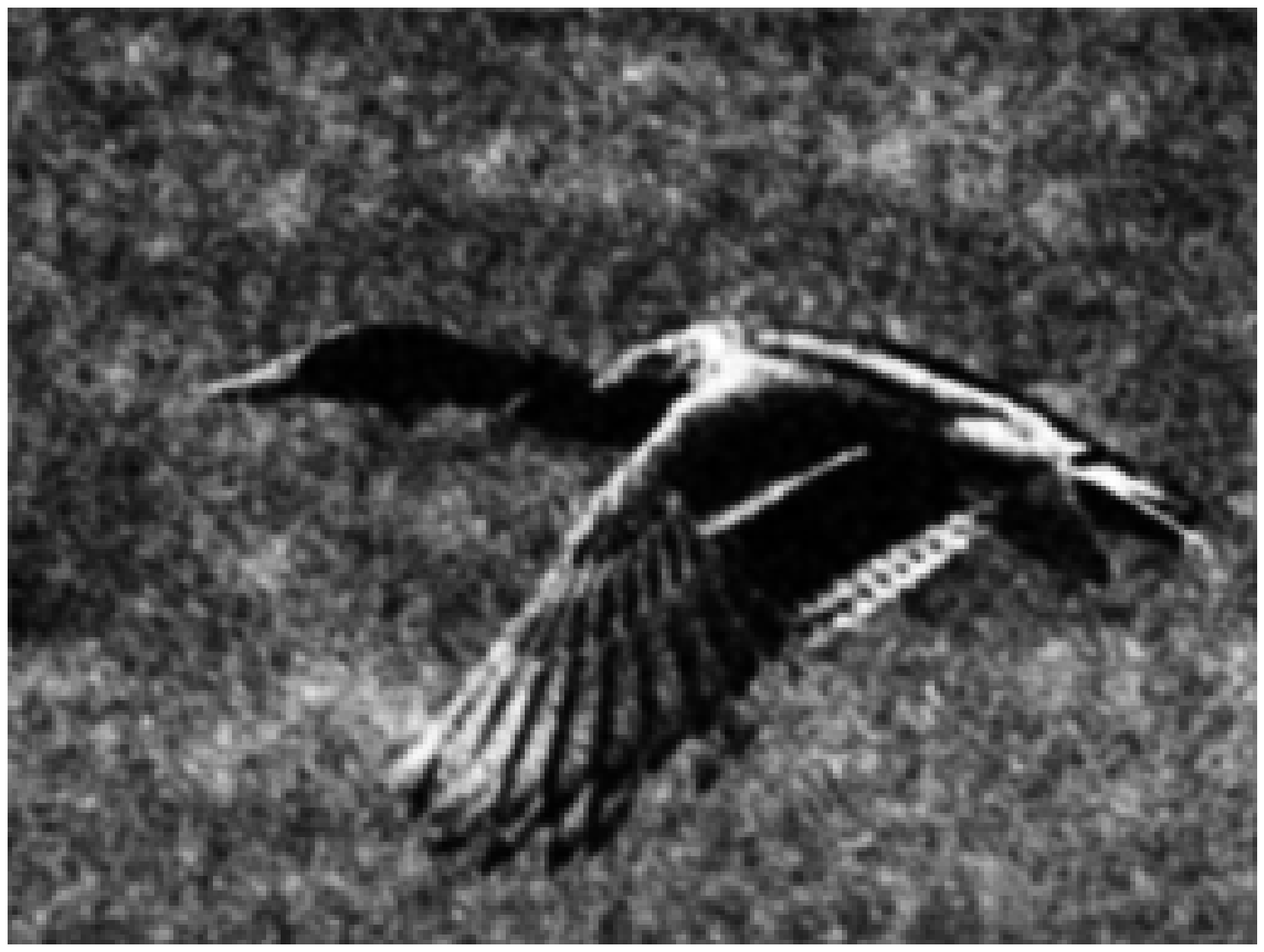} & 
		\includegraphics[width=0.16\textwidth]{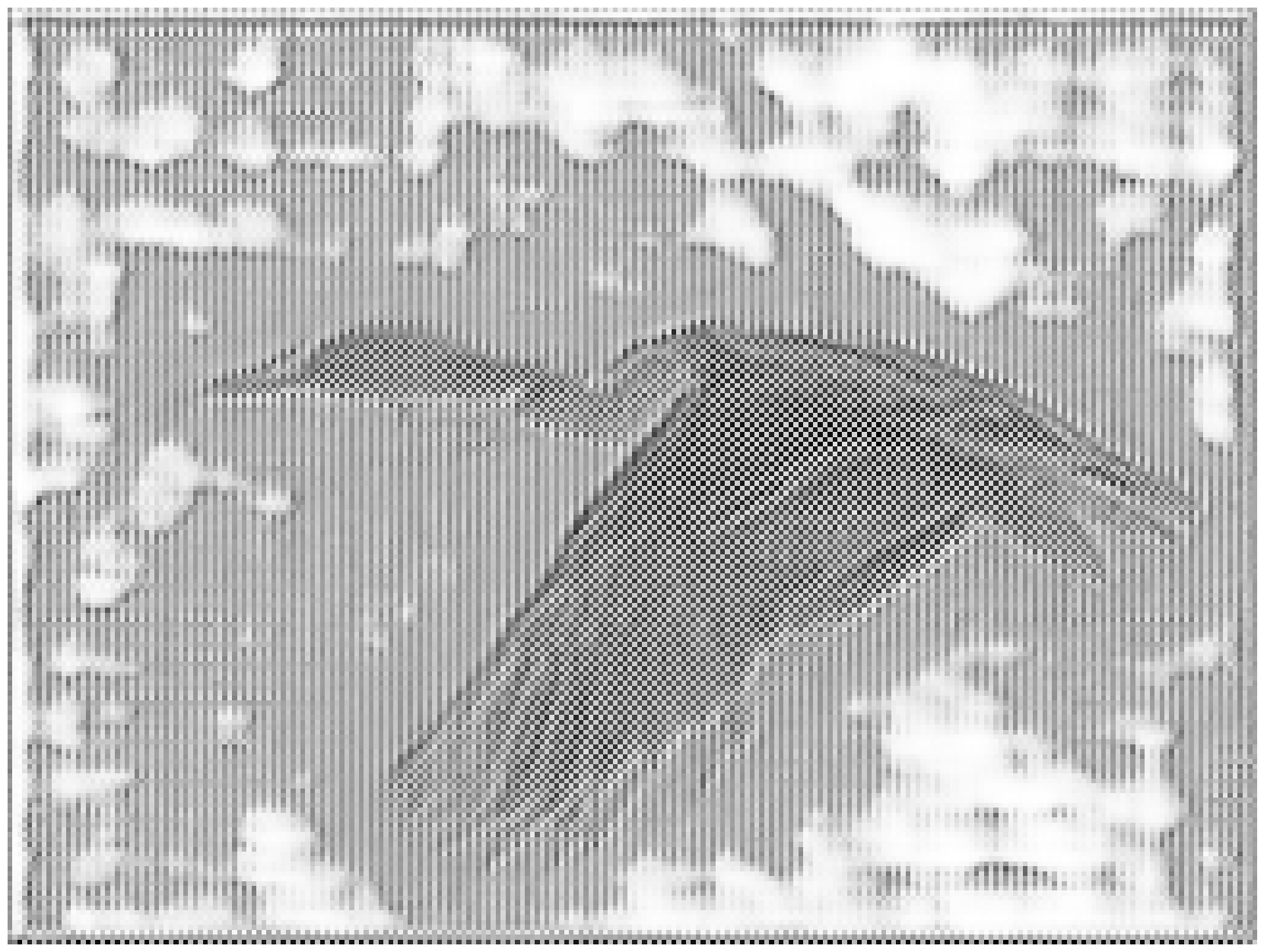} & 
		\includegraphics[width=0.16\textwidth]{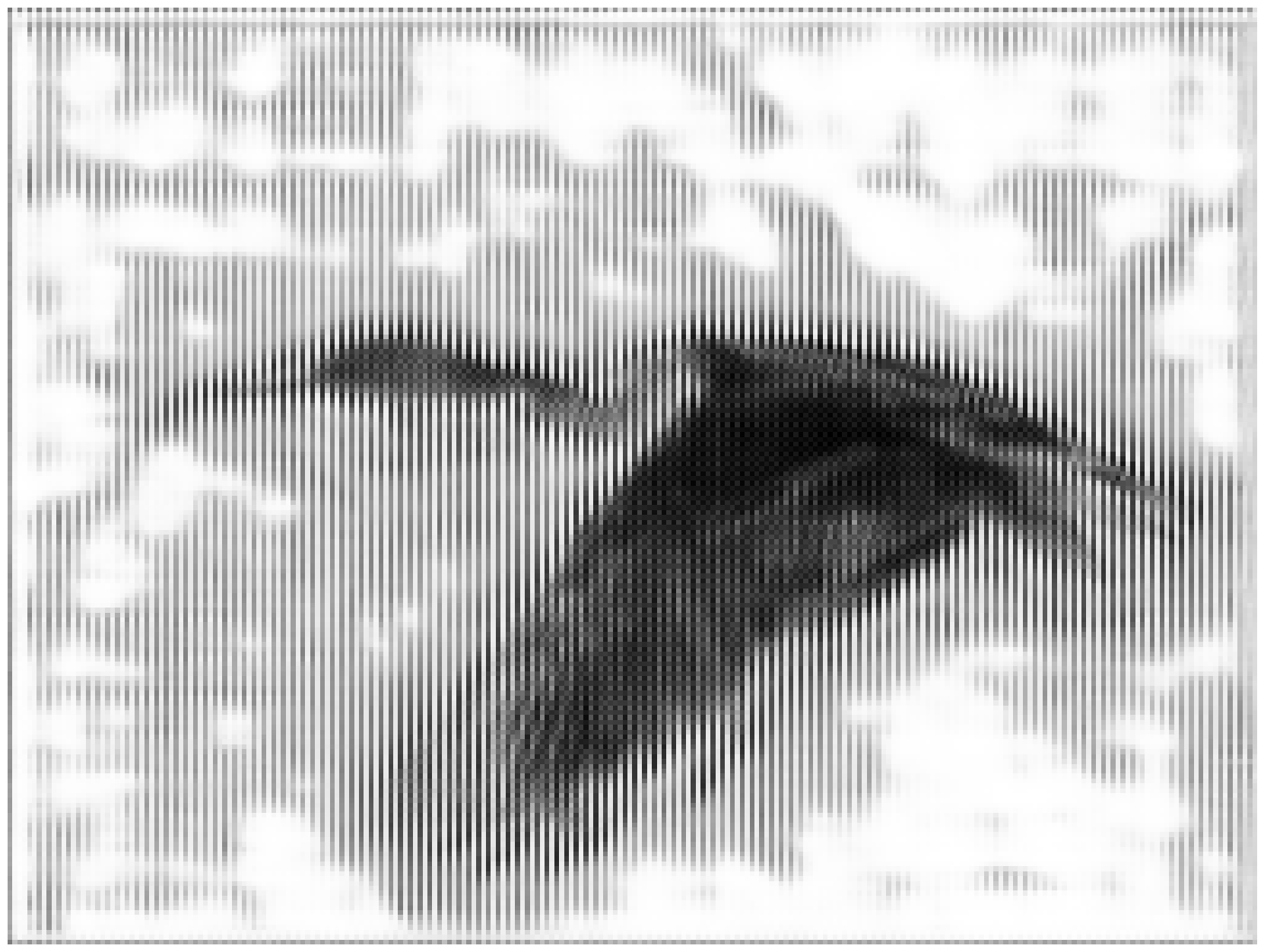} & 
		\includegraphics[width=0.16\textwidth]{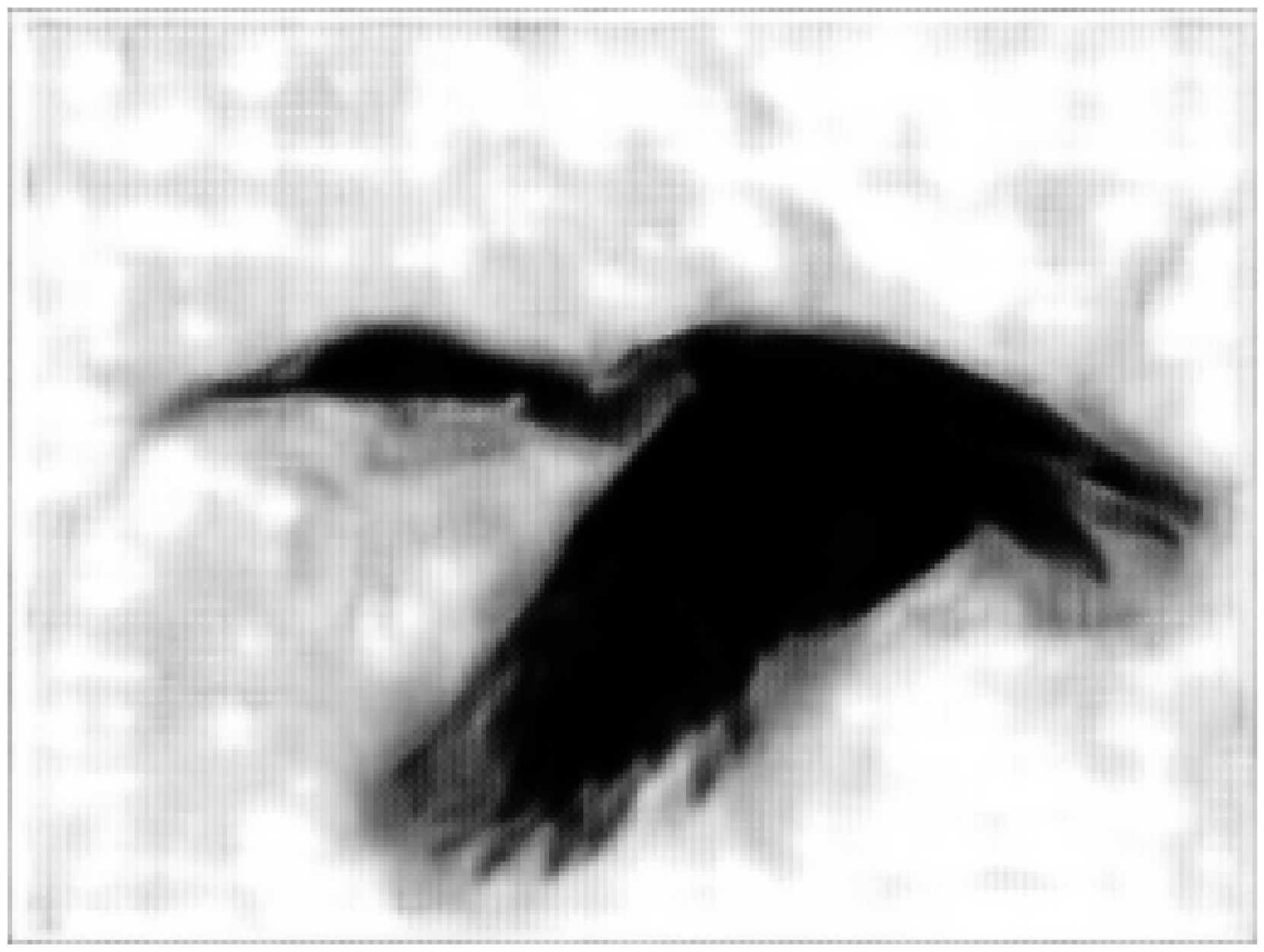} &
		\includegraphics[width=0.16\textwidth]{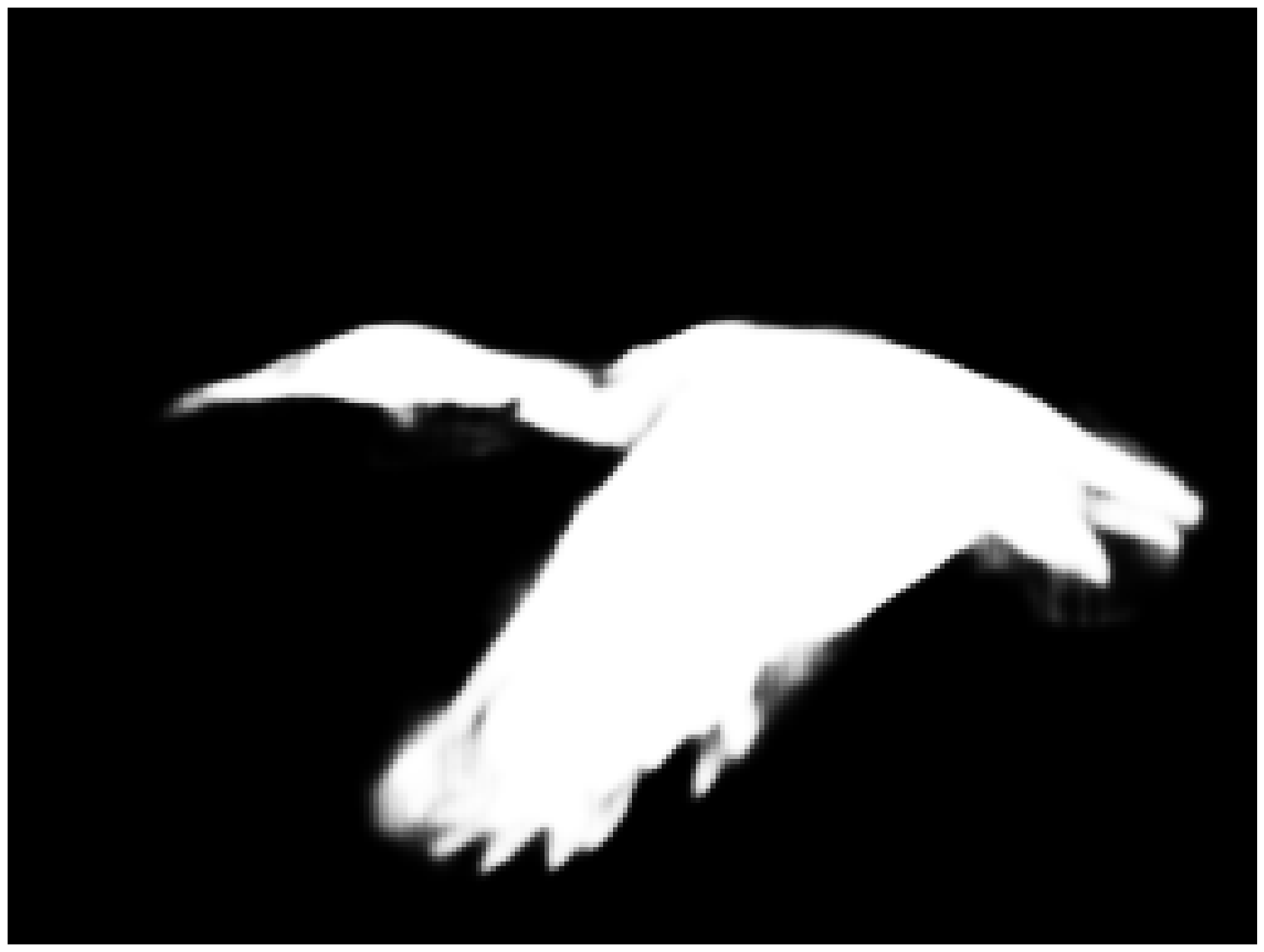} \\
		\includegraphics[width=0.16\textwidth]{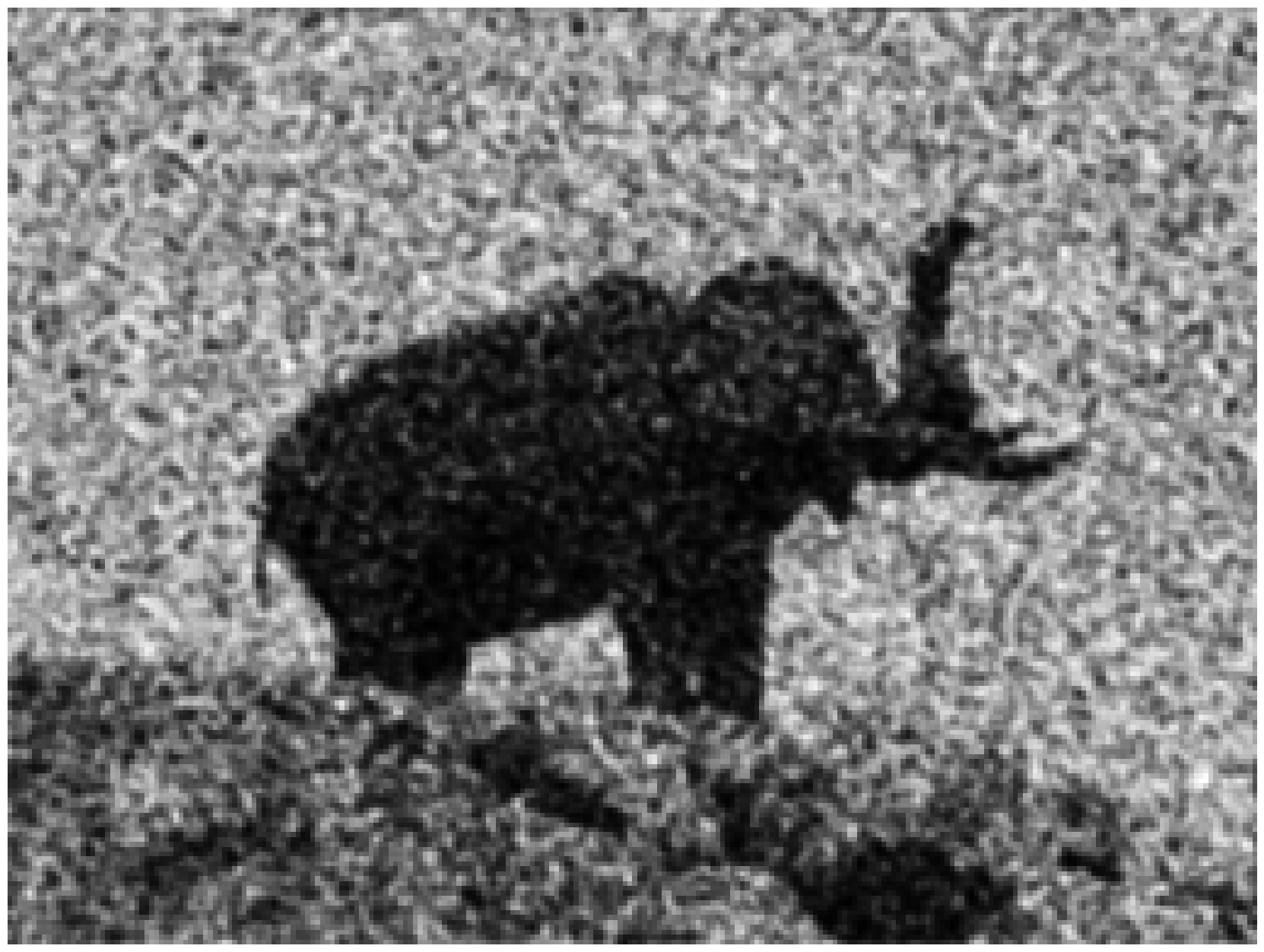} & 
		\includegraphics[width=0.16\textwidth]{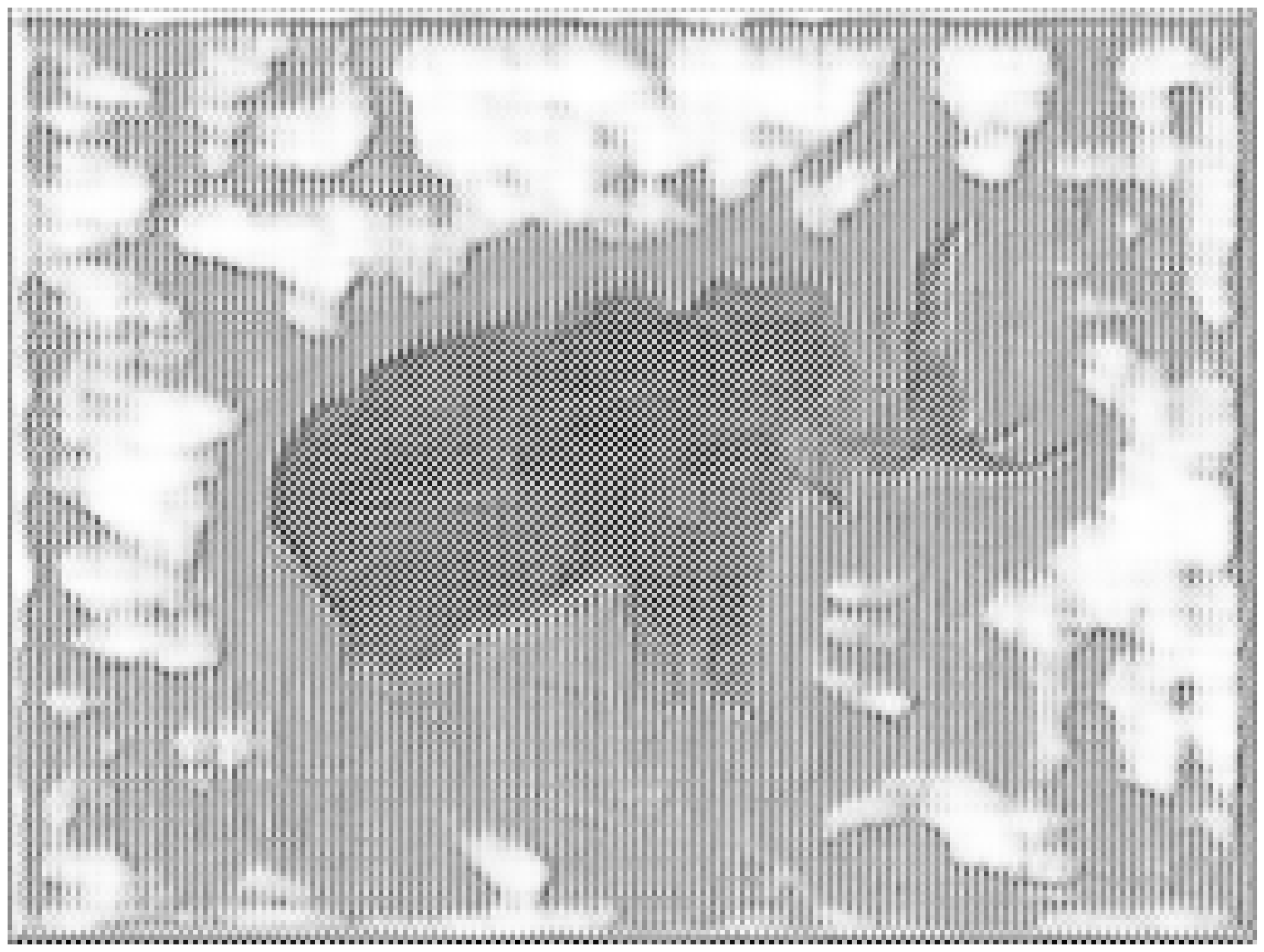} & 
		\includegraphics[width=0.16\textwidth]{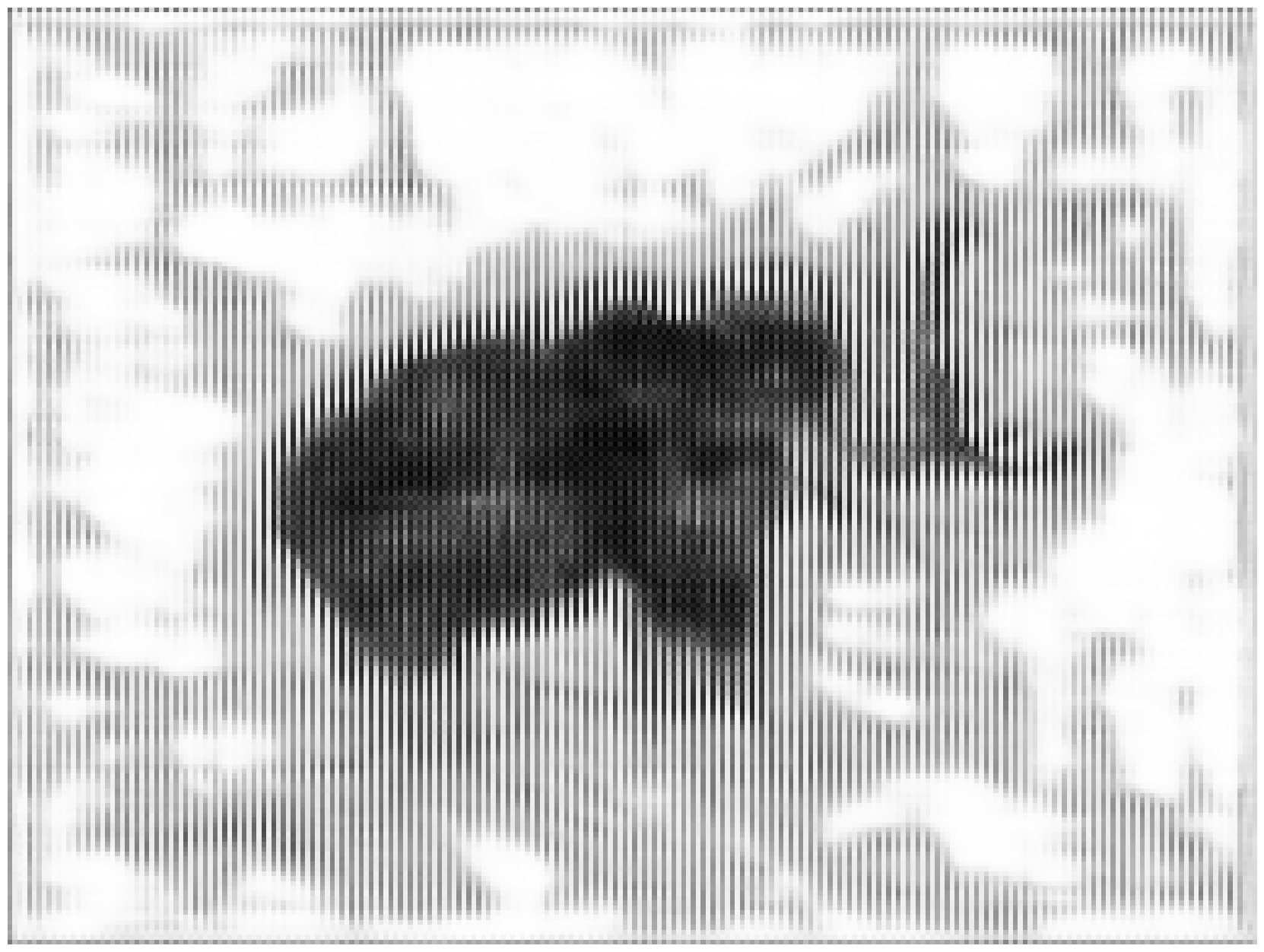} & 
		\includegraphics[width=0.16\textwidth]{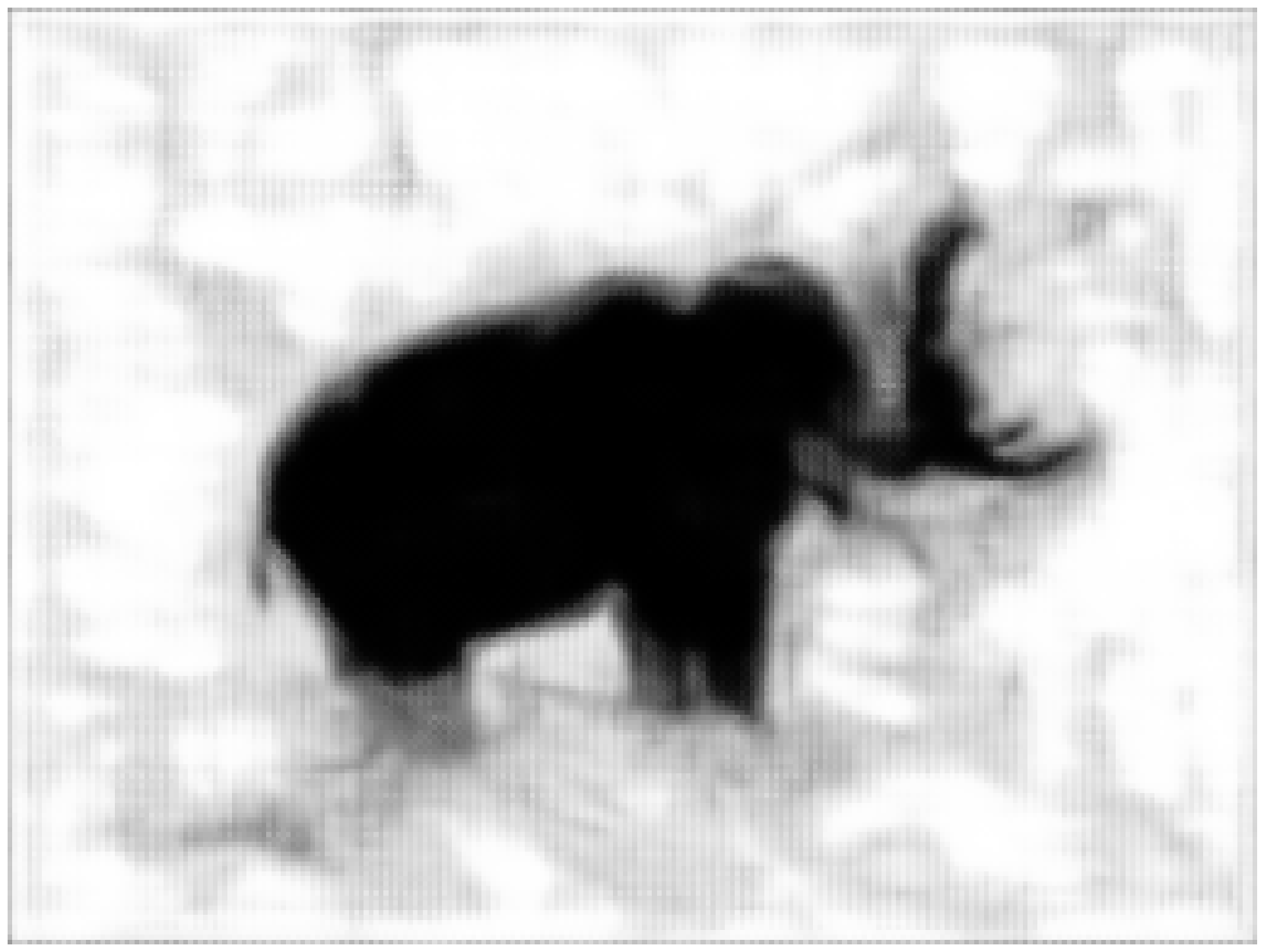} &
		\includegraphics[width=0.16\textwidth]{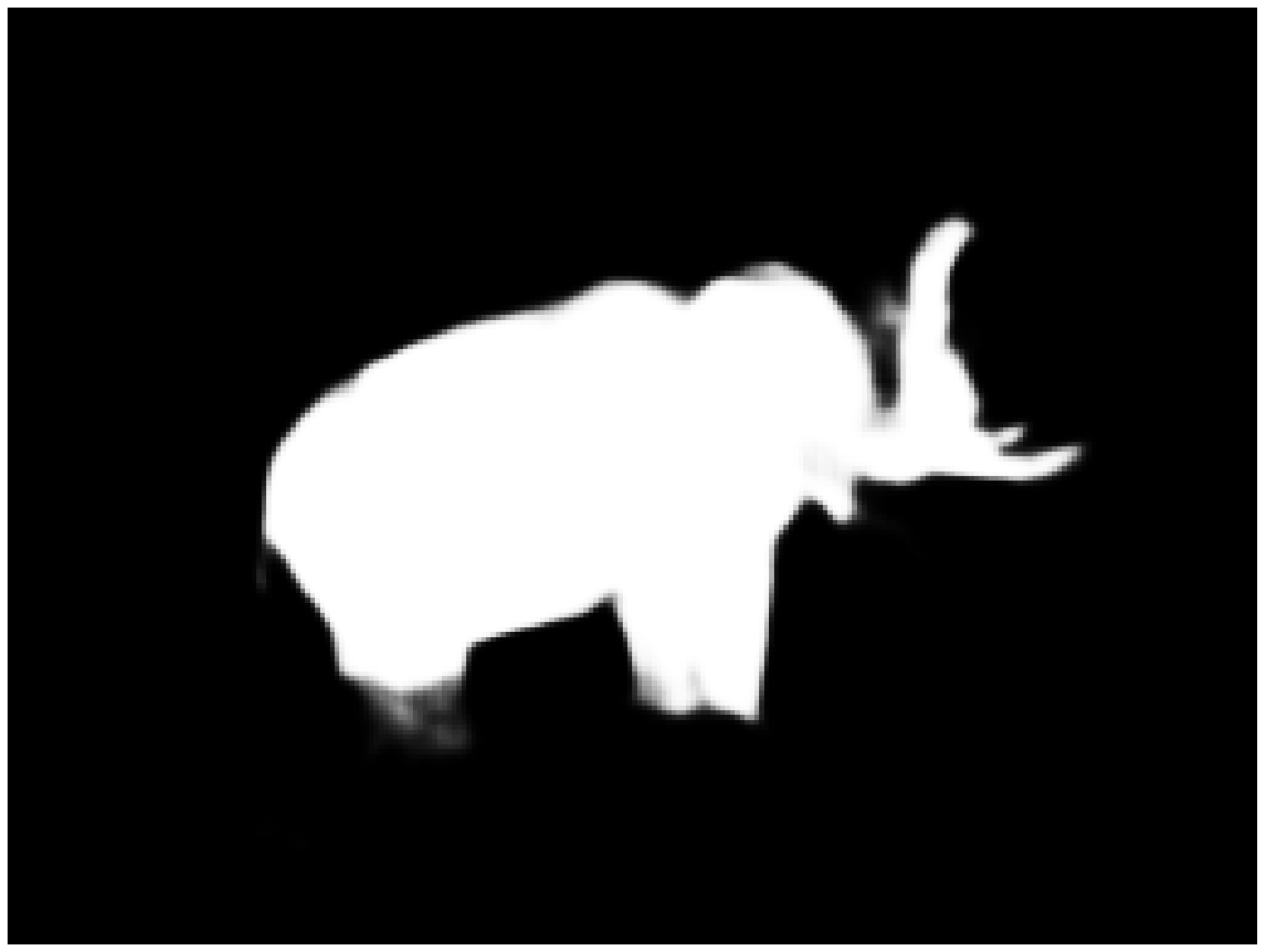} 
	\end{tabular}
	\caption{Evolution of the segmentation across DB-I's in DN-I on noisy images with (a) SD 0.2 and (b) SD 0.5. The DN-I is trained by progressive training.. Ten DB-I's are used in this experiment.}
	\label{fig.F_int.prog}
\end{figure}

\section{Conclusion} \label{sec.conclusion}
In this paper, we propose two models for image segmentation, Double-well Net I and Double-well Net II, which are based on the Potts model, network approximation theory and operator-splitting methods. Starting from the Potts model, we approximate it using a double-well potential and then propose two control problems to find its minimizer. The control problems are time discretized by operator splitting methods whose structures are similar to the building block of neural networks. We then define a UNet class to represent some operators in the operator-splitting methods, which together with control variables are trained from data, leading to DN-I and DN-II. The UNet class is designed to capture multiscale features of images. DN-I and DN-II consist of several blocks, each of which corresponds to a one-time stepping of the operator-splitting schemes. All blocks use an activation that is a fixed-point iteration to minimize a double-well potential with a proximal term. Both networks have a mathematical explanation: they are operator-splitting methods to approximately solve the Potts model, which provide a new perspective for network design. The proposed models also bridge the MBO scheme with deep neural networks. In addition, since the region force functional in the Potts model is represented as a network in DN-I, DN-I provides a data-driven way to learn the region force functional. Systematic numerical experiments are conducted to study the performance of the proposed models. Compared to state-of-the-art networks, the proposed models provide higher accuracy and dice scores with a much smaller number of parameters.

	\bibliographystyle{abbrv}
	\bibliography{ref}
\end{document}